\documentclass{article} 
\usepackage{iclr2027_conference,times}

\usepackage[T1]{fontenc}

\usepackage{amsmath,amsfonts,bm}

\def\eqref#1{equation~\ref{#1}}
\def\1{\bm{1}}

\DeclareMathAlphabet{\mathsfit}{\encodingdefault}{\sfdefault}{m}{sl}
\SetMathAlphabet{\mathsfit}{bold}{\encodingdefault}{\sfdefault}{bx}{n}

\newcommand{\R}{\mathbb{R}}

\newcommand{\softmax}{\mathrm{softmax}}

\newcommand{\KL}{D_{\mathrm{KL}}}

\usepackage{hyperref}
\hypersetup{hidelinks}
\usepackage{url}
\usepackage{microtype}
\usepackage{graphicx}
\usepackage{wrapfig}
\usepackage{subcaption}
\usepackage{placeins}
\usepackage{needspace}
\usepackage{booktabs}
\usepackage{amsmath}
\usepackage{amssymb}
\usepackage{mathtools}
\usepackage{amsthm}
\usepackage{xcolor}
\usepackage[capitalize,noabbrev]{cleveref}
\usepackage{tikz}
\usepackage[most]{tcolorbox}
\usepackage{fvextra}

\newtcolorbox{promptbox}[1]{
    title={#1},
    colback=gray!5,
    colframe=black!60,
    boxrule=0.5pt,
    arc=2pt,
    breakable
}

\usetikzlibrary{arrows.meta,positioning,calc}

\theoremstyle{plain}
\newtheorem{theorem}{Theorem}[section]
\newtheorem{proposition}[theorem]{Proposition}
\newtheorem{lemma}[theorem]{Lemma}
\newtheorem{corollary}[theorem]{Corollary}
\theoremstyle{definition}

\newtheorem{assumption}[theorem]{Assumption}
\theoremstyle{remark}
\newtheorem{remark}[theorem]{Remark}

\providecommand{\cS}{\mathcal{S}}
\providecommand{\cV}{\mathcal{V}}

\providecommand{\cM}{\mathcal{M}}    
\providecommand{\cG}{\mathcal{G}}    
\providecommand{\Ex}{\mathbb{E}}
\providecommand{\R}{\mathbb{R}}
\providecommand{\softmax}{\mathrm{softmax}}
\renewcommand{\KL}[2]{D_{\mathrm{KL}}\!\left(#1 \,\Vert\, #2\right)}
\title{Learning Dynamics of On-policy Distillation}

\author{Guan Zhe Hong$^{1}$ \quad Sumeet R. Motwani$^{1}$ \quad  Liang Chen$^{2}$ \quad Yangchen Pan$^{3}$ \\
\textbf{Junchi Yu}$^{1}$ \quad \textbf{Jialin Yu}$^{1}$ \quad \textbf{Philip H.S. Torr}$^{1}$ \quad \textbf{Yi Ren}$^{1}$\\
$^1$University of Oxford \quad $^2$EPFL \quad $^3$City University of Hong Kong
}

\iclrfinalcopy

\begin{document}

\maketitle

\begin{abstract}
On-policy distillation (OPD) is emerging as an important paradigm for language-model post-training, combining dense teacher supervision with training on student-generated trajectories. This creates a feedback loop where each update changes the student and thereby the training distribution that generates subsequent updates. Yet, how this feedback shapes learning and generalization remains poorly understood.
We conduct a mechanistic analysis of OPD's learning dynamics, distinguishing the two feedback loop stages: conversion of the distillation signal into parameter updates, and broadcast through which these updates change predictions across prefixes and reshape the rollout distribution.
This framework helps us understand both limited student responsiveness despite substantial teacher-student disagreement, and degeneration when updates that improve local teacher agreement also increase the probability of unsuccessful continuations, exposing the student to unreliable supervision in subsequent rollouts.
Finally, we show that offline distillation can prepare a student for OPD without significantly improving immediate held-out accuracy. While OPD alone may degrade generation, it can yield substantial accuracy gains after sufficient offline preparation. Together, our results connect teacher supervision, student geometry, and the evolving rollout distribution in determining the learning dynamics of OPD and its effects on downstream performance.
\end{abstract}

% ======================================================================
% Focused motivation/framing update based on the user's latest
% Learning_Dynamics_of_On_policy_Distillation__Copy_ (2).zip.
% The contribution structure and the user's substantive edits are retained.
% Preparation's geometric/source interpretation is evidence-informed, not
% presented as a separately isolated causal effect on both components.
% ======================================================================

\section{Introduction}
\label{sec:intro}
On-policy distillation (OPD) is becoming an important component of language-model post-training \citep{tinker_opd_2025,opd_survey_2026}. By matching a teacher’s next-token distributions at student-generated prefixes, it combines dense supervision with training on the states the student actually visits \citep{gu2023minillm,agarwal2024onpolicydistillation}. Yet, how this local supervision becomes useful learning over an evolving training distribution remains insufficiently understood. It is therefore important to theoretically understand both the conditions under which OPD improves generalization and its role alongside offline distillation.

The difficulty is that each OPD update changes both the student’s current behavior and the prefixes on which it will subsequently learn. Reducing teacher-student divergence on currently supervised predictions does not necessarily produce parameters from which further updates are useful \citep{rethinking_opd_2026, armandpour2026unmaskingonpolicydistillationhelps}. Modelling this requires a view of how supervision changes behavior beyond the current batch and how that impacts learning at later stages. We conduct a mechanistic analysis, aided by a first-principles account, of OPD learning dynamics to study this process, explain existing empirical challenges, and justify their empirical diagnostics and fixes.

We first characterize how the teacher-student log-probability difference can lead to meaningful weight updates, which depends on the overlap between the distillation signal and the logit-space directions the student’s current parameters can express. We then describe how learning at one prefix affects predictions at future ones through their dependence on shared parameter directions. We formalize this broadcast relationship through the student’s tangent kernel between the two prefixes, building on \citet{ren2025learning, malladi2023kernelbasedviewlanguagemodel}. Finally, we connect each update to its effect on the outcomes of student-generated sequences. The effect of these updates in broader settings also depends on what the training data covers: matching the teacher on narrow tasks may transfer only the capabilities that those tasks exercise. Integrating these effects over training characterizes what OPD can reach from a given starting point, accounting for how much successive updates improve or degrade performance.

\begin{figure}[!t]
\centering
\includegraphics[width=1.0\linewidth]{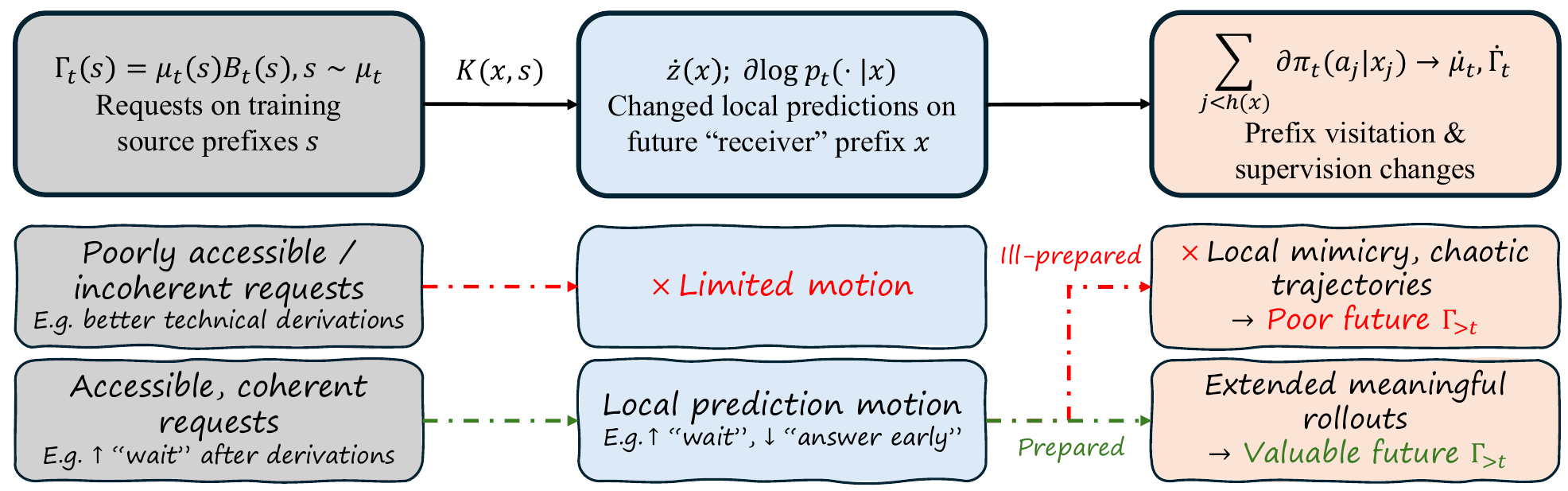}
\caption{OPD's learning dynamics, visualizing the feedback loop at the top (detailed in \Cref{sec:loop}), and illustrative successful and failing processes at the bottom in reasoning settings we analyzed empirically. In principle, OPD propagates current teacher requests $\Gamma_t(s)\vert_{s\sim\mu_t}$ to motion at any future prefix $x$ through shared parameters, captured by the Neural Tangent Kernel $K(x,s)$. Such policy motion shapes how the autoregressive rollout (prefix visitation) and supervision distribution changes ($\dot{\mu}_t, \dot{\Gamma}_t$). In practice, a healthy dynamics needs not only locally ``tangent-accessible'' and coherent requests (\Cref{sec:conversion}), but also the student to be sufficiently ``prepared'', so that it can sustain meaningful rollouts with learning value as it absorbs corrections requested by the teacher, instead of primarily performing local mimicry and falling into chaotic rollouts hard for the teacher to rescue, let alone improve on (\Cref{sec:broadcast,sec:preparation}).}
\label{fig:opd_illustration}
\end{figure}

This framework helps us understand two primary failure modes of OPD: stagnation under stronger teachers and degeneration into incoherent outputs \citep{rethinking_opd_2026,fu2026revisitingonpolicydistillationempirical,wang2026demystifying}. In the first case, despite substantial teacher–student log-probability differences, we show that the initial distillation signal can induce smaller parameter gradients as training proceeds, contributing to stalled improvement. In the second case, updates that reduce teacher–student disagreement at supervised prefixes also change predictions at other prefixes through shared parameters. These changes can shift the rollout distribution toward continuations the student cannot execute reliably, making those unstable continuations the source of subsequent supervision, making the teacher struggle to rescue, let alone improve the student.
%A well-prepared student, in contrast, can sustain useful learning as its rollout distribution changes. 
We illustrate the dynamical process of OPD and its possible failure modes in \cref{fig:opd_illustration} (we will elaborate on the intuitions behind this further in \cref{sec:loop}).

The dependence of OPD on the student sustaining stable rollouts (prefixes) which induce meaningful supervision motivates offline preparation. Here, the weak student is first forced to visit the teacher's prefixes and learn predictions there, enabling the subsequent OPD process to sustain productive learning as the student exposes finer correction opportunities to the teacher, while maintaining stabler and more recoverable changes in its rollout distribution. Across Qwen-2.5 and Llama-3.1 models on AIME 2024 and 2025 problems, we find that under sufficient offline training, the subsequent OPD phase substantially improve pass@$k$ (for $k$ up to 32) and per-problem reliability compared to offline training alone. In the rest of the paper, we present these results and insights organized under the following three questions. What determines the \emph{conversion} of teacher supervision into policy motion (\Cref{sec:conversion})? Does the \emph{broadcast} of these updates across prefixes improve subsequent reasoning or steer the student toward unreliable continuations and supervision (\Cref{sec:broadcast})? Can offline \emph{preparation} strengthen the student’s continuation ability and enable productive on-policy learning (\ref{sec:preparation})?

\textbf{Related works}. We briefly discuss related works here, and leave the expanded version to Appendix \cref{app:related-work}. \emph{On-policy distillation.} Training on learner-generated states with expert feedback is a classical approach to addressing distribution shift in imitation learning \citep{ross2011reduction}, applied to autoregressive distillation by \citet{lin2020autoregressiveknowledgedistillationimitation}. \citet{gu2023minillm} develop reverse-KL distillation using student rollouts, while \citet{agarwal2024onpolicydistillation} unify on-policy and off-policy training across multiple divergences. \citet{tinker_opd_2025} demonstrate its effectiveness for reasoning and personalization in language-model post-training. More recently, \citet{yang2026gopd} interpret OPD as dense KL-constrained reinforcement learning and introduce reward extrapolation. Empirical studies identify teacher-student compatibility on student prefixes as important factors in transfer and training failure \citep{rethinking_opd_2026,fu2026revisitingonpolicydistillationempirical}.

\emph{Learning dynamics of finetuning.} Kernel-based analyses connect parameter updates to changes in language-model predictions \citep{malladi2023kernelbasedviewlanguagemodel}. \citet{ren2025learning} decompose per-step prediction changes using the loss gradient and empirical tangent kernel, analyzing confidence dynamics in SFT and DPO \citep{rafailov2024directpreferenceoptimizationlanguage}. \citet{tomihari2026rl} study how RL updates propagate across examples, while \citet{shen2026geometryonpolicydistillation} identify persistent low-dimensional subspaces in OPD updates. Building on these perspectives, we connect the student’s evolving tangent kernel to its response to teacher supervision, the quality of its continuations, and the resulting distribution of future training prefixes. This distinguishes limited response to supervision from policy changes that worsen generation, and explains how either can shape subsequent learning.

\section{Setup}
\label{sec:setup}

On-policy distillation (OPD) trains a student on its own samples. The student generates a continuation from a prompt. At every prefix of that continuation, the training loss pulls the student's next-token distribution toward the next-token distribution of a fixed teacher \citep{gu2023minillm,agarwal2024onpolicydistillation,tinker_opd_2025}. This section fixes the notation for the objects that recur in the paper, including the policies, the source distribution that a run trains on, the request field that the teacher induces on that source, and the evaluation targets that we score runs against.

\paragraph{Policies and rollouts.}
A prefix $s$ is a prompt followed by a partial continuation. Prefixes range over a space $\cS$, and tokens range over a finite vocabulary $\cV$. The student has parameters $\theta \in \R^d$, logits $z_\theta(s) \in \R^{|\cV|}$, and next-token distribution $p_\theta(a \mid s) = \softmax(z_\theta(s))_a$. The teacher $q(\cdot \mid s)$ is fixed throughout training. Training proceeds in steps $t = 0, 1, 2, \ldots$. Gradient flow, the continuous-time version of the training process, has the update rule $\dot\theta_t = u_t$. For discrete gradient steps this gives a first-order local approximation \cref{app:ftc}. Two distributions share the student's weights. The \emph{model policy} $p_t$ is the distribution that the loss acts on. The \emph{rollout policy} $\pi_t = \cG[p_t]$ is the distribution that generates the samples (the transform $\cG$ collects the temperature, end-of-sequence conventions, etc). We sample prompts $b\sim\beta$ and continue them under $\pi_t$. The step-$t$ \emph{rollout population} is $\Pi_t(\xi)=\beta(b)\prod_{j}\pi_t(a_j\mid s_j)$ for a trajectory $\xi=(b,a_0,a_1,\ldots)$, with $s_0=b$ and $s_{j+1}=s_j\oplus a_j$ up to termination or the cap. The \emph{training source distribution} $\mu_t = \cM(\Pi_t)$ is the distribution over prefixes that the model trains on ($\cM$ normalizes over the rollouts to form the probability distribution).

\paragraph{Loss, request field, and update.}
\label{def:features}
For reverse Kullback--Leibler divergence, the per-prefix loss is $c_t(s)=\KL{p_t(\cdot\mid s)}{q(\cdot\mid s)}$. Write $\ell_t(s,a) = \log\big(p_t(a \mid s)/q(a \mid s)\big)$ for the log-ratio of the two policies at a token, and $\bar\ell_t(s) = \Ex_{a \sim p_t}[\ell_t(s,a)]$ for its mean under the student. The negative gradient of the loss with respect to the logits is
$B_t(s,a) \;=\; p_t(a \mid s)\,\big[\bar{\ell}_t(s) - \ell_t(s,a)\big]$.
We call $B_t$ the \emph{request field}\footnote{This is the ``student-teacher disagreement'' abstraction our theory is primarily concerned with: we adopt the specific reverse KL due to its popularity in OPD, but the theory covers general disagreement-measuring losses.}. It captures the student-teacher disagreement on a given prefix. The product $\Gamma_t = \mu_t B_t$, visitation times request, is the \emph{request population}. The student's \emph{tangent map} $\Phi_t(s) = \nabla_\theta z_{\theta_t}(s) \in \R^{|\cV| \times d}$. The map pulls a request in logit space back into parameter space, giving the per-prefix gradient $g_t(s) = \Phi_t(s)^{\!\top} B_t(s) = -\nabla_\theta c_t(s)$ and the update direction $u_t = \Ex_{s \sim \mu_t}[g_t(s)]$. See \cref{app:standing} for further details on the setup.

\section{The Learning Dynamics of the OPD loop}
\label{sec:theory}
\label{sec:calculus}\label{sec:anatomy}

\subsection{The OPD loop and the response law}
\label{sec:loop}\label{sec:conversion:request}

Recall that $\Pi_t$ denotes the distribution of student rollouts and $\mu_t=\cM(\Pi_t)$ the distribution of prefixes used by the loss. At a sampled prefix $s \sim \mu_t$, the student $p_t$ and the teacher $q$ supply the request $B_t(s)=-\nabla_{z_t(s)}c_t(s)$. The request population $\Gamma_t(s)=\mu_t(s)B_t(s)$ specifies \emph{what change} it requests on a prefix, across the rollout disitribution $\mu_t$. These requests then convert into parameter motion through the tangent map $\Phi_t(s)=\nabla_\theta z_t(s)$, which changes the future policy and rollouts. More explicitly, one training step follows
\begin{equation}
\resizebox{0.97\linewidth}{!}{$
\underbrace{\Pi_t}_{\text{rollouts}}
\xrightarrow{\;\cM\;}
\underbrace{\mu_t}_{\text{source}}
\xrightarrow{\;q\;}
\underbrace{\Gamma_t=\mu_tB_t}_{\text{weighted requests}}
\xrightarrow{\;\Phi\;}
\underbrace{\Delta\theta_t}_{\text{parameter step}}
\xrightarrow{\;\text{response}\;}
\underbrace{\Delta \log p_t(\cdot)}_{\text{policy changes}}
\longrightarrow
\underbrace{\Pi_{t+1}}_{\text{next rollouts}}
$}
\label{eq:loop}
\end{equation}
%The response $\delta^p_t(x,a)=\log p_{t+1}(a\mid x)-\log p_t(a\mid x)$ measures the change in output probability at a probing prefix $x$.
%Sampling with the changed policy regenerates the prefix (and rollout) distribution, $\mu_{t+1}=\cM(\Pi_{t+1})$.

\paragraph{Local requests shape future motion.}
Training imposes the gradient-based update \mbox{$u_t=\Ex_{\mu_t}[\Phi_t(s)^\top B_t(s)]$}.
For a gradient step $\Delta\theta_t=\eta u_t$, the response at \emph{any} receiver prefix $x$ is
\begin{equation}
\Delta p_t(a \mid x)
%\delta^p_t(x,a)
=
\eta\sum_s
\underbrace{\mu_t(s)}_{\text{source weight}}
\Big[
\underbrace{\mathsf A_t^p(x)}_{\text{normalization}}\;
\overbrace{\underbrace{K_t(x,s)}_{\text{student coupling}}\;
\underbrace{B_t(s)}_{\text{request}}}^{\Delta z_t(x)\text{: logit response}}
\Big]_a+O(\eta^2),
\label{eq:response-law}
\end{equation}
where $K_t(x,s)=\Phi_t(x)\Phi_t(s)^\top$ and $\mathsf A_t^p(x)=I-\mathbf1p_t(\cdot\mid x)^\top$.
The kernel $K_t$ measures how source and receiver logits move together through shared parameters; $\mathsf A_t^p$ converts logit motion into log-probability motion. Therefore, the model's prediction at any ``receiver'' prefix $x$ can change without appearing in the training batch.

In particular, since every prefix $x$ in a \emph{future} rollout (from the future policy) is a receiver, (1) OPD updates will affect \emph{local per-prefix predictions} of the student, captured by the logit motion $\dot{z}(x)$ and policy motion $\partial \log p(\cdot \mid x)$, (2) the updates will also shape \emph{which prefixes the student visits} (changed $\mu$) and expose to teacher supervision (changed $\Gamma$), as the policy motion along the ancestor tokens of any prefix $x$ accumulate to change its reach-probability. 

Let us look at the illustrative process in \cref{fig:opd_illustration}. Intuitively, imagine that the earlier requests locally promote the student to reason more and in specific manners (for instance, by requesting more ``wait''-based backtracking, suppressing early ``Answer:'' commitment, asking for a specific theorem, and so on). Due to the autoregressive nature of the policies, this will not only affect how likely the student emits specific tokens such as ``wait'' or ``answer:'' in a future rollout of the student, but will also alter the overall rollout distribution, by lengthening the reasoning trace, exposing new reasoning sub-trajectories, different errors, etc. In the \emph{benign} case, such online dynamics can be beneficial, as the teacher can correct the student in real time as if in a ``tutorial'' session. 

However, one can also imagine that such a dynamical process can be quite \emph{fragile} for a ill-prepared student. A weak student's tangent map $\Phi$ simply might not be very sensitive to the requests, resulting in \emph{limited motion} propagated from training to future prefixes via kernel $K$, thus limited changes in task accuracy. Moreover, even if there is substantial motion, such motion can be harmful. Taking our analogy further, an ill-prepared student which has not ``gone to class before the tutorial'' might benefit very little from online corrections on its own rollouts and even become more confused. That is, the ill-prepared student which has not been pretrained properly on the task and sustain sufficiently strong rollouts might only mimic superficial local patterns requested by the teacher, and \emph{push its rollout population $\mu$ towards chaos}, which can then be hard for the teacher to generate the right request population $\Gamma$ to even rescue, let alone improve.

The rest of our analysis builds on the OPD loop and the response law to make precise the above intuitions, complemented by empirical analysis of real OPD runs, and examine how offline-distillation preparation before OPD helps alleviate the problems for weak student models. \Cref{sec:local} (conversion) studies the student’s response to the supervised requests, which primarily concerns whether and how well a student model can convert different request populations $\Gamma$ (from different teachers) into policy motion $\dot{z}$ along the request fields. We develop early-training diagnostics of OPD runs from this perspective. \Cref{sec:broadcast} (broadcast) then focuses on \emph{where} policy motion $\dot{z}$ goes and how it influences future training sources $\mu$, and analyze how and why policy motion which improves local per-prefix readouts (such as training loss) might not make positive progress on terminal readouts such as answer accuracy. Motivated by the failure mechanisms, we then show in \Cref{sec:preparation}, across Qwen-2.5 and Llama-3.1 models, that with sufficient offline preparation, OPD runs improve pooled held-out accuracy (up to pass@32) and aggregate per-problem reliability on AIME 2024 and 2025 problems.

In the rest of this section, we primarily rely on the continuous gradient flow for cleaner theoretical discussions. Please see \cref{app:ftc} for further details on the finite step approximations. 

\subsection{Conversion: Teacher requests and the student's responsiveness}
\label{sec:local}\label{sec:motif-attenuated}\label{sec:response}\label{sec:modes}
\label{sec:conversion}\label{sec:conversion:geometry}

\leavevmode

\begin{wrapfigure}{r}{.44\textwidth}
\vspace{-9pt}
\centering
\includegraphics[width=\linewidth]{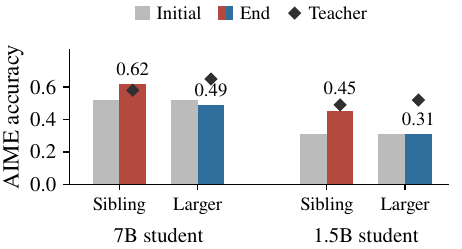}
\caption{Initial/end AIME accuracy; diamonds mark teachers. Red: sibling-taught; blue: larger-teacher taught.}
\label{fig:conversion-bench}\label{fig:hook}
\vspace{-8pt}
\end{wrapfigure}

\noindent\textbf{A strong teacher can transfer little.}\quad
We open with an intriguing empirical puzzle identified in the OPD community \citep{rethinking_opd_2026}: a larger (same-family) teacher can transfer less than an RL-trained sibling even when they are nearly matched in bechmark performance, and \citet{rethinking_opd_2026} suggests cancellation among token-level gradient signals as a possible explanation. We revisit this comparison at 1.5B and 7B (\cref{fig:conversion-bench}), holding the starting DeepSeek-R1-Distill-Qwen student and OPD recipe fixed within each scale. JustRL-1.5B and AceReason-Nemotron-7B yield substantial gains, whereas the larger DeepSeek-7B and DeepSeek-14B teachers transfer little throughout training despite higher benchmark scores (\cref{app:collective-protocol}). We call these runs \emph{healthy} and \emph{stalled}. We find that the teachers' request fields differ substantially in semantic functions, and that the student's responsiveness to these requests separates in two axes: coherence of the gradient signals, and kernel adaptation toward the requests. Along the way, we develop \emph{early-training diagnostics} that help reveal the ``request-conversion'' health of a training run.

\begin{figure}[!h]
\centering
\includegraphics[width=0.85\linewidth]{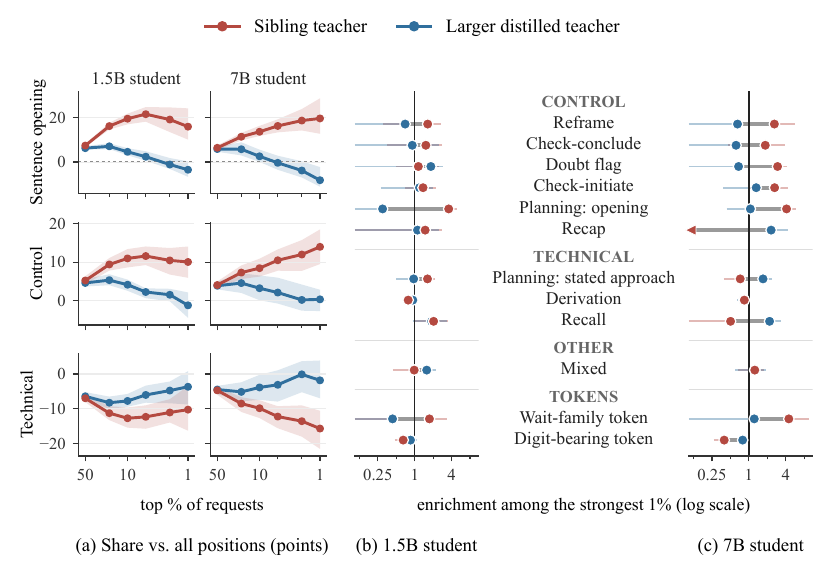}
\caption{\textbf{Teacher request fields differ systematically.}
In (a), each curve shows how much more (or less) often a semantic or token-position class appears among a teacher's top 50\%, 20\%, \ldots, 1\% of requests than among all annotated positions, in percentage points (so 0 = no preference). The RL-sibling teachers' increasingly emphasize requests on sentence openings and ``control-oriented'' moves, while the larger teachers' stay in sentence interiors and ``technical-oriented'' content. In (b) we show finer semantic categories of the request fields. Each dumbbell pairs the teachers' enrichment among their $1\%$ largest-normed of initial requests. $1$ denotes corpus-level prevalence. See \cref{app:collective-types} for finer details on the classes and how we conduct the interpretability analysis.}
\label{fig:conversion-tokens}
\end{figure}

\paragraph{How do the request fields differ?} As we see in the OPD update loop in \cref{sec:loop}, the teacher's request population $\Gamma_t$ at time 0 ``kickstart'' the whole learning process, from which the shared-parameter student model propagates into policy motion via its kernel $K_t$. Since the student starts out to be the same while the teachers are different across the healthy and stalled runs, their request fields become the natural entry point for our analysis.

On fixed initial-student rollouts, \cref{fig:conversion-tokens} compares where each teacher's largest-normed requests occur. (a) shows the coarse tendencies of the semantics of the requests as we move up the request-norm percentile, while (b) and (c) show the finer semantic categories and contrast the teachers' distinct request fields. We find that the RL-sibling teachers increasingly emphasize reasoning-management and initiating positions, while the larger teachers place relatively more of their strongest requests at technically oriented (e.g. digit-bearing, recalling mathematical fact) positions. This separation connects to findings in the reasoning-interpretability community, that RL tends to improve how well the model ``orchestrates'' the base reasoning mechanisms (including epistemic control abilities) \citep{venhoff2025base}. Full setup and analysis results are in \cref{app:collective-types}.

\paragraph{From requests to motion.}
Figure \ref{fig:conversion-tokens} shows that the healthy and stalled teachers provide systematically different request fields in the large-normed stratum, but this alone does not help us understand which teacher's request field converts better into policy motion along along the request field. 
Recall that for a prefix population $\nu$, the gradient is $g_t(s)=\Phi_t(s)^\top B_t(s)$ in terms of the tangent map and the request, $u_t=\Ex_\nu g_t(s)$ the average gradient, and $P_t=\Ex_\nu\|B_t(s)\|^2$ the request norm, or ``pressure'' (recall $B$ is the request field induced by the teacher-student disagreement).
Under gradient flow $\dot\theta_t=u_t$, the \emph{coefficient of the logit response $\dot{z}_t$ along the request field $B_t$ on $\nu$} can be written as
\begin{equation}
\Xi_t[B_t]
=\frac{\langle B_t,\dot z_t\rangle_\nu}{P_t}
=\frac{\|u_t\|^2}{P_t},
\qquad
\dot z_t=\Xi_t[B_t]B_t+r_t,\quad
\langle B_t,r_t\rangle_\nu=0,
\label{eq:collective-response}
\end{equation}
where we define the distribution-level inner product $\langle F,G\rangle_\nu=\Ex_\nu[F^\top G]$ and $P_t>0$. With nonzero denominators, this \emph{collective conversion gain} factors as

\begin{equation}
\Xi[B]=
\underbrace{\frac{\Ex_{s\sim\nu}\|\Phi(s)^\top B(s)\|^2}{P}}_{\text{accessibility }A}
\underbrace{\frac{\|\Ex_{s\sim\nu} g(s)\|^2}{\Ex_{s\sim\nu}\|g(s)\|^2}}_{\text{coherence }C},
\qquad \|u\|^2=P\Xi=PAC.
\label{eq:pac}
\end{equation}
\emph{Accessibility} averages the self-kernel Rayleigh coefficients $B(s)^\top K(s,s)B(s)/\|B(s)\|^2$, weighted by request energy: it measures how \emph{responsive} the student's tangent map $\Phi(s)$ is to the requests $B(s)$. Coherence measures how well the resulting parameter updates $g(s)=\Phi(s)^\top B(s)$ combine: aligned updates reinforce one another, while cancellation lowers $C$. Thus ``conversion'' of requests into parameter-update-induced policy motion not only depend on pressure $\Ex_\nu\|B_t(s)\|^2$, but even if two runs have similar gradient norm with high pressure, the requests need not convert efficiently due to low accessibility (poor parameter-level motion per unit request norm at a prefix) or coherence (poor cross-prefix aggregation of parameter motion). The deeper geometric interpretations and derivations are in \cref{app:collective-proof}. We now examine what the actual training runs reveal.

\begin{figure}[!h]
\centering
\begin{subfigure}[t]{.35\linewidth}
\includegraphics[width=\linewidth]{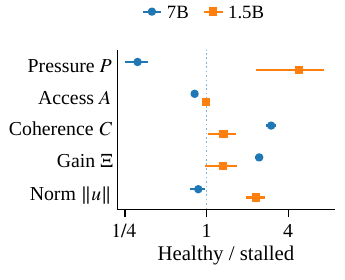}
\caption{Initial conversion.}
\label{fig:conversion-initial}
\end{subfigure}\hfill
\begin{subfigure}[t]{.32\linewidth}
\includegraphics[width=\linewidth]{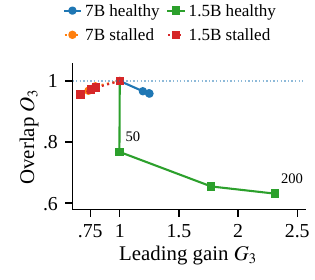}
\caption{Amplification and reorientation.}
\label{fig:conversion-geometry}
\end{subfigure}\hfill
\begin{subfigure}[t]{.30\linewidth}
\includegraphics[width=\linewidth]{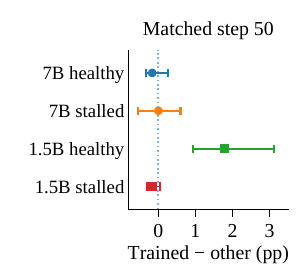}
\caption{Teacher-relative reorientation.}
\label{fig:conversion-turn}
\end{subfigure}
\caption{\textbf{Collective conversion and acquired responsiveness distinguish the pairings.} (a) Healthy/stalled ratios; segments show the range across the \textsc{control} and \textsc{technical} strata. (b) Median leading-three eigenvalue gain $G_3$ and eigen-subspace overlap $O_3$. (c) Step-50 paired trained-minus-other-field projection gain. Definitions and full results are in \cref{app:collective-geometry,app:collective-specificity}.}
\label{fig:collective-dynamics}\label{fig:absorb}
\end{figure}

\paragraph{Coherence contrasts at 7B.}
At 7B, healthy requests have near equal accessibility as the stalled requests but $2.75$--$3.24\times$ higher coherence, yielding $2.27$--$2.62\times$ higher collective gain $\Xi$ across the two sampled high-strength strata (\cref{fig:conversion-initial}). Therefore, even though the gradient norms across the healthy and stalled runs are similar, we see that it is the greater pressure in the stalled run which masks the weaker coherence of the update forces for the stalled run. For this 7B case, the gradient-incoherence conjecture in \citet{rethinking_opd_2026} is corroborated.

\paragraph{Evolving accessibility contrasts at 1.5B.}
At 1.5B the mechanism is richer: \emph{initial} accessibility is similar and coherence differs less; while we do see greater pressure for the healthy run (roughly $2$--$2.7\times$, which yields larger gradient norm), conversion of the requests at time 0 contrasts less across the two runs. This motivates examining how responsiveness \emph{evolves} as the student learns.

First, we note that live accessibility changes with \emph{both} the request $B_t$ and the kernel $K_t$, for instance, in a healthy run, $B_t$ can deplete learnable components and rotate away from the student's tangent map. To isolate kernel motion and how accessibility changes, we freeze each teacher's initial field $B_0$, prefix bank, and token coordinates, then reapply the same request field at every checkpoint\footnote{Time-$0$ request field is the least confounded as the initial student is the same in both the healthy and stalled runs. In practice, any representative probe bank of prefixes can be used to examine, for instance, the conversion health of more specific reasoning behaviors.}. The common initialization anchors both fields to the same student, while preserving their distinct request profiles we saw earlier with the different teachers. With $b_0(s)=B_0(s)/\|B_0(s)\|$,
$a_t(s)=b_0(s)^\top K_t(s,s)b_0(s)
       =\sum_j\lambda_{j,t}(s)|v_{j,t}(s)^\top b_0(s)|^2.
$
Here $(\lambda_{j,t},v_{j,t})$ are eigenpairs (eigenvalues and eigenvectors) of the self-kernel $K_t(s,s)$, and the request-energy-weighted average of $a_t(s)$ is $A_t[B_0]$. Larger eigenvalues amplify accessible directions; turning strong eigendirections toward $B_0$ makes more of the same request visible. Both can increase local responsiveness (\cref{app:collective-frozen}).

At 1.5B in the healthy run, leading-subspace overlap falls to $0.77$ by step 50 while leading strength remains near its initial value (\cref{fig:conversion-geometry}). This means its eigenframe \emph{re-orients towards the request field} in the early phase of training with nearly no scaling-amplification in the leading eigenvectors: the fixed request gains a median $5.5$ percentage points of top-three projection, and its paired gain exceeds the other teacher's by $1.79$ points (\cref{fig:conversion-turn}). 
By step 200, leading eigenvalue strength reaches $2.31\times$ and median request-direction sensitivity $2.08\times$ the baseline. In contrast, the stalled run \emph{contracts} in leading eigenvalue strengths and sensitivity ($0.67\times$ and $0.83\times$ respectively). 
At 7B in the healthy run, eigenvector rotation is much less pronounced and adaptation is primarily amplification of the existing eigenframe: leading strength reaches $1.25\times$ with overlap $0.96$ against the initial frame, while stalled leading strength falls to $0.74\times$.

These results distinguish healthy and stalled pairings \emph{early on} in an OPD training run in the requests they receive and the responsiveness their tangent geometry develop, giving diagnostics beyond the standard gradient norm and training loss from the request-conversion aspect. The next sub-section will study what comes next: when there is significant policy motion, cumulated over time, where do they head to? Why do two student models OPD-trained under the same teacher, both reducing training loss substantially, have different generalization performance under OPD?

\subsection{Broadcast: why local agreement need not improve reasoning}
\label{sec:broadcast}\label{sec:trajectory}

Under JustRL-1.5B, both DeepSeek-R1-Distill-Qwen-1.5B and Qwen2.5-1.5B-Instruct substantially reduce their training loss. Yet DeepSeek gains accuracy (\cref{sec:local}), while the base Qwen 1.5B model declines substantially on MATH within 100 steps. Why does learning from the same teacher help one learner and harm the other?

\paragraph{How do OPD updates change a per-prefix readout function?}
Let us first examine how an OPD update influences a local, per-prefix readout function of a rollout, such as the training loss or entropy. It acts on two coupled axes: it changes \emph{which prefixes} are encountered in future rollouts from the model, and it changes the model's \emph{predictions at a fixed prefix}. 

More precisely, let us define a local \emph{prefix readout} $Y_t(x)$. Define $J_Y(t)=\Ex_{\nu_t}Y_t(x)$, where $\nu_t$ is the observed prefix population. Then its rate of change (due to the parameter update) decomposes into ``routing'' and ``drift'':
\begin{equation}
 \dot J_Y(t)=
 \underbrace{\sum_x\dot\nu_t(x)Y_t(x)}_{R_Y(t):\ \text{routing}}
 +\underbrace{\Ex_{\nu_t}\dot Y_t(x)}_{D_Y(t):\ \text{drift}},
 \qquad J_Y(b)-J_Y(a)=\int_a^b(R_Y+D_Y)\,dt.
 \label{eq:rd-law}
\end{equation}
Drift changes the local, per-prefix readout over the current training population, while routing changes the probability weight assigned to the current prefixes, that is, which prefixes get exposed to teacher supervision in future updates.

At the more fundamental level of OPD's dynamics, both of these channels are projections of the $K$-$\Gamma$ broadcasts. The distinction is \emph{where the same update broadcasts}. Recall from the response law that the weighted request population $\Gamma_t$ propagate through $K_t$ to induce a logit response $v_t(x)$ at every receiver prefix $x$. With a temperature-$\tau$ sampler, receiver $x$ and prediction token $a$, this gives token log-probability velocity
\begin{equation}
 \zeta_t(x,a)=\partial_t\log\pi_t(a\mid x)
 =\frac{(e_a-\pi_t(\cdot\mid x))^\top v_t(x)}{\tau},
 \qquad v_t(x)=\sum_sK_t(x,s)\Gamma_t(s).
 \label{eq:rd-velocity}
\end{equation}
At the observed prefix $x$ itself, $v_t(x)$ changes its prediction, giving \emph{drift} $D_Y=\Ex_{\nu_t}[\nabla_zY_t(x)^\top v_t(x)]$. To reach that prefix, however, the autoregressive model must first generate its ancestor tokens, which also are the receivers of the $K$-$\Gamma$ broadcast. Summing the log-probability velocity of the ancestor tokens gives the velocity of reach-probability of prefix $x$, $\lambda_t(x)=\sum_{j<h(x)}\zeta_t(x_j,a_j)$. This gives \emph{routing} $R_Y=\operatorname{Cov}_{\nu_t}(Y_t,\lambda_t)$. 
Furthermore, the prefix visitation change causes the future training prefix distribution $\mu$ to change: within any time interval $[a,b]$,
\begin{equation}
 \mu_b(s)\propto\mu_a(s)e^{\Lambda_{a:b}(s)},\qquad
 \Lambda_{a:b}(s)=\int_a^b\lambda_t(s)\,dt\;\;\rightarrow \;\Gamma_b=\mu_bB_b.
 \label{eq:broadcast-source-renewal}
\end{equation}
The new rollouts (i.e. visited prefixes) supply the next requests and the induced $\Gamma$ field, on which the new round of updates and broadcasts are made. 

Intuitively, significant prefix-visitation change can be \emph{either} beneficial or damaging. For a weak student, it can route into low-quality, chaotic rollout distributions on which the teacher struggle to give requests that can help recover (or raise) the student's intermediate reasoning performance and terminal scores. On the flip side, the continually refreshed training rollouts also allow the possibility of ``online corrections'', where problematic reasoning behaviors broadcasted from early training can be corrected by the teacher mid-training. We will see these effects in our empirical analysis later.

To empirically study the effects of routing and drift in influencing the local readouts, we estimate their effects by \emph{cross-scoring training rollout banks and learners} at time intervals. Write $M_{ij}=\Ex_{x\sim\nu_i}Y_j(x)$ for the average readout of learner $j$ on the frozen bank of rollouts extracted at time step $i$, with $i,j\in\{a,b\}$. Replacing the bank at a fixed learner checkpoint measures routing; replacing the checkpoint on a fixed bank measures drift. Averaging the two replacement orders gives
\begin{equation}
 \widehat R_{a:b}=\tfrac12[(M_{ba}-M_{aa})+(M_{bb}-M_{ab})],\qquad
 \widehat D_{a:b}=\tfrac12[(M_{ab}-M_{aa})+(M_{bb}-M_{ba})].
 \label{eq:rd-estimators}
\end{equation}
These contributions sum exactly to $M_{bb}-M_{aa}$. Our curves accumulate them over consecutive windows; its relation to the path integrals and further details are in \cref{app:rd-theory,app:rd-protocol}\footnote{While a step-wise estimate for routing and drift is theoretically recommended, it is too expensive for our setting, and we resort to intervals of 20 update steps.}.

\paragraph{From local readouts to rollout-outcome readouts.}
So far we have discussed how the $K$-$\Gamma$ broadcasts change the model's predictions per prefix, how they accumulate along ancestor tokens to change prefix visitation, and how they influence a local per-prefix readout function. \emph{Correctness}, by contrast, is \emph{non-local}, and known only \emph{after completing the full rollout}, so its $J_Y$ expression must consider how ``aligned'' the policy motion is with terminal score when cumulated along a rollout trajectory. 

Generally, given a trajectory-determined terminal readout function $Y$, set $J_Y=\Ex_{\Pi_t}Y$ and use the student's continuation values
$Q_t^Y(x,a)=\Ex_t[Y\mid x,a]$ and $V_t^Y(x)=\Ex_{a\sim\pi_t}Q_t^Y(x,a)$.
We obtain:
\begin{equation}
 \begin{aligned}
 J_Y(T)-J_Y(t_\star)
 &=\int_{t_\star}^{T}\sum_xd_t(x)\Ex_{a\sim\pi_t}\!\left[\zeta_t(x,a)\underbrace{(Q_t^Y(x,a)-V_t^Y(x))}_{\text{continuation advantage}}\right]dt\\[-1pt]
 &=\frac1\tau\int_{t_\star}^{T}\sum_{x,s}d_t(x)
 [\pi_t\odot(Q_t^Y-V_t^Y)](x)^\top K_t(x,s)\Gamma_t(s)\,dt,
 \end{aligned}
 \label{eq:broadcast-continuations}
\end{equation}
where $d_t(x)$ is the probability of reaching $x$. This weights motion by what the \emph{student} can achieve. When the favored alternatives have similarly unsuccessful continuations, fitting their next-token distributions can yield little accuracy progress overall (see \cref{app:outcome-proof} for the proof).

\begin{figure}[!t]
\centering
\captionsetup[subfigure]{font=footnotesize,skip=2pt,justification=centering,singlelinecheck=false}
\begin{subfigure}[t]{.243636\linewidth}
\centering
\includegraphics[width=\linewidth]{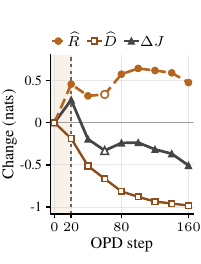}
\caption{Loss change}
\label{fig:rd-contrast-loss}
\end{subfigure}\hfill
\begin{subfigure}[t]{.243636\linewidth}
\centering
\includegraphics[width=\linewidth]{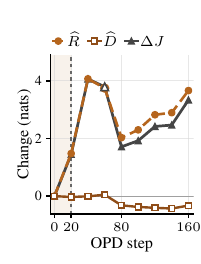}
\caption{Entropy change}
\label{fig:rd-contrast-entropy}
\end{subfigure}\hfill
\begin{subfigure}[t]{.243636\linewidth}
\centering
\includegraphics[width=\linewidth]{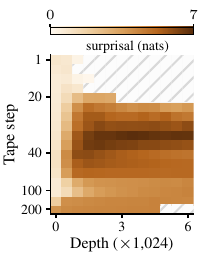}
\caption{Student uncertainty}
\label{fig:rd-contrast-student}
\end{subfigure}\hfill
\begin{subfigure}[t]{.243636\linewidth}
\centering
\includegraphics[width=\linewidth]{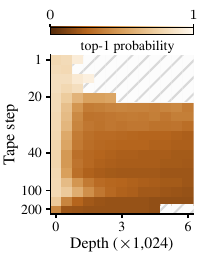}
\caption{Teacher confidence}
\label{fig:rd-contrast-teacher}
\end{subfigure}
\caption{\textbf{Routing into unreliable prefixes, then fitting them.} Weak Qwen: (a,b) cumulative routing, drift, and total changes in local loss and entropy, each on its own scale. The vertical guide at step 20 marks the approximate onset-to-fitting transition; the effects overlap. (c,d) Student surprisal and teacher top-1 probability; darker means greater uncertainty. Columns are 512-token windows. Details are in \cref{app:rd-protocol}.}
\label{fig:rd-contrast}
\end{figure}

\paragraph{Weak 1.5B Qwen: routing into unreliable reasoning, then fitting uncertain supervision.}
In the early phase of training, the teacher pushes the weak learner toward longer rollouts (such as heavy negative requests on the student's ``overly early'' answer commitments), but these extensions become high-entropy and incoherent. The teacher supervision there also becomes much less concentrated than in the shallower rollout windows (\cref{fig:rd-contrast}c,d). As seen in \cref{fig:rd-contrast}(a) and (b), from step 1 to 20, loss rises $0.58\to0.85$ (routing $+0.455$ outweighs drift $-0.187$); entropy also rises mainly through routing. Intuitively, the student, which has never been trained to emit long reasoning rollouts, gets pushed into reasoning for long before committing an answer, and executing reasoning states which it is highly unfamiliar with. The teacher, conditioned on these low-quality states, also give much less confident guidance. 
The drift of the loss is small in this phase, however: the student has not learned the flatter prediction patterns of the teacher (uniformly) across the visited prefixes yet. 

What happens after the initial routing-dominated phase is dominated by drift (local fitting) on the unreliable request distributions. 
As seen in \cref{fig:rd-contrast}, entropy remains high within the band around $2-4$ nats, and loss reduction is increasingly driven by drift, that is, the student is indeed fitting the teacher's low confidence predictions on the chaotic, low-quality prefixes continually generated during training. To further test this, we find that, on the same batch-41 training prefixes, loss falls $0.79\to0.06$ while entropy falls $4.99\to3.99$: drift over the course of training actually broadcasts and sharpens predictions on this rollout bank toward an uncertain but slightly less entropic teacher (\cref{app:rd-entropy-response,app:rd-entropy-evidence}).

\paragraph{Policy motion and the shifting value of continuations.} So far we see that the weak Qwen model's failed run has the OPD update loop spending later, more ``stable'' drift-dominated updates learning from unreliable requests on the low-quality generative distribution. This helps us better understand how the training loss declines. Intuitively, fitting unreliable teacher supervision on high-entropy rollouts already suggests that this will not yield a better-reasoning student model. To understand how and why this process worsens terminal score more precisely, we need to turn to \Cref{eq:broadcast-continuations}: policy motion influences \emph{non-local} readouts (including terminal score) through the \emph{value / advantage} of the student's continuations (namely the $Q - V$ advantage summed across the prefixes), which can be in conflict with \emph{local} teacher agreement. 

While it is prohibitively expensive to directly compute the ``velocity$\times$advantage'' term in \cref{eq:broadcast-continuations}, we find on select banks of probing rollouts and continuations that as training proceeds, the model increasingly make local decisions whose continuation correctness decline, and complementarily, visit prefixes that are difficult to reach the right answer at increasingly shallower depth. This tells us that the per-token policy velocity of the model tend to align with ``negative advantage'' over the course of training, resulting in the overall decline of terminal accuracy. We elaborate on these studies below.

First, to probe the continuation-value term in \Cref{eq:broadcast-continuations}, we freeze a set of answer-decision prefixes from the initial student's held-out MATH rollouts and intervene on the local ``answer-versus-continue'' decision token position. The student's probability of the answer branch falls from $0.74\to0.19\to0.07$ at steps 0, 20, 100, below the teacher's $0.30$ on the same prefixes. At update 20, while ``routing'' is dominating, counterfactual continuations from the two branches have similar downstream value (correct $0.30$ for the ``answer'' branch and $0.29$ for the ``continue'' branch), i.e. the student still maintains reasonable answer-reaching abilities. By update 100, however, the continue-branch rarely produces the answer ($8\%$ of continuations). The teacher's own continuations from the same prefixes is correct $0.60$ of the time, which is much higher, so the gap is in the student's degraded execution of the action branch, not in the branch itself. For a complete treatment of how we decompose and estimate \cref{eq:broadcast-continuations} see \cref{app:steer-probe}.

\begin{wrapfigure}{r}{.32\textwidth}
\vspace{-12pt}
\centering
\includegraphics[width=\linewidth]{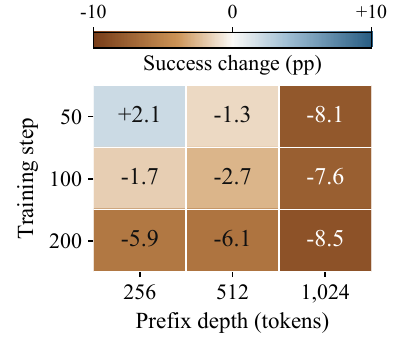}
\caption{\textbf{Initial-policy continuation success rate on trained prefixes.} Negative values indicate harder prefixes for reaching the right answer.}
\label{fig:rd-initial-continuation}
\vspace{-8pt}
\end{wrapfigure}

Now, these probes fix the prefixes and probe how the student checkpoints continues them. As we discussed before, the broadcast mechanism changes both prefix \emph{visitation} and continuations from them, therefore we next ask the complementary question of \emph{whether the prefixes visited during OPD also become increasingly harder to complete}. We fix a few rollout depths and student checkpoints, sample prefixes generated from these checkpoints at those depths, and then let the initial student (as a reference policy) to finish. Keeping the continuer \emph{fixed} makes changes in success reflect changes in prefix visitation. To understand the principled connections between these reference-continuer effects and \cref{eq:broadcast-continuations}, please refer to \cref{app:continuation-profile}. Empirically, \Cref{fig:rd-initial-continuation} shows a clear increase in continuation difficulty due to the prefix visitation shift, on the axes of rollout depths and training length. At step 50, while the shallow prefixes remains usable (even slightly better than the initial policy's own prefixes), it already yields less successful continuations on deep prefixes. As training goes on, the decline show up at increasingly shallower cuts of the rollout.

\begin{wrapfigure}{r}{.32\textwidth}
\vspace{-5.2em}
\centering
\includegraphics[width=\linewidth]{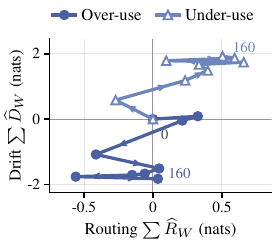}
\caption{\textbf{Conditional checking refinement.} Filled circles: initial over-use; open triangles: initial under-use.}
\label{fig:rd-learning}
\vspace{-17pt}
\end{wrapfigure}

\paragraph{DeepSeek: stable learning acquiring epistemic behaviors.}

DeepSeek supplies the clean contrast. Loss falls $0.263\to0.015$, mainly through drift ($-0.214$) rather than routing ($-0.033$), indicating more more stable prefix visitations and local learning of requests. Furthermore, motivated by the epistemic-control request distribution found in \cref{sec:local}, we design a local readout to examine the dynamics of how the model acquires intermediate reasoning behaviors, focusing on ``wait''-family initiation at sentence starts (observed to initiate important backtracking and checking behaviors in R1-styled models \citep{marjanovic2026deepseekr}). On each prefix, let $w_t$ and $q_W$ denote student and teacher WAIT-family probabilities. We freeze two classes: \emph{over-use} ($w_0>q_W$) and \emph{under-use} ($w_0<q_W$), and set the local readout to be $\log[w_t/(1-w_t)]$ (log-odds of ``wait'' initiation).

\Cref{fig:rd-learning} plots cumulative routing and drift of the ``wait''-readout. Filled circles move downward: drift suppresses initial over-use. Open triangles move upward: drift promotes initial under-use. The corresponding net changes are $-1.82$ and $+1.92$ nats. While there are short overshoots of routing-dominated changes (early false suppression of under-use positions and promotion of over-use positions), they are quickly stabilized. Moreover, teacher-relative disagreement also decreases on frozen and held-out rollout banks (\cref{app:rd-checking}). Thus, unlike the failed learner's case where drift-dominated reduction of training loss happened on unreliable prefixes, DeepSeek's stabler drift-dominated reduction on training loss not only improves on its terminal score, but is accompanied by improvements in intermediate reasoning behaviors, with it sustaining prefix populations that allows to learn \emph{when} to initiate a ``wait''-based check as a representative example.

\paragraph{Preparation permits a different trajectory.}
The base Qwen model, prepared with offline distillation training on teacher rollouts, also undergoes substantial prefix visitation shifts, but sustains useful learning. Unlike DeepSeek's smoother refinement, its loss and checking behavior are nonmonotone: a routing-dominated overshoot in loss and reasoning (epistemic) behaviors are followed by lower-entropy rollouts and conditional moderation of reasoning behaviors dominated by drift. Its held-out gains suggest that a stronger starting policy with higher prefix quality and continuation-readiness makes these excursions amenable to correction. Full analysis in \cref{app:rd-prepared}.

\subsection{Preparation: improving the starting point of on-policy learning}
\label{sec:source}
\label{sec:reachability}
\label{sec:preparation}

\begin{figure}[!h]
%\vspace{-1em}
\centering
\includegraphics[width=1.0\linewidth]{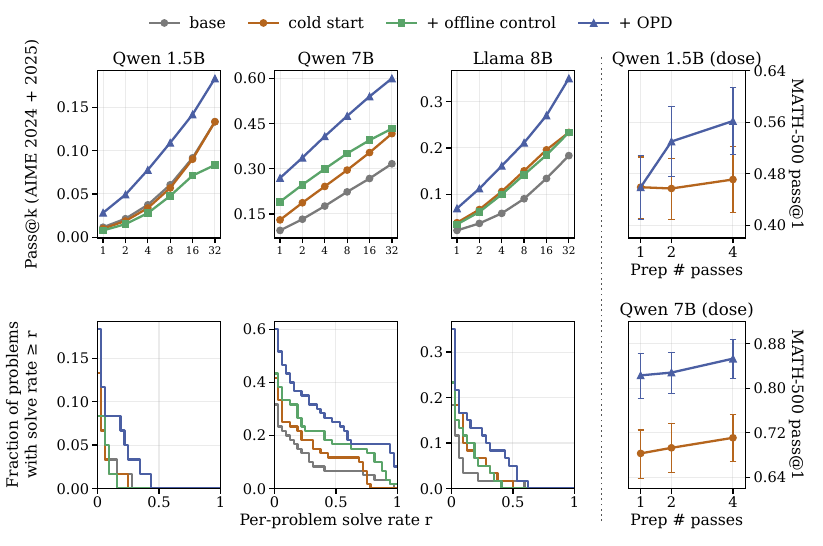}
\caption{\textbf{Offline-prepared OPD improves sampled coverage and reliability.} Left: pooled AIME 2024/2025 pass@$k$ for $k\le32$ (top) and the fraction of problems with empirical solve rate $\ge r>0$ (bottom), using 32 responses per problem from batch~a. Right: preparation dose versus strict MATH-500 pass@1, using the step-100 OPD checkpoint at 1.5B and the best evaluated checkpoint within 100 steps at 7B; error bars are 95\% problem-bootstrap intervals. See \cref{app:preparation-multisetup,app:preparation-legacy} for protocols and paired comparisons.}
\label{fig:preparation-aime}
\end{figure}

\Cref{sec:broadcast} suggests a direct intervention. As we saw in the last sub-section, OPD can fail because early policy changes route the weak student model into reasoning states it cannot yet execute reliably, and the training run spends a large chunk of time fitting unreliable requests on increasingly lower-quality prefixes. Then, a natural candidate intervention is that, before OPD, we force the student to visit higher-quality prefixes from the teacher itself and learn the teacher's stronger continuations on those prefix populations. This suggests offline distillation to prepare the student model's continuation abilities before running OPD, which can then expose finer student errors to the teacher and sustain more stable learning trajectory for online corrections when OPD changes prefix visitation, without the requests looking so foreign to the student in the OPD phase that it routes into irrecoverable rollout distributions.
%It may therefore improve continuation skills and reasoning-state management that become useful as OPD changes visitation. This motivates a cold start \citep{rethinking_opd_2026} as preparation for subsequent on-policy learning.

We prepare with student/teacher pairings Qwen2.5-1.5B-Instruct/JustRL-1.5B \citep{qwen2025qwen25technicalreport,he2025justrlscaling15bllm}, Qwen2.5-7B-Instruct/AceReason-Nemotron-7B \citep{liu2025acereasonnemotron11advancingmath}, and Llama-3.1-8B-Instruct/Llama-3.1-Nemotron-Nano-8B-v1 \citep{grattafiori2024llama3herdmodels,bercovich2025llamanemotronefficientreasoningmodels}, with students trained on their teachers' solutions to OpenMathInstruct-2 \citep{toshniwal2024openmathinstruct2}. We then compare OPD with continued offline distillation on the same DAPO-Math-17k prompt blocks \citep{yu2025dapo}, matching optimizer steps and sequence presentations.

\Cref{fig:preparation-aime} shows higher pooled pass@$k$ accuracy for OPD than offline training through $k=32$, using 32 responses per problem. The gains extend beyond pass@1 to sampled coverage and the distribution of per-problem solve rates. The \emph{reliability} improvement is especially interesting: looking at pass@$k$ alone might give us the impression that OPD just solves more problems by rare chances, however, we find that OPD with appropriate preparation also enhances the per-problem solve rates for the models beyond the offline-distillation-only alternatives. Additionally, the preparation ``dose study'' (\cref{fig:preparation-aime}, right) shows increasing value of offline-preparation for the later OPD phase. At 1.5B, one to four passes leaves the starting strict MATH pass@1 at $45.7$--$47.1\%$, but raises the 100-step OPD endpoint from $45.9\%$ to $56.2\%$. Similar trends are observed at 7B (\cref{app:preparation-legacy}). Please refer to \cref{app:preparation-multisetup} for further details.

\section{Conclusion}
\label{sec:conclusion}

\textbf{Findings.}
We studied on-policy distillation as a feedback loop between teacher supervision, student response, and the rollout distribution that supplies future supervision.
Our analysis separates two mechanisms in this loop: \emph{conversion}, and \emph{broadcast}. Across our settings, these mechanisms explain both stalled transfer and deterioration despite improving local teacher agreement. Offline preparation changes the starting point of this loop: even with little immediate accuracy improvement, it can enable
substantial gains from subsequent OPD.

\textbf{Limitations and discussion.}
Our empirical analysis is confined to mathematical reasoning and a select set of LLM models due to resource limitations. We also present a more detailed Q\&A in Appendix \ref{app:clarifications} to further clarify potential ambiguities in our work.  

\section*{AI Use Statement}
Generative AI tools were also used for language editing, rewriting and presentation. We used generative AI tools to assist with code implementation, debugging, experimental scripting, and parts of experiment execution.
The research questions, theoretical development, experimental design, interpretation, and scientific
conclusions were determined by the authors. All AI-assisted code and text were reviewed and validated by the authors. We take responsibility for the final content of this work, including text, claims, and artifacts produced with the aid of generative AI.

\section*{Reproducibility statement}

Precise experimental details are provided in the appendix, including model specifications, training pipeline and evaluation details.

\bibliography{iclr2027_conference,opd_appendix_references}
\bibliographystyle{iclr2027_conference}

\appendix
\clearpage
\section{Mechanistic interpretation and key questions}
\label{app:clarifications}

\noindent\textbf{Q1. What do the response law and decompositions in \cref{sec:loop,sec:local,sec:broadcast} explain beyond their familiar mathematical ingredients?}
\par\noindent
They provide a connected mechanistic account of how dense local supervision can produce either weak transfer or deteriorating generation. OPD updates the policy that generates its next training population. Understanding its dynamics therefore requires following supervision beyond the current loss, through its effects on future behavior and training histories (student's rollouts and the induced supervision signals / request field).

The theory separates three links in this process. First, \cref{sec:local} explains how a teacher's requests become parameter motion: accessibility determines their expression through the student's tangent map, while coherence determines how the resulting gradients combine. Teacher benchmark performance, magnitude of (KL-based) disagreement, or gradient norm alone cannot characterize this conversion. Second, \cref{sec:broadcast} distinguishes changing a prediction from changing the ancestral decisions that make its history likely. Shared-parameter broadcast can increase exposure to a continuation without making the student better at executing it; the resulting histories then change the supervision received from the same fixed teacher. Third, \cref{eq:broadcast-continuations} connects the delivered response to completed outcomes through what the \emph{student's remaining policy} can achieve, separating teacher matching from useful reasoning.

Together, these links explain the structure of the empirical puzzles. The stronger-teacher study exposes the importance of gradient coherence and tangent accessibility; the degenerate run exposes that, under substantial policy motion, the degree of freedom granted to the learner to generate its own rollouts (training histories) can make it generate the new rollouts (reasoning states) which it has difficulty managing and grows uncertain on, inducing poorer-quality supervision signals. Loss reduction dominated by local fitting on these poorly supervised rollouts cannot improve the learner's terminal scores. The mechanisms revealed in the degenerate run motivates preparing the student for the continuations that OPD will elicit, even when preparation does not substantially improve current accuracy by itself. The theoretical contribution is the integrated explanation that makes these mechanisms distinguishable, connects them to outcomes, and motivates an intervention that boosts the model's learning stability and generalization.

\Needspace{5\baselineskip}
\noindent\textbf{Q2. Why is every training run reported with a single seed?}

\textbf{Cost}. The runs behind this paper were trained and evaluated on GH200 nodes, and the single-seed set already consumed roughly 4{,}000 GPU-hours, which is at the limit of our computing resources. OPD is known to be computationally demanding, we therefore designed the experiments so that the main comparisons do not rest on seed variance. The mechanism measurements of Sections 3.2 and 3.3 are exact identities evaluated within a run, on fixed receiver banks and fixed checkpoints, and their uncertainty is reported at the level of problems or sites (paired bootstrap intervals). The outcome comparisons of Section 3.4 are paired within a student-teacher pair on the same problems and replicated across three pairs and two student scales. What single seeds leave unquantified is run-to-run variation of the training trajectories themselves.

\Needspace{5\baselineskip}
\noindent\textbf{Q3. Why does \cref{sec:broadcast} need both prefix-level routing/drift (\cref{eq:rd-law}) and trajectory-level continuation advantages (\cref{eq:broadcast-continuations}) to explain degeneration?}
\par\noindent
For loss or entropy, $J_Y=\mathbb{E}_{\nu_t}Y_t(x)$ describes predictions on the currently observed histories. Cross-scoring separates changes in those histories from changes in predictions on the \emph{same} histories. In \cref{fig:rd-contrast}, the initial student is already highly uncertain on the deep histories that later become prevalent, while subsequent fitting sharpens them. On identical batch-41 histories, discrepancy falls $0.79\to0.06$ and entropy $4.99\to3.99$. Together with the stopping/reach probes, these measurements distinguish routing followed by fitting from uniform entropy-increasing drift (\cref{app:rd-failure,app:rd-entropy-evidence}).

But fitting a prefix and improving the completion from it are different effects. For terminal correctness, $J_Y=\mathbb{E}_{\Pi_t}Y$, and the conditional value $V_t^Y(x)$ depends on \emph{every downstream decision}. At an intermediate cut of the autoregressive trace, its derivative separates changes in reaching the cut from changes in the remaining policy. Unrolling the latter gives the continuation-advantage weighting in \cref{eq:broadcast-continuations}; replacing it with a derivative of the logits at $x$ alone would omit downstream effects (\cref{app:outcome-proof}). Thus, a teacher-preferred next token need not help if the student cannot complete the reasoning it initiates. The prefix analysis establishes the observed routing-then-fitting pattern; the outcome analysis explains why local fitting need not restore correctness. Entropy enters the argument only as a prefix-level readout.

\Needspace{5\baselineskip}
\noindent\textbf{Q4. Could the responsiveness changes in \cref{sec:local} (Figure~\ref{fig:collective-dynamics}) simply reflect changing teacher--student disagreement or generic kernel movement?}
\par\noindent
A live accessibility measure $A_t[B_t]$ can change because either the requests or the kernel change. We therefore fix the initial numerical requests $B_0$, exact histories, and output coordinates across checkpoints. Since only the kernel varies in $A_t[B_0]$, its change measures responsiveness to the \emph{same supervision} (\cref{app:collective-frozen}).

To test whether eigenspace movement reflects teacher-specific adaptation, we evaluate both teachers' frozen fields in the \emph{same checkpoint kernel} and compare their projection gains, separating preferential reorientation from shared movement that benefits both fields. Healthy 1.5B exhibits preferential reorientation; healthy 7B mainly amplifies existing directions; stalled runs lose sensitivity to the fixed requests (\cref{app:collective-specificity}). These findings complement the initial accessibility/coherence decomposition, which exposes conversion differences concealed by comparable gradient norms across the runs.

\Needspace{5\baselineskip}
\noindent\textbf{Q5. What does preparation in \cref{sec:preparation,fig:preparation-aime} contribute beyond offline warm-up, and how does \cref{sec:broadcast} motivate it?}
\par\noindent
The contribution is an explanation of why the value of a warm-up stage, already used in prior work (\cref{app:related-work}), should be assessed through \emph{subsequent learning}. The failure mechanism identifies a mismatch between making extended reasoning likely and being able to execute it. Offline distillation supplies teacher-generated reasoning histories without requiring the initial student to generate them first. It can therefore prepare continuation skills and reasoning-state management that become useful as OPD changes visitation: \emph{continuation readiness}, a benefit beyond higher starting accuracy -- our pass@k results clearly demonstrate this.

\FloatBarrier
\clearpage
\section{Expanded related work}
\label{app:related-work}
\subsection{Core related works}
\noindent\textbf{Distillation and the OPD loop.}
Knowledge distillation transfers teacher outputs \citep{hinton2015distillingknowledgeneuralnetwork}, including teacher-generated sequences \citep{kim2016sequence}. DAgger, the seminal online imitation algorithm, addresses distribution shift through expert feedback on learner-visited states \citep{ross2011reduction}; \citet{lin2020autoregressiveknowledgedistillationimitation} develop this connection for autoregressive distillation. MiniLLM uses reverse KL and student-generated samples \citep{gu2023minillm},  \citep{agarwal2024onpolicydistillation,piskorz2026onpolicyoffpolicylearningsystematic} study the choice of divergence loss, and \citet{tinker_opd_2025} demonstrate practical OPD for reasoning and personalization. These methods motivate our question: how does dense supervision on student histories produce successful transfer, stagnation, or degeneration? We develop a systematic first-principles account of the coupled loop through which shared-parameter updates change predictions, renew visited histories, and reshape subsequent supervision.

\noindent\textbf{Kernel response laws.}
The neural tangent kernel connects parameter learning to function-space change \citep{jacot2018neural}. \citet{malladi2023kernelbasedviewlanguagemodel} analyze language-model finetuning through kernels, including an Adam extension; \citet{ren2025learning} decompose prediction changes through normalization, empirical-kernel, and loss-gradient terms for SFT and preference finetuning, including offline DPO. \citet{tomihari2026rl} studies update propagation in RL post-training. OPD needs different theoretical and empirical machinery for two reasons. Its on-policy sampling moves the learning distribution itself, so the analysis must track this motion, which is absent from the offline algorithms (SFT and offline DPO) studied by \citet{ren2025learning}. Its supervision is also dense and token-level and can be disconnected from the reasoning outcome (such as accuracy), whereas RL gives only sparse, outcome-level learning signals.

Nevertheless, the fundamental response law of the shared-parameter neural learner is shared with \citet{ren2025learning}. We then follow the induced motion through the rest of the OPD loop: changes at ancestor decisions alter reach probabilities, renew the training source, and thereby change the requests driving future updates. Combining the evolving kernel, routing/drift split, and continuation-value analysis yields a mechanistic account of how optimization changes future supervision and outcomes (\cref{sec:theory,app:derivations}).

\noindent\textbf{Cumulative weight geometry vs the geometry of request conversion.}
\citet{shen2026geometryonpolicydistillation} find low stable rank and early stabilization of dominant subspaces in cumulative weight-matrix changes; constrained-training experiments test those subspaces' functional sufficiency. We instead analyze the current logit Jacobian and the response it delivers to specified teacher requests. The Jacobian can change even when cumulative displacement is low-rank. Frozen histories, coordinates, and requests isolate amplification and reorientation; crossed teacher fields test preferential adaptation in the same kernel (\cref{app:collective-frozen,app:collective-specificity}). Distinguishing local accessibility from collective coherence further explains why individually expressible requests can cancel. These analyses concern how geometry enables or obstructs transfer.

\noindent\textbf{Local fitting vs task performance.}
Policy-gradient theory \citep{sutton2000policy} and cost-to-go-aware imitation learning \citep{ross2014reinforcementimitationlearninginteractive} connect actions to subsequent outcomes. G-OPD interprets distillation through KL-constrained RL and reward extrapolation \citep{yang2026gopd}. \citet{armandpour2026unmaskingonpolicydistillationhelps} use student continuation-success probabilities to derive an ideal local logit direction and measure its alignment with distillation signals (however, they acknowledge that such estimations in long-CoT reasoning can be very noisy). \cite{uemura2026learningoverwritingtheoryselfdistillation} theoretically studies the dynamics of self-distillation for a single-layer neural network, and analyzes how the supervision affects long-horizon traversal of directed acyclic graphs, invoking tools such as future advantage in their analysis. Our analysis connects this type of outcome perspective to \emph{delivered} motion: $\sum_s K_t(x,s)\Gamma_t(s)$ includes the effects of all coupled source prefixes in addition to the request at $x$. Crucially, prefix readouts $\mathbb{E}_{\nu_t}Y_t(x)$ and terminal outcomes $\mathbb{E}_{\Pi_t}Y$ are different objects. At an intermediate cut-position in the autoregressive rollout, terminal change splits into the change in reaching that prefix and the change in the continuations; unrolling the latter and recombining both terms yields the advantage-weighted contribution of every decision (\cref{eq:broadcast-continuations,app:outcome-proof}). This separates reaching a prefix, fitting there, and completing it successfully --the distinction underlying the degeneration mechanism and continuation readiness.

\noindent\textbf{Degeneration involves source feedback as well as supervision quality.}
\citet{fu2026revisitingonpolicydistillationempirical} study unreliable supervision and implementation choices; \citet{jin2026entropyawareonpolicydistillationlanguage} augment high-entropy teacher supervision with forward KL; and \citet{zhu2026manyfaces} analyze signals induced by top-$K$ truncation, renormalization, and stop-gradient conventions. \citet{rethinking_opd_2026} also observe deterioration with prefix depth. Contemporaneous work by \citet{wang2026demystifying} studies teacher mismatch and length exploitation. Our first-principled analysis reveals a phased failure mechanism: early routing makes already-uncertain continuations prevalent, after which substantial fitting can \emph{sharpen those very histories} without restoring useful generation (\cref{app:rd-failure,app:rd-entropy-evidence}). Our routing-drift machinery, coupled with the learning-value decomposition (for terminal-accuracy readout) distinguish the failure mechanism from a naive, uniform entropy-increasing learning process. The important feedback is that the histories made common by earlier updates become the source of later supervision. The prepared learner's different trajectory further shows why source movement itself is not synonymous with failure, and its continuation-readiness helps in exposing useful subsequent learning.

\noindent\textbf{Offline preparation can improve readiness for learning without improving immediate accuracy.}
\citet{rethinking_opd_2026} motivate cold-start SFT by reducing thinking-pattern mismatch and improving overlap. Their \S5.1/Figure~8 studies a strong standalone-SFT regime: the SFT initialization already exceeds the direct-OPD endpoint on all three plotted benchmarks, with little further gains from OPD. We study a different regime, where the student-teacher gap is significant, and preparation leaves starting accuracy near the unprepared student's but changes subsequent OPD from deterioration to substantial improvement. The principled motivation is \emph{continuation readiness}: the outcome analysis and observed failure mechanisms in \cref{sec:broadcast} identify a distinction between increasing a history's reach and executing its continuation successfully. Teacher-generated trajectories provide training on such histories before the student must reach them itself. Our dose study exposes deferred benefits and a sharply nonmonotonic stability pattern at 7B; the prepared-run analysis traces a different subsequent routing/fitting process (\cref{tab:dose-figure,app:rd-prepared}). 

In \emph{contemporaneous work} Simple-OPD \citep{liu2026simpleopd} studies warm-up data, duration, and LoRA parameterization, including benefits from incorrect teacher-compatible traces and differences between SFT and downstream rankings. These findings complement our mechanistic account linking preparation to source feedback and continuation values. The key result is an altered learning trajectory; the preparation dose-response experiments demonstrate this interesting benefit.

\subsection{More related works}
\noindent\textbf{Reasoning interpretability}. Part of the focus of our work is on how LLMs learn to reason (from a teacher model), and our analysis methodology is partly inspired by the discoveries in the reasoning interpretability community. \citet{venhoff2025base} is especially relevant, as they discovered that RL-trained models tend to acquire the ability to manage ``when'' to fire a reasoning mechanism (be it an ``epistemic control'' mechanism such as backtracking or planning, or computation mechanisms like estimating an inequality); \cite{marjanovic2026deepseekr} directly studies the chain-of-thought reasoning mechanisms of DeepSeek-R1 models, which our DeepSeek student models are distilled from, also inspiring our analysis on the epistemic control transmission in OPD. More broadly, \cite{bogdan2025thoughtanchorsllmreasoning, baek2025understandingdistilledreasoningmodels, venhoff2025understandingreasoningthinkinglanguage, troitskii-etal-2025-internal, NEURIPS2025_2c447757, ward2025reasoningfinetuningrepurposeslatentrepresentations} identify causal mechanisms behind LLMs' reasoning behaviors in logic, mathematics and other related domains.

\textbf{OPD and its variants}. On-policy distillation is becoming increasingly popular in industry post-training pipelines \citep{deepseekai2026deepseekv4highlyefficientmilliontoken, kimiteam2026kimik3openfrontier}. Many variants are being proposed, including selecting which tokens to perform supervision on \citep{jin2026entropyaware,yu2026mismatchmattersonpolicydistillation}, and curriculum-styled variants \citep{ziheng2026moreearlystoppingrollout,zhang2026rolloutsnecessaryonpolicydistillation,li2026onpolicydistillationcurriculumturnlevel}. The survey on OPD \citet{song2026surveyonpolicydistillationlarge} is also a good reference for understanding the research landscape in this area.

\textbf{Learning dynamics}. Many classic works lie in this space and are predecessors to recent studies of LLM learning dynamics. \cite{Biehl_1995,PhysRevLett.74.4337,saxe2014exactsolutionsnonlineardynamics,NEURIPS2019_cab070d5} are early representative works, studying how learning trajectories and generalization depend on the model, initialization, and training process, and providing foundations for our analysis of how teacher supervision changes the student over time. Furthermore, a particular line of work study feature learning in nonlinear neural networks: \cite{pmlr-v130-baratin21a,atanasov2022neural,hong2024why,9719702,pmlr-v195-allen-zhu23a} study how the nonlinear neural network’s geometry evolves and aligns with the task, showing feature re-orientation early on in (pre-)training that precedes appreciable loss reduction (sometimes called the ``lottery winning'' phase, where certain neurons or features better align towards loss-reducing directions early in training without substantial loss-reduction yet). These works are especially relevant to our analysis of the student’s changing responsiveness to teacher requests in \cref{sec:conversion}, and to offline preparation whose benefits become apparent through subsequent learning. Additionally, another relevant line of work in the pre-LLM era study dynamics of DNN distillation. \cite{pmlr-v97-phuong19a,NEURIPS2020_ef0d3930,Hong_2021_CVPR,allenzhu2023understandingensembleknowledgedistillation,10.1145/3511808.3557237,he2022feature} examine how the geometry of the learning problem, teacher knowledge, and the student’s representation affect transfer, using linear, deep linear, or kernel-based analyses of the SGD-based learning trajectory.
\FloatBarrier
\clearpage
\section{Derivations and proofs for the main text}
\label{app:derivations}
This appendix follows \cref{sec:setup,sec:theory}. It gives the assumptions and proofs for the main-text identities, then the specializations used to interpret the experiments.

\begin{center}
\small
\begin{tabular}{@{}p{.36\linewidth}p{.59\linewidth}@{}}
\toprule
Main-text result & Complete derivation \\
\midrule
\Cref{sec:setup}: policies and request field & \Cref{app:standing}: finite-tree conventions and exact-loss request \\
\Cref{sec:loop}, \cref{eq:response-law} & \Cref{app:ftc}: exact response and explicit finite-step remainder \\
\Cref{sec:local}, \cref{eq:collective-response,eq:pac} & \Cref{app:collective-proof,app:collective-frozen}: conversion and frozen-field geometry \\
\Cref{sec:broadcast}, \cref{eq:rd-law,eq:rd-velocity} & \Cref{app:rd-flow}: routing and conditional drift \\
\Cref{sec:broadcast}, \cref{eq:broadcast-source-renewal} & \Cref{app:broadcast-prefix-reweighting}: source normalization and support \\
\Cref{sec:broadcast}, \cref{eq:rd-estimators} & \Cref{app:rd-theory}: endpoint accounting and path-integral error \\
\Cref{sec:broadcast}, \cref{eq:broadcast-continuations} & \Cref{app:outcome-proof}: continuation values and terminal outcomes \\
\Cref{sec:broadcast}: entropy versus fitting & \Cref{app:rd-entropy-response}: entropy and checking sensitivities \\
\bottomrule
\end{tabular}
\end{center}

\subsection{Setup: policies, requests, and standing assumptions}
\label{app:standing}
\textbf{Connection to \cref{sec:setup}.} We distinguish the model distribution, the sampling distribution, and the numerical field pulled back into parameter space. This distinction is needed for both the proofs and the experimental diagnostics.

\paragraph{Notation.}
A prompt is denoted by $b$, a source prefix by $s$, and a receiver prefix by $x$. A complete trajectory is $\xi$; $\tau>0$ is reserved for sampling temperature. The integer $k$ indexes optimizer updates, $j$ indexes generated-token depth, and $t$ indexes continuous training time. All gradients of scalar functions are column vectors. The matrix $\Phi_\theta(x)\in\mathbb R^{|\cV|\times d}$ has row $a$ equal to $\nabla_\theta z_\theta(x,a)^\top$, so $K_\theta(x,s)=\Phi_\theta(x)\Phi_\theta(s)^\top$. We write $\mathsf A^p=I-\mathbf1p^\top$ for the log-softmax normalization matrix, $A$ for scalar accessibility, and $\mathcal A^Y$ for continuation advantage. A logit-space readout gradient is denoted by $\chi\in\mathbb R^{|\cV|}$; its parameter-space pullback is $\psi=\Phi^\top\chi\in\mathbb R^d$. Norms without a subscript are Euclidean. $\|\cdot\|_{\rm op}$ denotes the induced operator norm.

\begin{assumption}[Finite generation trees and regular parameter paths]
\label[assumption]{ass:regularity}
The prompt set $\mathcal B$ and vocabulary $\cV$ are finite, and generation has a finite cap $H_{\max}$. Each prefix contains its prompt and its full generated ancestry. Appending a token is deterministic. A node is terminal after a stopping token or at the cap, and has no further decisions. The terminal rule and chat template are fixed. The prompt distribution $\beta$ is fixed (and independent of model parameters); prompts of zero probability are omitted.

The parameter path $\theta:I\to\mathbb R^d$, on a compact interval $I$, is continuous and piecewise $C^1$ with finitely many pieces. For every prefix under consideration, $z_\theta(x)$ is $C^2$ on an open neighborhood of the path and of any parameter segment used for a finite-step calculation. The model distribution is the full-vocabulary policy $p_\theta=\softmax(z_\theta)$. The teacher $q$ is fixed and is positive on every coordinate used in an exact-KL calculation (which softmax-output models do satisfy).

For the sampling identities, each nonterminal prefix $x$ has a fixed nonempty allowed vocabulary set $\mathcal V_\pi(x)\subseteq\cV$ and
\[
 \pi_\theta(a\mid x)=
 \frac{\exp(z_\theta(x,a)/\tau)}{\sum_{c\in\mathcal V_\pi(x)}\exp(z_\theta(x,c)/\tau)}
 \quad(a\in\mathcal V_\pi(x)),\qquad
 \pi_\theta(a\mid x)=0\quad(a\notin\mathcal V_\pi(x)).
\]
The temperature $\tau>0$, allowed sets, and any observation or source masks are fixed. The full-vocabulary sampler is the case $\mathcal V_\pi(x)=\cV$.
\end{assumption}

Let $\mathcal S$ and $\mathcal T$ be the resulting finite sets of nonterminal and terminal prefixes. A trajectory $\xi=(b,a_0,\ldots,a_{N-1})$ ends at its first terminal node, with $N\le H_{\max}$, and has probability
\[
 \Pi_t(\xi)=\beta(b)\prod_{j=0}^{N-1}\pi_t(a_j\mid x_j).
\]
For any node $x$, including a terminal node, $d_t(x)$ is its reach-probability. It is the same product truncated at depth $h(x)$. On terminal nodes $d_t(x)=\Pi_t(x)$; on nonterminal nodes the sum $\sum_xd_t(x)$ is generally not one. Every allowed trajectory has positive probability. Finiteness and the stated regularity justify every interchange of sums and derivatives below. Identities for derivatives hold on each smooth piece, hence almost everywhere, and integrate across the entire interval. The results take the path as given and are independent of any existence or uniqueness theorem for an autonomous training flow.

The raw direction is
\[
 g_t(s)=\Phi_t(s)^\top B_t(s),\qquad
 u_t=\sum_s\mu_t(s)g_t(s),\qquad \Gamma_t(s)=\mu_t(s)B_t(s).
\]
Here $B_t$ is a declared numerical logit-space field. The specialization $\dot\theta_t=u_t$ is called the raw flow. An actual optimizer path uses its realized $\dot\theta_t$. Restricted diagnostic fields are extended by zero outside their recorded token coordinates, and their Jacobians are restricted consistently; these diagnostic supports leave the sampler's allowed set unchanged.

\begin{lemma}[Log-softmax differentiation and centering]
\label[lemma]{lem:logsoftmax}\label[lemma]{lem:centering}
Write $\pi=\pi_\theta(\cdot\mid x)$ as a vector in the full vocabulary, with zeros outside its fixed allowed set, and let $\mathsf A^\pi=I-\mathbf1\pi^\top$. For each allowed token $a$,
\[
 \nabla_\theta\log\pi_\theta(a\mid x)
 =\tau^{-1}\Phi_\theta(x)^\top(e_a-\pi),\qquad
 \sum_a\pi(a\mid x)\nabla_\theta\log\pi(a\mid x)=0.
\]
Consequently $\zeta_t(x,a):=\partial_t\log\pi_t(a\mid x)$ satisfies $\sum_a\pi_t(a\mid x)\zeta_t(x,a)=0$. If $\mathbf1^\top w=0$, then $(\mathsf A^\pi)^\top w=w$. Sums involving $\log\pi$ or $\zeta$ range only over allowed tokens.
\end{lemma}
\begin{proof}
For allowed $a$, the logarithm of the displayed sampler in Assumption~\ref{ass:regularity} is
\[
 \log\pi(a\mid x)=\tau^{-1}z(x,a)
 -\log\sum_{c\in\mathcal V_\pi(x)}\exp(z(x,c)/\tau).
\]
Its partial derivative with respect to any full-vocabulary logit $z(x,c)$ is $\tau^{-1}(\mathbf1\{a=c\}-\pi(c\mid x))$, including the value zero when $c$ is disallowed. The chain rule proves the gradient formula. Averaging over allowed $a$ yields
\[
 \tau^{-1}\Phi(x)^\top\left(\sum_a\pi(a\mid x)e_a
                 -\pi\sum_a\pi(a\mid x)\right)=0.
\]
Pairing this equality with $\dot\theta_t$ gives the zero-mean velocity. Finally,
$(\mathsf A^\pi)^\top w=(I-\pi\mathbf1^\top)w=w-\pi(\mathbf1^\top w)=w$.
\end{proof}

\begin{proposition}[Exact reverse-KL request]
\label[proposition]{prop:request}
At a fixed prefix, let $c=\sum_a p_a\log(p_a/q_a)$, $\ell_a=\log(p_a/q_a)$, and $\bar\ell=\sum_a p_a\ell_a$, with $p=\softmax(z)$ and $q_a>0$. Then
\begin{equation}
 B_a=-\partial_{z_a}c=p_a(\bar\ell-\ell_a),\qquad
 \mathbf1^\top B=0,\qquad g=\Phi^\top B=-\nabla_\theta c.
 \label{eq:request}
\end{equation}
For fixed source weights $\mu$, $u=\sum_s\mu(s)g(s)$ is the negative gradient of $\sum_s\mu(s)c(s)$.
\end{proposition}
\begin{proof}
Positivity makes the logarithms differentiable. Since
$\partial_{z_a}p_b=p_b(\mathbf1\{a=b\}-p_a)$ and
$\partial_{p_b}[p_b\log(p_b/q_b)]=\ell_b+1$, we have
\begin{align*}
 \partial_{z_a}c
 &=\sum_b p_b(\mathbf1\{a=b\}-p_a)(\ell_b+1)\\
 &=p_a(\ell_a+1)-p_a\sum_b p_b(\ell_b+1)
 =p_a(\ell_a-\bar\ell).
\end{align*}
Thus $\sum_a B_a=\bar\ell\sum_a p_a-\sum_a p_a\ell_a=0$.
The chain rule gives $\nabla_\theta c=\Phi^\top\nabla_z c$; linearity of the finite sum proves the last assertion when $\mu$ is held fixed.
\end{proof}

This is a full-distribution identity. For an implemented objective, such as a truncated-support loss or detached-coefficient surrogate, $B$ must be that objective's own complete numerical negative logit derivative, including the softmax coupling. For the conversion diagnostic, $B$ denotes the explicitly specified audit field in \cref{app:collective-conventions}. The response and conversion results hold for this field's pullback, whether or not that pullback coincides with exact-KL descent. Holding $\mu$ fixed applies to computing one update; subsequent source populations can change.

\subsection{Response law: exact finite steps and controlled remainders}
\label{app:ftc}
\textbf{Connection to \cref{sec:loop}, \cref{eq:response-law}.} The integral is exact for an arbitrary realized step. The main-text kernel expression is its raw-gradient, first-order specialization.

\begin{proposition}[Exact response and Taylor remainder]
\label[proposition]{prop:ftc}
Let $\Delta\theta=\theta_{k+1}-\theta_k$, and let $\theta_r=\theta_k+r\Delta\theta$ for $0\le r\le1$. For any fixed receiver $x$ and token $a$,
\begin{equation}
 \delta_k^p(x,a)=\int_0^1\nabla_\theta\log p_{\theta_r}(a\mid x)^\top\Delta\theta\,dr.
 \label{eq:ftc}
\end{equation}
If $M_k(x,a)=\sup_{0\le r\le1}\|\nabla_\theta^2\log p_{\theta_r}(a\mid x)\|_{\rm op}$, then
\[
 \delta_k^p(x,a)=\nabla_\theta\log p_k(a\mid x)^\top\Delta\theta+R_k(x,a),
 \qquad |R_k(x,a)|\le\tfrac12M_k(x,a)\|\Delta\theta\|^2.
\]
The corresponding exact logit displacement is
$z_{k+1}(x)-z_k(x)=\int_0^1\Phi_{\theta_r}(x)\Delta\theta\,dr$.
\end{proposition}
\begin{proof}
Define $f(r)=\log p_{\theta_r}(a\mid x)$. Its first two derivatives are
\[
 f'(r)=\nabla_\theta\log p_{\theta_r}(a\mid x)^\top\Delta\theta,
 \qquad
 f''(r)=\Delta\theta^\top\nabla_\theta^2\log p_{\theta_r}(a\mid x)\Delta\theta.
\]
The fundamental theorem of calculus gives \cref{eq:ftc}. Applying it also to $f'$ and changing the order of integration over $0\le v\le r\le1$ gives
\[
 f(1)-f(0)-f'(0)=\int_0^1\int_0^r f''(v)\,dv\,dr
               =\int_0^1(1-v)f''(v)\,dv.
\]
The Hessian is continuous on a compact segment, so $M_k(x,a)<\infty$.
The bound follows from $|f''(v)|\le M_k(x,a)\|\Delta\theta\|^2$ and $\int_0^1(1-v)\,dv=1/2$.
Applying the same first-derivative calculation to each logit proves the vector identity.
\end{proof}

\begin{corollary}[Small-step response law]
\label[corollary]{cor:response}
For $\Delta\theta_k=\eta u_k$ with $u_k=\sum_s\mu_k(s)\Phi_k(s)^\top B_k(s)$,
\[
 \delta_k^p(x,a)=\eta\sum_s\mu_k(s)
 [\mathsf A_k^p(x)K_k(x,s)B_k(s)]_a+R_k(x,a),\qquad
 |R_k(x,a)|\le\tfrac12\eta^2M_k(x,a)\|u_k\|^2.
\]
For fixed $\theta_k,u_k$ and sufficiently small $\eta$, the remainder is $O(\eta^2)$, as in \cref{eq:response-law}.
\end{corollary}
\begin{proof}
At temperature one, \cref{lem:logsoftmax} gives
$\nabla_\theta\log p_k(a\mid x)^\top=e_a^\top\mathsf A_k^p(x)\Phi_k(x)$.
Substituting $u_k$ in \cref{prop:ftc} therefore yields
\[
 \nabla_\theta\log p_k(a\mid x)^\top u_k
 =\sum_s\mu_k(s)e_a^\top\mathsf A_k^p(x)
        \Phi_k(x)\Phi_k(s)^\top B_k(s),
\]
which is the asserted kernel term. The remainder bound follows by substituting $\Delta\theta_k=\eta u_k$. To justify the asymptotic notation, choose $\eta_0>0$ so that all segments with $|\eta|\le\eta_0$ lie in a compact neighborhood where the Hessian is continuous. Its supremum on this neighborhood supplies a single bound independent of $\eta$. A family of varying directions requires a uniform bound on their norms as well.
\end{proof}

\paragraph{Velocity and the optimizer.}
At every differentiable time on an actual parameter path,
\[
 v_t(x):=\dot z_t(x)=\Phi_t(x)\dot\theta_t,
 \qquad \zeta_t(x,a)=\tau^{-1}(e_a-\pi_t(\cdot\mid x))^\top v_t(x).
\]
These follow directly by the chain rule and \cref{lem:logsoftmax}. Under gradient flow, substituting $\dot\theta_t=u_t$ gives $v_t(x)=\sum_sK_t(x,s)\Gamma_t(s)$ and hence \cref{eq:rd-velocity}. With momentum, preconditioning, clipping, or weight decay, the actual $\dot\theta_t$ in general differs from the raw $u_t$. However, with the rich tradition of theoretical analysis on the learning dynamics of neural networks including the seminal \cite{ren2025learning, jacot2018neural, pmlr-v97-arora19a}, gradient-flow analysis is still genuinely informative in obtaining rich theoretical insights. We leave extensions to the more convoluted optimizer settings to future work.

\subsection{Collective response and its geometry}
\label{app:collective-proof}\label{app:pac}
\textbf{Connection to \cref{sec:local}, \cref{eq:collective-response,eq:pac}.} We fix the checkpoint, output coordinates, and normalized bank $\nu$ in this subsection. Zero-weight sites are omitted. The space $\mathcal H_\nu$ of fields on the remaining finite bank has the inner product
\[
 \langle F,G\rangle_\nu=\sum_s\nu(s)F(s)^\top G(s),\qquad
 \|F\|_\nu^2=\langle F,F\rangle_\nu.
\]
Define $\mathcal J:\mathbb R^d\to\mathcal H_\nu$ by $(\mathcal Jv)(s)=\Phi(s)v$. This is a finite-dimensional Hilbert space, so orthogonal projections and adjoints are well-defined.

\begin{proposition}[Collective response and conversion factorization]
\label[proposition]{prop:collective}\label[proposition]{prop:pac}
Let $g(s)=\Phi(s)^\top B(s)$, $u=\Ex_\nu g$, $P=\|B\|_\nu^2$, and $D_g=\Ex_\nu\|g\|^2$. The adjoint is $\mathcal J^*B=u$, and $\mathcal K=\mathcal J\mathcal J^*$ is self-adjoint and positive semidefinite. For the same-bank raw direction and $P>0$,
\[
 \dot z=\mathcal K B=\Xi B+r,\qquad
 \Xi=\frac{\langle B,\mathcal K B\rangle_\nu}{P}
     =\frac{\|u\|^2}{P}\ge0,\qquad \langle B,r\rangle_\nu=0.
\]
If also $D_g>0$, then
\[
 A=\frac{D_g}{P},\qquad C=\frac{\|u\|^2}{D_g},\qquad
 \Xi=AC,\qquad \|u\|^2=PAC,\qquad 0\le C\le1.
\]
Here $C=1$ exactly when $g(s)=u$ at every positive-weight site, and $C=0$ exactly when $u=0$. If $P>0$ but $D_g=0$, then $u=0$, $A=\Xi=0$, and $C$ is undefined. If $P=0$, then $B=g=u=0$ on the bank and $A,C,\Xi$ are not defined by these ratios.
\end{proposition}
\begin{proof}
For any field $F$ and parameter vector $v$,
\[
 \langle F,\mathcal Jv\rangle_\nu
 =\sum_s\nu(s)F(s)^\top\Phi(s)v
 =\left(\sum_s\nu(s)\Phi(s)^\top F(s)\right)^\top v.
\]
This proves the adjoint formula. It also gives
\[
 \langle F,\mathcal KG\rangle_\nu
 =(\mathcal J^*F)^\top(\mathcal J^*G)
 =\langle\mathcal KF,G\rangle_\nu,
 \qquad \langle F,\mathcal KF\rangle_\nu=\|\mathcal J^*F\|^2\ge0.
\]
By the chain rule, the logit velocity generated by $u$ is $\mathcal Ju=\mathcal KB$. Thus $\langle B,\dot z\rangle_\nu=\|u\|^2$. Since $P>0$, the strictly convex quadratic $\alpha\mapsto\|\dot z-\alpha B\|_\nu^2$ has its unique minimum at $\alpha=\langle B,\dot z\rangle_\nu/P$. This is $\Xi$, and the residual has the stated orthogonality.

For $D_g>0$, multiplication of the defined ratios gives $\Xi=AC$ and $\|u\|^2=PAC$. Expanding a centered squared norm gives
\[
 \Ex_\nu\|g-u\|^2
 =\Ex_\nu\|g\|^2-2u^\top\Ex_\nu g+\|u\|^2
 =D_g-\|u\|^2\ge0.
\]
This proves $C\le1$ and its equality condition; nonnegativity of $C$ and its zero condition follow from its numerator. If $D_g=0$, every nonnegative summand $\nu(s)\|g(s)\|^2$ vanishes, so $g=u=0$. If $P=0$, the same argument first gives $B=0$, then $g=u=0$. These cases cannot be handled by cancelling zero denominators.
\end{proof}

\begin{proposition}[Joint singular modes]
\label[proposition]{prop:joint-modes}
Stack the blocks $\sqrt{\nu(s)}\Phi(s)$ into $\mathsf J\in\mathbb R^{N_\nu\times d}$ and $\sqrt{\nu(s)}B(s)$ into $\mathbf B\in\mathbb R^{N_\nu}$. Let the compact singular-value decomposition be $\mathsf J=U\Sigma V^\top$, of rank $r$, with orthonormal columns $U_j,V_j$ and positive singular values $\sigma_j$. If $P>0$,
\begin{equation}
 u=\sum_{j=1}^r\sigma_j(U_j^\top\mathbf B)V_j,\qquad
 \Xi=\sum_{j=1}^r\sigma_j^2\frac{|U_j^\top\mathbf B|^2}{\|\mathbf B\|^2}.
 \label{eq:collective-modes}
\end{equation}
The component of $\mathbf B$ orthogonal to $\operatorname{range}(\mathsf J)$ produces zero parameter response.
\end{proposition}
\begin{proof}
The stacking gives $\mathsf J^\top\mathbf B=\sum_s\nu(s)\Phi(s)^\top B(s)=u$ and $\|\mathbf B\|^2=P$. Substitution of the singular-value decomposition proves the first identity. Since $V_i^\top V_j=\mathbf1\{i=j\}$, its squared norm is $\sum_j\sigma_j^2|U_j^\top\mathbf B|^2$; division by $P$ proves the second. Finally $\mathsf J^\top\mathbf B_\perp=V\Sigma U^\top\mathbf B_\perp=0$ when $U^\top\mathbf B_\perp=0$. If $r=0$, these are empty sums and $u=\Xi=0$.
\end{proof}

Joint modes span multiple prefixes and tokens. Their parameter-space components are orthogonal, whereas individual per-prefix gradients need not be. Strong cancellation between local gradients is therefore compatible with large self-kernel responses but a small joint response; independently sorted self-kernel spectra do not determine \cref{eq:collective-modes}.

\begin{proposition}[Coherence and interference]
\label[proposition]{prop:coherence}
For independent $s,s'\sim\nu$,
\[
 \|u\|^2=\Ex\langle g(s),g(s')\rangle
        =\Ex_\nu\|g(s)\|^2-\tfrac12\Ex\|g(s)-g(s')\|^2.
\]
If $B=\sum_{m=1}^M B^{(m)}$ is any finite field decomposition and $u^{(m)}=\Ex_\nu[\Phi^\top B^{(m)}]$, then
$\|u\|^2=\sum_m\|u^{(m)}\|^2+2\sum_{m<n}\langle u^{(m)},u^{(n)}\rangle$.
\end{proposition}
\begin{proof}
Independence gives the explicit finite-sum identity
\[
 \Ex\langle g(s),g(s')\rangle
 =\sum_{s,s'}\nu(s)\nu(s')g(s)^\top g(s')
 =\left(\sum_s\nu(s)g(s)\right)^\top\left(\sum_{s'}\nu(s')g(s')\right).
\]
This is $\|u\|^2$. Expanding the squared difference and using the identical marginals yields $\Ex\|g(s)-g(s')\|^2=2D_g-2\|u\|^2$. The field decomposition gives $u=\sum_m u^{(m)}$ by linearity, and the last formula follows by expanding the square and collecting both copies of each off-diagonal inner product.
\end{proof}

\begin{proposition}[Finite-sample correction of the mean-gradient energy]
\label[proposition]{prop:gradient-norm-correction}
Let $g_1,\ldots,g_n$, $n>1$, be independent identically distributed parameter vectors with finite second moment and mean $u$. Define $\bar g=n^{-1}\sum_i g_i$ and $\widehat D_g=n^{-1}\sum_i\|g_i\|^2$. Then
\[
 \widehat U=\frac{n\|\bar g\|^2-\widehat D_g}{n-1}
           =\frac{1}{n(n-1)}\sum_{i\ne j}g_i^\top g_j
\]
has expectation $\|u\|^2$.
\end{proposition}
\begin{proof}
Expanding $n^2\|\bar g\|^2$ gives $\sum_i\|g_i\|^2+\sum_{i\ne j}g_i^\top g_j$. Subtracting $n\widehat D_g=\sum_i\|g_i\|^2$ and dividing by $n(n-1)$ proves the second expression. Independence gives $\Ex[g_i^\top g_j]=(\Ex g_i)^\top(\Ex g_j)=\|u\|^2$ for each of its $n(n-1)$ terms. Thus its expectation is $\|u\|^2$. The finite second moment makes every term integrable. Positive clipping and division by the random $\widehat D_g$ are nonlinear, so the bias of the resulting coherence ratio falls outside this linear expectation calculation. Other sampling designs, including sampling without replacement from a fixed finite population, require their own design-specific bias calculation.
\end{proof}

\paragraph{Policy-level observation of a logit response.}
\label{app:contrast-observation}
For a fixed vector $w(x)$ with $\mathbf1^\top w(x)=0$,
\[
 w(x)^\top[z_{\theta'}(x)-z_\theta(x)]
 =w(x)^\top[\log p_{\theta'}(\cdot\mid x)-\log p_\theta(\cdot\mid x)].
\]
Indeed, $\log p_\theta=z_\theta-\log\sum_a e^{z_\theta(a)}\mathbf1$, and both scalar-normalizer terms disappear under $w^\top$. The identity holds for finite displacements, for zero-extended contrasts on a recorded support, and for centered and uncentered training fields alike. The experimental contrast is specified in \cref{app:collective-conventions}.

\subsection{Why the frozen spectral probe measures accessibility}
\label{app:collective-frozen}
\textbf{Connection to \cref{sec:local} and the fixed-bank diagnostic in \cref{sec:preparation}.} All time dependence below is in the tangent map; the bank, coordinates, and numerical field are fixed.

\begin{proposition}[Frozen-field Rayleigh expansion]
\label[proposition]{prop:frozen-rayleigh}
Let $P_0=\Ex_{\nu_0}\|B_0\|^2>0$, and define on nonzero-request sites
\[
 \omega_0(s)=\frac{\nu_0(s)\|B_0(s)\|^2}{P_0},\qquad
 b_0(s)=\frac{B_0(s)}{\|B_0(s)\|}.
\]
Zero-request sites have weight zero and are omitted from sums involving $b_0$. For any nonzero field on a specified bank,
\begin{equation}
 A[B]=\frac{\Ex_\nu[B(s)^\top K(s,s)B(s)]}{\Ex_\nu\|B(s)\|^2}.
 \label{eq:rayleigh}
\end{equation}
Let $(\lambda_{j,t}(s),v_{j,t}(s))$ be a complete orthonormal eigendecomposition of $K_t(s,s)$. Then
\begin{equation}
 \begin{aligned}
 a_t(s)&=b_0(s)^\top K_t(s,s)b_0(s)
       =\sum_j\lambda_{j,t}(s)|v_{j,t}(s)^\top b_0(s)|^2,\\
 A_t[B_0]&=\sum_s\omega_0(s)a_t(s).
 \end{aligned}
 \label{eq:afrozen}
\end{equation}
The weights sum to one. In particular, $A_t[B_0]$ is constant whenever the self-kernels are constant on the bank.
\end{proposition}
\begin{proof}
For each site,
$\|\Phi(s)^\top B(s)\|^2=B(s)^\top\Phi(s)\Phi(s)^\top B(s)=B(s)^\top K(s,s)B(s)$.
Substituting into the definition $A=D_g/P$ proves \cref{eq:rayleigh}. In the frozen case, write $B_0(s)=\|B_0(s)\|b_0(s)$ at each nonzero site and divide the weighted sum by $P_0$; this gives $A_t[B_0]=\sum_s\omega_0(s)b_0(s)^\top K_t(s,s)b_0(s)$. The weights sum to $P_0/P_0=1$.

Each $K_t(s,s)=\Phi_t(s)\Phi_t(s)^\top$ is symmetric and positive semidefinite: for every vector $v$, $v^\top K_t(s,s)v=\|\Phi_t(s)^\top v\|^2\ge0$. The finite-dimensional spectral theorem therefore gives $K_t=\sum_j\lambda_{j,t}v_{j,t}v_{j,t}^\top$ with $\lambda_{j,t}\ge0$. Inserting this sum into the quadratic form proves \cref{eq:afrozen}. All coefficients except $K_t$ are fixed, which proves the constancy assertion. Repeated eigenvalues do not affect the full sum: the sum of squared projections within an eigenspace is basis-invariant.
\end{proof}

The live quantity $A_t[B_t]$ can also change through its field and energy weights. The frozen collective gain is $\Xi_t[B_0]=A_t[B_0]C_t[B_0]$ when the denominators are nonzero; both factors use $B_0$, whereas $A_t[B_0]C_t[B_t]$ mixes two fields and is a different quantity.

\subsection{Routing and conditional drift on moving prefix populations}
\label{app:rd-flow}
\textbf{Connection to \cref{sec:broadcast}, \cref{eq:rd-law,eq:rd-velocity}.} Let $m^{\rm obs}(x)\ge0$ be a fixed weight on the nonterminal prefixes of the observation tree. Assume at least one eligible prefix has positive reach probability, and define
\begin{equation}
 \nu_t(x)=\frac{m^{\rm obs}(x)d_t(x)}{Z_t^{\rm obs}},\qquad
 Z_t^{\rm obs}=\sum_xm^{\rm obs}(x)d_t(x)>0.
 \label{eq:rd-observation}
\end{equation}
This is normalization of expected weighted counts, which allows fixed depth masks, context classes defined by a fixed reference learner, and empirical minibatches of unequal length.

\begin{proposition}[Routing and conditional drift]
\label[proposition]{prop:rd-law}
Let $Y_t(x)$ be a scalar readout that is continuous and piecewise $C^1$ along the parameter path, and put $J_Y(t)=\sum_x\nu_t(x)Y_t(x)$. For a positive-reach prefix with ancestry $(x_j,a_j)_{j<h(x)}$, define
\[
 \lambda_t(x)=\partial_t\log d_t(x)
            =\sum_{j<h(x)}\zeta_t(x_j,a_j).
\]
Then, almost everywhere,
\begin{align*}
 \dot\nu_t(x)&=\nu_t(x)\bigl(\lambda_t(x)-\Ex_{\nu_t}\lambda_t\bigr),\\
 \dot J_Y(t)&=R_Y(t)+D_Y(t),\qquad
 R_Y(t)=\operatorname{Cov}_{\nu_t}(Y_t,\lambda_t),\quad
 D_Y(t)=\Ex_{\nu_t}\dot Y_t.
\end{align*}
If $Y_t(x)=Y(z_t(x),x)$ for a $C^1$ local function $Y$ with no explicit time dependence, then $\dot Y_t(x)=\nabla_zY(z_t(x),x)^\top v_t(x)$. Under raw flow, $v_t=K_t\Gamma_t$. For every subinterval $[a,b]$, $J_Y(b)-J_Y(a)=\int_a^b(R_Y+D_Y)\,dt$.
\end{proposition}
\begin{proof}
For fixed $x$, the prompt $b_x$ and all tokens in its ancestry are fixed while training time changes. Thus
\[
 d_t(x)=\beta(b_x)\prod_{j<h(x)}\pi_t(a_j\mid x_j),\qquad
 \partial_t\log d_t(x)=\sum_{j<h(x)}\partial_t\log\pi_t(a_j\mid x_j).
\]
This proves the formula for $\lambda_t$; at a root the product is empty and $\lambda_t=0$. Also $\dot d_t(x)=d_t(x)\lambda_t(x)$. Differentiating the finite normalizer gives
\[
 \frac{\dot Z_t^{\rm obs}}{Z_t^{\rm obs}}
 =\frac{\sum_xm^{\rm obs}(x)d_t(x)\lambda_t(x)}{Z_t^{\rm obs}}
 =\Ex_{\nu_t}\lambda_t.
\]
The quotient rule now gives the asserted formula for $\dot\nu_t$ at each eligible site. For a site with zero fixed mask, both $\nu_t$ and $\dot\nu_t$ are zero, so the same expression holds with that site omitted.

The product rule and the definition of covariance yield
\begin{align*}
 \dot J_Y
 &=\sum_x\dot\nu_t(x)Y_t(x)+\sum_x\nu_t(x)\dot Y_t(x)\\
 &=\Ex_{\nu_t}(Y_t\lambda_t)
   -(\Ex_{\nu_t}Y_t)(\Ex_{\nu_t}\lambda_t)
   +\Ex_{\nu_t}\dot Y_t.
\end{align*}
For a local readout, the chain rule gives $\dot Y_t=\nabla_zY^\top\dot z_t$; the raw specialization follows from \cref{app:ftc}. All functions are continuous and piecewise $C^1$ on a finite partition of the compact interval. Integrating the derivative on each piece and telescoping the boundary values proves the final equality.
\end{proof}

\paragraph{The two roles of the same parameter change.}
Define $\mathcal F(\theta_g,\theta_e)=\sum_x\nu_{\theta_g}(x)Y_{\theta_e}(x)$. Applying the multivariable chain rule to $\mathcal F(\theta_t,\theta_t)$ gives its generator derivative $\sum_x\dot\nu_tY_t=R_Y$ and evaluator derivative $\sum_x\nu_t\dot Y_t=D_Y$. Both are contributions of the same parameter change to one derivative. A readout depending on the full parameter vector, rather than only the logits at $x$, uses $\dot Y_t(x)=\nabla_\theta Y_\theta(x)^\top\dot\theta_t$ instead. This distinction will matter for continuation values.

If prompt weights or observation masks vary explicitly, the formula for $\lambda_t$ alone is insufficient. On positive coordinates the quotient-rule calculation instead centers $\partial_t\log[m_t^{\rm obs}(x)d_t(x)]$; $\partial_t\log d_t$ includes $\partial_t\log\beta_t(b_x)$ for a changing prompt law. Zero weights or changes of support require separate boundary accounting. The fixed-weight assumptions are therefore substantive.

\subsection{Finite-window exposure and regenerated source weights}
\label{app:broadcast-prefix-reweighting}\label{app:broadcast-finite-window}
\textbf{Connection to \cref{sec:broadcast}, \cref{eq:broadcast-source-renewal}.} Use a superscript ${\rm tr}$ when training and observation employ different fixed prompt laws or samplers. In particular, $\tau_{\rm tr}$ can differ from the observation temperature. For a fixed training mask $m^{\rm tr}(s)\ge0$, let
\[
 \mu_t(s)=\frac{m^{\rm tr}(s)d_t^{\rm tr}(s)}{Z_t^{\rm tr}},\qquad
 Z_t^{\rm tr}=\sum_s m^{\rm tr}(s)d_t^{\rm tr}(s)>0.
\]
The following proposition assumes one fixed eligible support throughout the interval, as guaranteed by Assumption~\ref{ass:regularity}. Merely sharing some positive-support states at the endpoints is not sufficient for the unrestricted normalization formula.

\begin{proposition}[Source reweighting across a training window]
\label[proposition]{prop:source-reweighting}
For $a\le b$, let $\Lambda_{a:b}(s)=\int_a^b\lambda_t^{\rm tr}(s)\,dt$, where $\lambda_t^{\rm tr}$ is the ancestral log reach-probability velocity under the training sampler. On every allowed prefix,
\begin{equation}
 \frac{d_b^{\rm tr}(s)}{d_a^{\rm tr}(s)}=e^{\Lambda_{a:b}(s)},\qquad
 \Lambda_{a:c}(s)=\Lambda_{a:b}(s)+\Lambda_{b:c}(s)\quad(a\le b\le c).
 \label{eq:broadcast-prefix-ratio}
\end{equation}
For each eligible source prefix,
\begin{equation}
 \mu_b(s)=\frac{\mu_a(s)e^{\Lambda_{a:b}(s)}}
 {\Ex_{v\sim\mu_a}e^{\Lambda_{a:b}(v)}},\qquad
 \Gamma_b(s)=\mu_b(s)B_b(s).
 \label{eq:broadcast-regenerated-requests}
\end{equation}
A zero fixed mask stays zero.
\end{proposition}
\begin{proof}
By \cref{prop:rd-law} with the training sampler, $\partial_t\log d_t^{\rm tr}(s)=\lambda_t^{\rm tr}(s)$ on each smooth piece. Integrating and telescoping gives $\log d_b^{\rm tr}(s)-\log d_a^{\rm tr}(s)=\Lambda_{a:b}(s)$. Exponentiating proves the ratio, and additivity of the integral proves the second identity.

Because the eligible set is the same at both endpoints, substitution into the entire source normalizer is valid:
\begin{align*}
 Z_b^{\rm tr}
 &=\sum_s m^{\rm tr}(s)d_a^{\rm tr}(s)e^{\Lambda_{a:b}(s)}\\
 &=Z_a^{\rm tr}\sum_s\mu_a(s)e^{\Lambda_{a:b}(s)}.
\end{align*}
The expectation is strictly positive and finite, since it is a finite sum of positive terms on the nonempty eligible set. Dividing $m^{\rm tr}(s)d_b^{\rm tr}(s)$ by this expression for $Z_b^{\rm tr}$ proves \cref{eq:broadcast-regenerated-requests}. The expression for $\Gamma_b$ defines the regenerated request population with the current numerical field $B_b$. If $m^{\rm tr}(s)=0$, both endpoint weights are zero directly from their definition.
\end{proof}

An eligible prefix can be absent from a finite recorded batch while having positive population weight at every checkpoint. The reference in \cref{eq:broadcast-regenerated-requests} is the population $\mu_a$, which includes such prefixes. At a root, $\Lambda=0$, although normalization by the total expected token count can change its source weight.

\subsubsection{Finite-batch extraction and native checkpoints}
\label{app:broadcast-source-conventions}
\begin{proposition}[Transport of a normalized batch extraction]
\label[proposition]{prop:batch-transport}
Let $\mathfrak B$ be a finite batch space and $P_a^{\rm tr},P_b^{\rm tr}$ two batch laws with identical positive support. For each batch $\mathcal B$, let $\widehat\mu_{\mathcal B}(s)\ge0$ be an extraction rule with $\sum_s\widehat\mu_{\mathcal B}(s)=1$, determined only by the batch text. Set $\mu_t(s)=\Ex_{P_t^{\rm tr}}\widehat\mu_{\mathcal B}(s)$ and $\Lambda_{a:b}(\mathcal B)=\log[P_b^{\rm tr}(\mathcal B)/P_a^{\rm tr}(\mathcal B)]$. Then
\begin{equation}
 \mu_b(s)=\Ex_{P_a^{\rm tr}}\!\bigl[\widehat\mu_{\mathcal B}(s)e^{\Lambda_{a:b}(\mathcal B)}\bigr],
 \qquad \Ex_{P_a^{\rm tr}}e^{\Lambda_{a:b}(\mathcal B)}=1.
 \label{eq:broadcast-batch-transport}
\end{equation}
\end{proposition}
\begin{proof}
Expand the first expectation and cancel its positive probability factor:
\[
 \sum_{\mathcal B}P_a^{\rm tr}(\mathcal B)
 \widehat\mu_{\mathcal B}(s)\frac{P_b^{\rm tr}(\mathcal B)}{P_a^{\rm tr}(\mathcal B)}
 =\sum_{\mathcal B}P_b^{\rm tr}(\mathcal B)\widehat\mu_{\mathcal B}(s)=\mu_b(s).
\]
Taking $\widehat\mu_{\mathcal B}(s)$ out of this calculation and replacing it by one gives $\sum_{\mathcal B}P_b^{\rm tr}(\mathcal B)=1$. If the batch responses are independent, its probability is their product, so its log likelihood ratio is the sum of their complete-trajectory log ratios. Conditional independence given a fixed shared prompt schedule gives the same factorization conditional on that schedule.
\end{proof}

For example, if $N_s(\mathcal B)$ is the eligible count at $s$, batch normalization gives $\widehat\mu_{\mathcal B}(s)=N_s(\mathcal B)/\sum_vN_v(\mathcal B)$ when the denominator is positive. Its expectation is generally not $\Ex N_s/\Ex\sum_vN_v$. Thus the random normalization must stay inside \cref{eq:broadcast-batch-transport}; no additional ratio of expectations is introduced. A rule for batches with no eligible positions must be specified before defining a normalized extraction. If extraction also depends on the learner after fixing the text, transport uses the endpoint rule $\widehat\mu_{b,\mathcal B}$, not an unchanged rule from time $a$.

\paragraph{Native checkpoint identity.}
For discrete checkpoints $k_1<k_2$, assume that every token in the chosen prefix has positive sampler probability at every intermediate checkpoint. Then
\[
 \begin{aligned}
 \Lambda_{k_1:k_2}(s)
 &:=\sum_{k=k_1}^{k_2-1}\sum_{j<h(s)}
       \log\frac{\pi_{k+1}^{\rm tr}(a_j\mid s_j)}{\pi_k^{\rm tr}(a_j\mid s_j)}\\
 &=\sum_{j<h(s)}\log\frac{\pi_{k_2}^{\rm tr}(a_j\mid s_j)}{\pi_{k_1}^{\rm tr}(a_j\mid s_j)}
 =\log\frac{d_{k_2}^{\rm tr}(s)}{d_{k_1}^{\rm tr}(s)}.
 \end{aligned}
\]
The first equality telescopes the finite sum over checkpoints for each token; the second multiplies the ancestral probability factors and cancels the fixed prompt law. The normalizer calculation in \cref{prop:source-reweighting} then applies unchanged on a fixed eligible support. The identity holds for any optimizer and any learning rate. If a token has zero probability at an intermediate checkpoint, the per-step logarithms above need not exist even when both endpoint probabilities are positive; the endpoint ratio can still be used directly where it is defined.

\subsubsection{Changing support and nonsmooth readouts}
\label{app:transport}
\begin{proposition}[Explicit support accounting]
\label[proposition]{prop:support-accounting}
Let $P_a,P_b$ be probability laws on a common finite universe, let $\mathcal C=\operatorname{supp}P_a\cap\operatorname{supp}P_b$, and let $F$ be a bounded real function. Define $e^{\Lambda(v)}=P_b(v)/P_a(v)$ only for $v\in\mathcal C$. Then
\begin{align*}
 \Ex_{P_b}F-\Ex_{P_a}F
 &=\sum_{v\in\mathcal C}P_a(v)F(v)(e^{\Lambda(v)}-1)\\
 &\quad+\sum_{v\in\operatorname{supp}P_b\setminus\mathcal C}P_b(v)F(v)
       -\sum_{v\in\operatorname{supp}P_a\setminus\mathcal C}P_a(v)F(v).
\end{align*}
The absolute value of the last two terms together is at most
$\|F\|_\infty[P_b(\mathcal C^c)+P_a(\mathcal C^c)]$.
\end{proposition}
\begin{proof}
Split each expectation into its contribution on $\mathcal C$ and on its complement. On $\mathcal C$, replace $P_b(v)$ by $P_a(v)e^{\Lambda(v)}$ and subtract. This gives the identity. Applying the triangle inequality and $|F(v)|\le\|F\|_\infty$ to the remaining sums gives the bound.
\end{proof}

In particular, $\sum_{v\in\mathcal C}P_a(v)e^{\Lambda(v)}=P_b(\mathcal C)$, which need not be one. Common-support ratios do not recover newly created mass or, by themselves, the full normalization of a changed source. The proposition applies to complete-trajectory laws, batch laws, or normalized source laws. Changed prompt sets and masks require the same care.

The local top-16 loss in \cref{app:rd-protocol} can change its expression when the selected support changes. If a readout path has finitely many such changes and is piecewise $C^1$ with one-sided limits, integration over each smooth piece gives
\[
 J(b)-J(a)=\int_a^b\dot J(t)\,dt+
       \sum_{t_i\in(a,b)}[J(t_i^+)-J(t_i^-)],
\]
with endpoints taken on their respective one-sided branches. The formula follows by telescoping the integrals and inserting the jumps; a continuous branch switch has zero jump. Smooth-flow integrals are asserted only under this regularity; the finite four-cell identity below remains valid whenever its four scores are finite. For an exogenously fixed source, the response and outcome laws still hold, while the source's own generating rule replaces the on-policy source-tilt formula.

\subsection{Exact symmetric endpoint accounting and its path limit}
\label{app:rd-theory}
\textbf{Connection to \cref{sec:broadcast}, \cref{eq:rd-estimators}.} The four-cell estimator is exactly additive at finite endpoints; the difference between its individual components and the continuous routing and drift integrals is computed and bounded in \cref{prop:endpoint-path-limit}.

\begin{proposition}[Symmetric endpoint accounting]
\label[proposition]{prop:rd-crossing}
Let $\nu_a,\nu_b$ be any two probability vectors on the union of two finite rollout banks, and let $Y_a,Y_b$ be finite readouts defined on that union. With $M_{ij}=\sum_x\nu_i(x)Y_j(x)$, define
\begin{equation}
 \widehat R_{a:b}=\tfrac12[(M_{ba}-M_{aa})+(M_{bb}-M_{ab})],\qquad
 \widehat D_{a:b}=\tfrac12[(M_{ab}-M_{aa})+(M_{bb}-M_{ba})].
 \label{eq:rd-crossing}
\end{equation}
Using $\langle\nu,Y\rangle:=\sum_x\nu(x)Y(x)$ for the distribution--readout pairing,
\[
 \widehat R_{a:b}=\left\langle\nu_b-\nu_a,\frac{Y_a+Y_b}{2}\right\rangle,
 \qquad
 \widehat D_{a:b}=\left\langle\frac{\nu_a+\nu_b}{2},Y_b-Y_a\right\rangle,
\]
and $\widehat R_{a:b}+\widehat D_{a:b}=M_{bb}-M_{aa}$. 
\end{proposition}
\begin{remark}
The inner products can be intuitively interpreted: for routing $\widehat R_{a:b}$, we measure the effect of changing the rollout distribution $v_t$ while holding the readout to the averaged $\frac{Y_a + Y_b}{2}$; the drift estimator holds the rollout distribution averaged while varying the readouts (how much fitting happened over the period $(a,b)$).
\end{remark}
\begin{proof}
Set $\Delta\nu=\nu_b-\nu_a$ and $\Delta Y=Y_b-Y_a$. Bilinearity gives
\[
 M_{bb}-M_{aa}
 =\langle\Delta\nu,Y_a\rangle+\langle\nu_a,\Delta Y\rangle
  +\langle\Delta\nu,\Delta Y\rangle.
\]
Expanding the four terms in \cref{eq:rd-crossing} yields
\[
 \widehat R_{a:b}=\langle\Delta\nu,Y_a\rangle+\tfrac12\langle\Delta\nu,\Delta Y\rangle,
 \qquad
 \widehat D_{a:b}=\langle\nu_a,\Delta Y\rangle+\tfrac12\langle\Delta\nu,\Delta Y\rangle.
\]

\end{proof}

\begin{proposition}[Difference from the path-integrated channels]
\label[proposition]{prop:endpoint-path-limit}
Suppose $\nu_t,Y_t$ are $C^1$ on time interval $[a,b]$, and write
$R_{a:b}=\int_a^b\langle\dot\nu_t,Y_t\rangle\,dt$ and
$D_{a:b}=\int_a^b\langle\nu_t,\dot Y_t\rangle\,dt$.
Then
\[
 \widehat R_{a:b}-R_{a:b}
 =\int_a^b\left\langle\dot\nu_t,\frac{Y_a+Y_b}{2}-Y_t\right\rangle dt
 =-(\widehat D_{a:b}-D_{a:b}).
\]
If $L_\nu=\sup_t\|\dot\nu_t\|_1$ and $L_Y=\sup_t\|\dot Y_t\|_\infty$, each channel error is at most $L_\nu L_Y(b-a)^2/2$. If both functions are $C^2$, define $M_\nu=\sup_t\|\ddot\nu_t\|_1$ and $M_Y=\sup_t\|\ddot Y_t\|_\infty$; the sharper bound is
\[
 |\widehat R_{a:b}-R_{a:b}|=|\widehat D_{a:b}-D_{a:b}|
 \le\frac{(b-a)^3}{12}(L_\nu M_Y+M_\nu L_Y).
\]
\end{proposition}
\begin{proof}
Substitute $\nu_b-\nu_a=\int_a^b\dot\nu_t\,dt$ in the pairing for $\widehat R$ and subtract $R_{a:b}$. Since the sums of the two channels both equal $J_Y(b)-J_Y(a)$, their errors have opposite signs. The fundamental theorem of calculus gives
\[
 \left\|\frac{Y_a+Y_b}{2}-Y_t\right\|_\infty
 \le\tfrac12L_Y[(t-a)+(b-t)]=\tfrac12L_Y(b-a).
\]
Applying $|\langle v,y\rangle|\le\|v\|_1\|y\|_\infty$ and integrating proves the $C^1$ bound.

For the sharper result, insert
$\frac12(Y_a+Y_b)-Y_t=\frac12(\int_t^b\dot Y_r\,dr-\int_a^t\dot Y_r\,dr)$
into the exact error. Reordering these finite-dimensional integrals gives
\[
 \widehat R-R=\tfrac12\int_{a\le r<t\le b}
 \bigl[\langle\dot\nu_r,\dot Y_t\rangle-
       \langle\dot\nu_t,\dot Y_r\rangle\bigr] \,dr\,dt.
\]
Add and subtract $\langle\dot\nu_r,\dot Y_r\rangle$ in the integrand. Its absolute value is bounded by
$L_\nu\|\dot Y_t-\dot Y_r\|_\infty+
 \|\dot\nu_t-\dot\nu_r\|_1L_Y
\le(L_\nu M_Y+M_\nu L_Y)(t-r)$.
Finally $\int_a^b\int_a^t(t-r)\,dr\,dt=(b-a)^3/6$, which proves the coefficient $1/12$.
\end{proof}

For a partition of a fixed interval with largest width $h$, the finite totals telescope exactly. Summing the $C^1$ bound gives a channel error at most $L_\nu L_Y(b-a)h/2$, tending to zero as $h\to0$; the $C^2$ bound gives order $h^2$ instead.

\paragraph{Scope.}
The path limit is a population statement; finite-sample differences between empirical banks, checkpoint substitutions, prompt-composition changes, and support-switch jumps lie outside it, and \cref{app:rd-protocol,app:rd-sensitivity} give the corresponding experimental controls. At a fixed finite partition the allocation is a symmetric endpoint convention.

\subsection{Continuation values and completed outcomes}
\label{app:outcome-proof}
\textbf{Connection to \cref{sec:broadcast}, \cref{eq:broadcast-continuations}.} Recall that a terminal score is a fixed function of a completed rollout. Now, fix a bounded, model-parameter-independent function $Y:\mathcal T\to\mathbb R$. Write $J_Y(t)=\Ex_{\Pi_t}Y$ for the expectation of the terminal score over the time-$t$ generative distribution. For a non-terminal prefix $x$, define
\[
 V_t^Y(x)=\Ex_t[Y\mid x\text{ is reached}],\qquad
 Q_t^Y(x,a)=V_t^Y(x\oplus a),\qquad
 \mathcal A_t^Y(x,a)=Q_t^Y(x,a)-V_t^Y(x).
\]
For terminal $x$, we simply adopt the convention of $V_t^Y(x)=Y(x)$. These are finite conditional expectations under the student's current sampler with its remaining generation budget. Positivity on the fixed allowed autoregressive tree makes them well-defined. Conditioning on the next action gives
$V_t^Y(x)=\sum_a\pi_t(a\mid x)Q_t^Y(x,a)$ and
$\sum_a\pi_t(a\mid x)\mathcal A_t^Y(x,a)=0$.

\begin{proposition}[Recombining exposure and continuation change]
\label[proposition]{prop:rd-value-cut}
For an integer $0\le h\le H_{\max}$, let $\mathcal C_h$ contain every allowed node at depth $h$ and every terminal node of depth less than $h$. Then
$\sum_{x\in\mathcal C_h}d_t(x)=1$, and
\begin{equation}
 \begin{aligned}
 J_Y(t)&=\sum_{x\in\mathcal C_h}d_t(x)V_t^Y(x),\\
 \dot J_Y(t)&=\sum_{x\in\mathcal C_h}\dot d_t(x)V_t^Y(x)
            +\sum_{x\in\mathcal C_h}d_t(x)\dot V_t^Y(x).
 \end{aligned}
 \label{eq:rd-value-cut}
\end{equation}
The first derivative term is the sum of the outcome contributions from decisions of depth less than $h$; the second is the sum from decisions of depth at least $h$. In particular, along any path satisfying the regularity conditions,
\begin{equation}
 \dot J_Y(t)=\sum_{x\in\mathcal S}d_t(x)
          \Ex_{a\sim\pi_t(\cdot\mid x)}[\zeta_t(x,a)\mathcal A_t^Y(x,a)].
 \label{eq:rd-value-rate}
\end{equation}
Under raw flow this equals
\[
 \frac1\tau\sum_{x\in\mathcal S}\sum_s d_t(x)
 [\pi_t\odot\mathcal A_t^Y](x)^\top K_t(x,s)\Gamma_t(s).
\]
Integrating from $t_\star$ to $T$ yields \cref{eq:broadcast-continuations} in \cref{sec:broadcast}. If training and evaluation samplers differ, $d_t,\pi_t,\tau$ in this expression are the evaluation quantities, while $\Gamma_t$ uses the training source and field.
\end{proposition}
\begin{proof}
\emph{Finite conditional values and the cut.}
Every trajectory ends by depth $H_{\max}$. It either reaches depth $h$ or terminates earlier, and therefore meets exactly one element of $\mathcal C_h$. The corresponding reach events are disjoint and exhaustive. Their probabilities sum to one, and conditioning the terminal score on this partition gives the first equality in \cref{eq:rd-value-cut}. At a nonterminal node, $V_t^Y(x)$ is a finite sum of terminal labels multiplied by products of suffix transition probabilities; at a terminal node it is the fixed label. Thus these values are continuous and piecewise $C^1$. The product rule proves the second equality.

\emph{Differentiating a continuation.}
For each nonterminal $x$, set
\[
 \mathfrak h_t(x)=\sum_a\pi_t(a\mid x)\zeta_t(x,a)\mathcal A_t^Y(x,a).
\]
Differentiate the finite recursion $V_t^Y(x)=\sum_a\pi_t(a\mid x)V_t^Y(x\oplus a)$. By \cref{lem:centering}, $\dot\pi_t=\pi_t\zeta_t$ and $\sum_a\pi_t\zeta_t=0$, so
\[
 \dot V_t^Y(x)=\mathfrak h_t(x)+
             \sum_a\pi_t(a\mid x)\dot V_t^Y(x\oplus a).
 \tag{*}
\]
At a terminal child the last derivative is zero. For a descendant $y$ of $x$, let $d_t(y\mid x)$ be the product of transition probabilities from $x$ to $y$, with $d_t(x\mid x)=1$. Induction on the remaining depth proves
\[
 \dot V_t^Y(x)=\sum_{\substack{y\in\mathcal S\,:\,y\succeq x}}
                     d_t(y\mid x)\mathfrak h_t(y).
 \tag{**}
\]
Indeed, if all children are terminal, (*) is precisely (**). Otherwise insert (**) for each nonterminal child into (*). Its descendant subtrees are disjoint, and multiplying a child's descendant reach probability by $\pi_t(a\mid x)$ gives $d_t(y\mid x)$. This proves the induction step and counts every remaining decision exactly once.

\emph{Decisions after the cut.}
For a nonterminal $y$ of depth at least $h$, the ancestry has a unique cut ancestor $x\in\mathcal C_h$. Also $d_t(x)d_t(y\mid x)=d_t(y)$. Multiplying (**) by $d_t(x)$ and summing over the cut therefore gives
\[
 \sum_{x\in\mathcal C_h}d_t(x)\dot V_t^Y(x)
 =\sum_{\substack{y\in\mathcal S\,:\,h(y)\ge h}}d_t(y)\mathfrak h_t(y).
\]
Terminal cut nodes have no descendants with decisions and contribute zero.

\emph{Decisions before the cut.}
The reach derivative at a cut node is $\dot d_t(c)=d_t(c)\sum_{j<h(c)}\zeta_t(c_j,a_j)$. Reorder the finite sum $\sum_{c\in\mathcal C_h}\dot d_t(c)V_t^Y(c)$ by each ancestral decision $(x,a)$ with $h(x)<h$. The coefficient of $\zeta_t(x,a)$ is
\[
 \sum_{\substack{c\in\mathcal C_h\,:\,c\succeq x\oplus a}}d_t(c)V_t^Y(c)
 =d_t(x)\pi_t(a\mid x)Q_t^Y(x,a).
\]
To verify this equality, condition on reaching $x$ and selecting $a$. The subsequent cut nodes form a disjoint, exhaustive partition of that conditional event, and the conditional expectation of their values is $Q_t^Y(x,a)$. This includes a terminal child $x\oplus a$, which is itself in the cut. Summing over $a$ and subtracting $V_t^Y(x)\sum_a\pi_t(a\mid x)\zeta_t(x,a)=0$ gives
\[
 \sum_{c\in\mathcal C_h}\dot d_t(c)V_t^Y(c)
 =\sum_{\substack{x\in\mathcal S\,:\,h(x)<h}}d_t(x)\mathfrak h_t(x).
\]
Combining the two depth ranges proves \cref{eq:rd-value-rate}, including the boundary cases $h=0$ and $h=H_{\max}$.

\emph{Kernel form and integration.}
Let $a_t^Y(x)=[\pi_t\odot\mathcal A_t^Y](x)$, extended by zero outside the allowed tokens. Its coordinates sum to zero. Substituting $\zeta_t=\tau^{-1}(e_a-\pi_t)^\top v_t$ gives
\[
 \mathfrak h_t(x)
 =\tau^{-1}\left[a_t^Y(x)^\top v_t(x)
  -\Bigl(\sum_a\pi_t(a\mid x)\mathcal A_t^Y(x,a)\Bigr)\pi_t^\top v_t(x)\right]
 =\tau^{-1}a_t^Y(x)^\top v_t(x).
\]
Under raw flow, $v_t(x)=\sum_sK_t(x,s)\Gamma_t(s)$, proving the bilinear formula. For an actual optimizer path, the same calculation uses $v_t(x)=\Phi_t(x)\dot\theta_t$. Integrating the continuous, piecewise differentiable $J_Y$ proves the accumulated outcome formula.
\end{proof}

The suffix derivative in (**) already includes changes to every downstream prediction. Adding a second derivative of $Q_t^Y$ to the full decision sum in \cref{eq:rd-value-rate} would just double-count that motion. Conversely, at a fixed cut prefix, $\dot V_t^Y(x)$ \emph{cannot} generally be replaced by a derivative with respect to the logits at $x$ alone. Local entropy and terminal correctness are very different observables even though their changes follow the same underlying parameter path.

Reordering the finite source-receiver sum gives the source-resolved form
\[
 \dot J_Y(t)=\sum_s\mu_t(s)\varphi_{Y,t}(s),\qquad
 \varphi_{Y,t}(s)=\frac1\tau\sum_xd_t(x)
 [\pi_t\odot\mathcal A_t^Y](x)^\top K_t(x,s)B_t(s).
\]
A source contributes through every coupled receiver, not just through its own continuation value. The experiments in \cref{app:rd} therefore combine prefix dynamics with held-out outcomes.

\subsection{Entropy sensitivity and ancestral exposure}
\label{app:rd-entropy-response}
\textbf{Connection to \cref{sec:broadcast}, the entropy-versus-fitting argument.} In this subsection $H_t(x)=-\sum_a p_t(a\mid x)\log p_t(a\mid x)$ is full-vocabulary model entropy. Its policy is $p_t=\softmax(z_t)$ at temperature one; the sampling policy in the routing term remains $\pi_t$ at temperature $\tau$. Fixed-bank evidence is in \cref{app:rd-entropy-evidence}.

\begin{proposition}[Entropy sensitivity]
\label[proposition]{prop:rd-entropy-response}
Under \cref{eq:rd-observation},
\begin{equation}
 \begin{aligned}
 R_H(t)&=\operatorname{Cov}_{\nu_t}\!\left(H_t(x),\sum_{j<h(x)}\zeta_t(x_j,a_j)\right),\\
 D_H(t)&=\Ex_{\nu_t}[\chi_{H,t}(x)^\top v_t(x)],\\
 \chi_{H,t}(x)&:=\nabla_zH_t(x)
 =-p_t(\cdot\mid x)\odot[\log p_t(\cdot\mid x)+H_t(x)\mathbf1].
 \end{aligned}
 \label{eq:rd-entropy-sensitivity}
\end{equation}
At a fixed receiver,
\begin{equation}
 \dot H_t(x)=-\operatorname{Cov}_{a\sim p_t(\cdot\mid x)}(\log p_t(a\mid x),v_t(x,a)),
 \qquad
 |\dot H_t(x)|\le\sqrt{\operatorname{Var}_{p_t}(\log p_t)\operatorname{Var}_{p_t}(v_t)}.
 \label{eq:rd-entropy-covariance}
\end{equation}
The parameter-space entropy gradient is $\Phi_t(x)^\top\chi_{H,t}(x)$. These identities hold with the actual logit velocity; the raw specialization is $v_t=K_t\Gamma_t$.
\end{proposition}
\begin{proof}
Fix $x$ and suppress it in the notation. Softmax differentiation gives
$\dot p_a=p_a(v_a-\bar v_p)$, where $\bar v_p=\sum_a p_av_a$. Since $\sum_a\dot p_a=0$,
\begin{align*}
 \dot H&=-\sum_a\dot p_a(\log p_a+1)
        =-\sum_a p_a(v_a-\bar v_p)\log p_a\\
       &=-\operatorname{Cov}_p(\log p,v)
        =-\sum_a p_a(\log p_a+H)v_a.
\end{align*}
The last equality uses $\sum_a p_a\log p_a=-H$. Since the directional identity holds for every logit direction $v$, its coefficient vector is $\chi_H=\nabla_zH$. Weighted Cauchy--Schwarz applied to the centered vectors $\log p-\Ex_p\log p$ and $v-\Ex_pv$ proves the variance bound. The chain rule gives the parameter gradient $\Phi^\top\chi_H$. Finally substitute the local derivative into \cref{prop:rd-law}; its ancestral score gives the stated $R_H$ and its fixed-prefix derivative gives $D_H$.
\end{proof}

\begin{proposition}[Ancestral exposure without first-order entropy increase]
\label[proposition]{prop:stop-entropy}
Consider a fixed path of $h\ge1$ binary continue/stop decisions. At every gate, let the conditional continuation probability be a differentiable $c_t\in(0,1)$, and hold the suffix executor fixed. Conditional on the path's starting node, its end is reached with probability $c_t^h$. With $H_{\rm bin}(c)=-c\log c-(1-c)\log(1-c)$,
\begin{equation}
 \partial_t\log\Pr_t(\text{reach depth }h)=h\dot c_t/c_t,
 \qquad \partial_tH_{\rm bin}(c_t)=\dot c_t\log[(1-c_t)/c_t].
 \label{eq:rd-stop-entropy-example}
\end{equation}
\end{proposition}
\begin{proof}
The probability of reaching the end is the product of the $h$ conditional gate probabilities, regardless of any independence interpretation. Its logarithm is $h\log c_t$, giving the first derivative. Direct differentiation yields $H_{\rm bin}'(c)=-\log c-1+\log(1-c)+1=\log[(1-c)/c]$; the chain rule gives the second.
\end{proof}
For example, at $c_t=1/2$ with $\dot c_t>0$, the reach derivative is positive and each gate's entropy derivative is zero. Thus a pre-existing uncertain suffix can receive more exposure in the rollout distribution \emph{without} its predictions becoming flatter.

\paragraph{Checking propensity and teacher-relative discrepancy.}
\label{app:checking-sensitivity}
For the fixed WAIT-initiated family $\mathcal W$, let $w_t=p_t(\mathcal W\mid x)\in(0,1)$ at a fixed prefix, $b_t=\log[w_t/(1-w_t)]$, and $q_W=q(\mathcal W\mid x)$ be fixed. Define
\[
 c_W(t)=D_{\rm KL}(\operatorname{Ber}(q_W)\Vert\operatorname{Ber}(w_t)).
\]
With the convention $0\log0=0$ at $q_W\in\{0,1\}$,
\[
 \partial_t c_W=(w_t-q_W)\partial_t b_t.
\]
Indeed, writing out the two Bernoulli terms gives
$\partial_wc_W=-q_W/w+(1-q_W)/(1-w)=(w-q_W)/[w(1-w)]$.
Also $\partial_t b_t=\dot w_t/[w_t(1-w_t)]$. Multiplication proves the identity, including the endpoint teacher probabilities by the same formula. A decrease in propensity is corrective exactly when its product with the current signed teacher discrepancy is negative; an initially above-teacher class need not remain above the teacher throughout training.

\FloatBarrier
\Needspace{12\baselineskip}
\section{Conversion: protocols and additional measurements}
\label{app:collective}\label{app:collective-data}
This appendix gives the experimental details behind \cref{sec:local}. We first describe the model pairs and the fixed prefixes used for measurement. We then ask where the teachers place their strongest requests, how well those requests produce parameter updates, and how the student's responsiveness changes during training. The derivations are in \cref{app:collective-proof,app:collective-frozen}.

\subsection{Setup}
\label{app:collective-protocol}
We compare two teachers for the same starting student at each scale. At 1.5B, the student is DeepSeek-R1-Distill-Qwen-1.5B and the teachers are JustRL-1.5B and DeepSeek-R1-Distill-Qwen-7B. At 7B, the student is DeepSeek-R1-Distill-Qwen-7B and the teachers are AceReason-Nemotron-7B and DeepSeek-R1-Distill-Qwen-14B. The AIME comparison uses 30 AIME 2024 problems at temperature $.6$. We draw 16 responses per problem at 1.5B and 4 at 7B.

At each scale, eleven rollouts from the initial student provide the prefixes for measurement. Each local kernel is computed from logit gradients with respect to all student parameters. Please note that, computing the gradients is expensive, the eigendecomposition itself is inexpensive. The kernel measurements therefore use sampled positions, while $\Xi$ serves as a less expensive monitor.

\subsection{Where the strongest requests occur}
\label{app:collective-types}
We analyze which kinds of positions receive the strongest teacher requests. The annotated rollouts contain 36,416 positions at 1.5B and 38,168 at 7B. For each teacher, the strongest $1\%$ contains 364 and 381 positions, respectively. 

\paragraph{Annotations and enrichment.}\label{app:semantic-protocol}
We use Opus 5 to label the sentences in the rollouts. Each sentence receives a semantic category, which describes its role in the reasoning. We also classify positions by their location within a sentence and by token kind. These are three separate classifications of the same positions, examined from different semantic aspects. We use the following prompt to create the semantic labels (this rubric is the result of iterating on the model's reasoning traces a few times, to settle on the most prevalent reasoning categories the model writes in its chain of thought).

\begin{promptbox}{LLM Judge Prompt}
\begin{Verbatim}[
    breaklines=true,
    breakanywhere=true,
    fontsize=\small,
    breaksymbolleft={}
]
You are annotating every sentence of a model's reasoning trace (for example, a model solving a math problem) with a label describing the act the sentence performs.

REASONING TRACE: "{reasoning_trace}"

Emit per sentence: {"sent_idx": int, "fine": [1 label], "uncertain": bool, "note": "..."} Read the FULL trace first; judge each sentence's FUNCTION IN CONTEXT.

## Fine labels

CONTROL family (steering the reasoning process):
- check-initiate: starting a verification act on ALREADY-ESTABLISHED material -- re-computing a prior result, re-reading the problem statement, re-deriving an earlier step. First-time testing of a fresh candidate or case is NOT check-initiate (see TECHNICAL).
- check-conclude: verdict closing a check episode. Covers: work-correctness verdicts ("Yes, that's correct", "that step was wrong"); verdicts about the check itself, including redundancy ("I already did this case", "we've already considered that"); resolutions of a problem re-read that settle interpretation ("So it's allowed to be prime").
- doubt-flag: hesitation OR bare deliberation cue WITHOUT an act ("Wait.", "Hmm, that seems too big.", "Let me think.", "Let's see."). Judged sentence-locally: a bare cue stays doubt-flag even when a check follows in the next sentence. Never label bare cues as plan.
- plan: stating an approach or opening a new sub-task, with the approach/task STATED ("I'll use coordinates", new-sub-task colon lead-ins like "Compute the exponent:"). Includes proceed-decisions ("Let me try that").
- reframe: abandoning/switching approach. NOT routine next-case moves in an enumeration ("Alternatively, x=2" = TECHNICAL).
- repair: announcing/making a correction of an earlier error.
- recap: MULTI-ITEM consolidation of established state used to steer. Single value-echoes = compute/derive.
- commit: FINAL-ANSWER commitment only ("So the answer is 34", "Answer: 34").

TECHNICAL family (object-level mathematics):
- derive: algebraic/logical derivation steps; this includes stating the associated problem facts, notation, coordinates, case setup, and within-computation connective lead-ins (including "Simplify:", "Comparing coefficients:").
- recall: invoking a theorem/definition/known mathematical fact.

OTHER family:
- style: pure filler/transition with no content and no steering.
- format: formatting artifacts, the end-of-reasoning marker (e.g. </think>), bare list numerals or LaTeX delimiters split onto their own lines (format, not fragment).
- unclear: genuinely unclassifiable.


## Fixed vocabulary and note rules
Use only the labels above; do not propose or introduce new labels. A sentence the rules do not settle or hard to classify takes the best available existing label, "uncertain": true, and leave a short note -- residual uncertainty is data itself, please do not treat it as a defect to legislate away. Furthermore, write in the note for the sentence if it genuinely has more than one of the labels above.
\end{Verbatim}
\end{promptbox}

%We compare the single-label semantic counts with counts that allow multiple labels; every headline enrichment value changes only by at most $0.1$.

For a category $c$, \emph{enrichment} compares its frequency among the teacher's strongest requests with its frequency among all annotated positions:
\[
 E_q(c)=\frac{\Pr(c\mid\text{strongest }1\%\text{ under teacher }q)}{\Pr(c)}.
\]

\begin{figure}[ht]
\centering
\includegraphics[width=\linewidth]{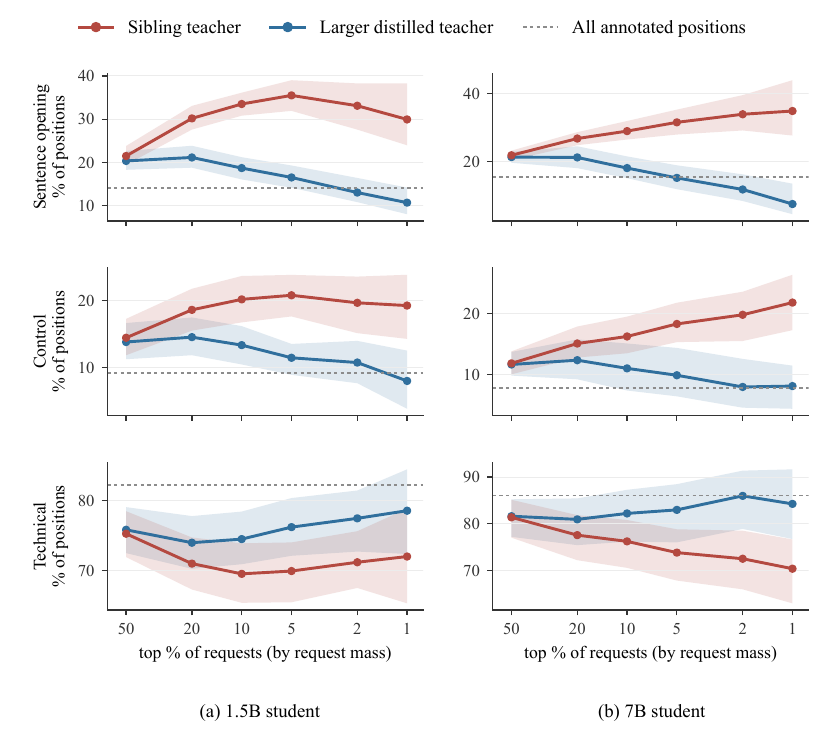}
\caption{\textbf{Composition of the strongest requests as the percentile rises.} Shares of sentence openings (first three tokens), control positions and technical positions among each teacher's top 50\%, 20\%, \ldots, 1\% of requests. Dotted: all annotated positions; bands: 95\% bootstrap intervals over the annotated rollouts.}
\label{fig:request-trends}
\end{figure}

\begin{figure}[ht]
\centering
\includegraphics[width=\linewidth]{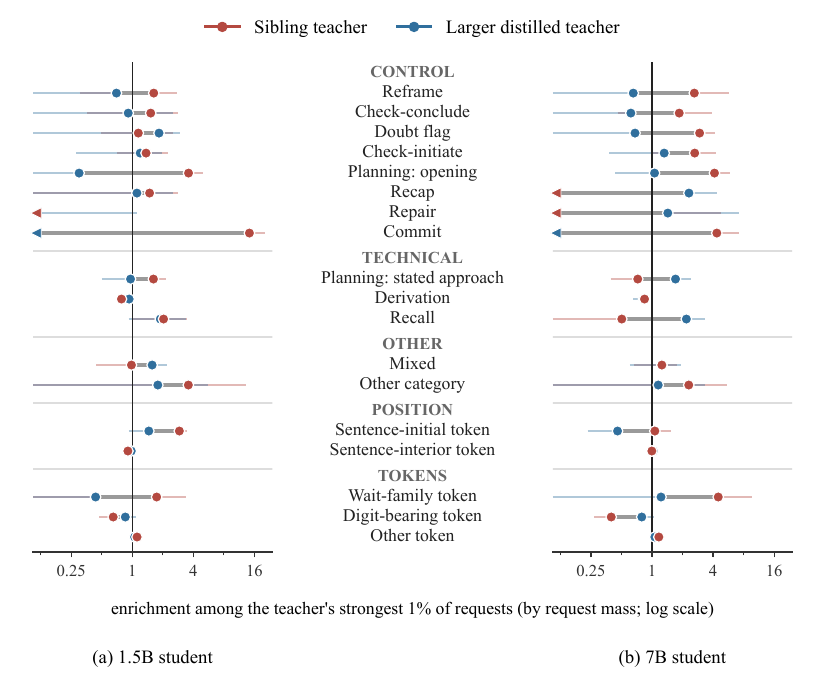}
\caption{\textbf{Where the strongest requests occur.} The full set of semantic categories, sentence positions, and token kinds. A ratio of one means that a category has the same frequency among the strongest requests as in the annotated rollouts. Planning sentences are split into their opening and stated approach. Intervals are clipped at the plot edges; left-pointing markers denote estimates below the plotted range, including zero. The main figure selects twelve rows.}
\label{fig:composition}
\end{figure}

A value above 1 means that the category is more common among the strongest requests. For planning, we split the sentences in this class by position: their first three tokens (the opening of a plan) belong to the control family, and the remaining tokens (the stated approach) belong to the technical family; mixed sentences belong to neither. Panel~(a) of \cref{fig:conversion-tokens} takes each teacher's top 50\%, 20\%, 10\%, 5\%, 2\% and 1\% of requests and shows the share of sentence openings (first three tokens), control positions and technical positions, \emph{minus} the corresponding share among all annotated positions. \cref{fig:request-trends} shows the raw shares. Bootstrap intervals resample the annotated rollouts. They are computed separately for each teacher. Panels~(b, c) of \cref{fig:conversion-tokens} show twelve selected rows, and \cref{fig:composition} shows the full set. \Cref{tab:semantic-enrichment} gives the values used in the main figure.

\paragraph{Differences between teachers.}
At 1.5B, planning openings are enriched $3.59\times$ under the sibling teacher and $0.30\times$ under the larger teacher. Sentence-initial positions are enriched $2.91\times$ and $1.46\times$, respectively. At 7B, the sibling teacher places relatively more of its strongest requests at reframing, checking, and doubt positions. Wait-family tokens are enriched $4.50\times$ under the sibling and $1.23\times$ under the larger teacher.

The difference depends on the category.  At 7B, planning openings are enriched $4.13\times$ and $1.06\times$, while the stated approach of a plan is enriched $0.72\times$ and $1.70\times$; the teachers' derivation enrichment is similar ($0.84\times$ and $0.83\times$). Digit-bearing positions are relatively more prominent under the larger teacher at both scales.

We find that, as the percentile goes up, the sibling teacher's shares of \emph{sentence openings and control positions} grow relative to all positions, while the larger teacher's return to or fall below them (\cref{fig:request-trends}). Among the strongest 1\%, the sibling teacher's share of sentence openings exceeds the larger teacher's by $19.2$ [$13.1$, $27.7$] percentage points at 1.5B and $27.6$ [$18.8$, $37.0$] at 7B.

\begin{table}[ht]
\centering\small
\begin{tabular}{lrrrr}
\toprule
& \multicolumn{2}{c}{1.5B student} & \multicolumn{2}{c}{7B student} \\
\cmidrule(lr){2-3}\cmidrule(lr){4-5}
Position category & Sibling & Larger & Sibling & Larger \\
\midrule
Reframe & 1.63 & 0.70 & 2.61 & 0.65 \\
Check-conclude & 1.52 & 0.91 & 1.86 & 0.62 \\
Doubt flag & 1.15 & 1.84 & 2.95 & 0.68 \\
Check-initiate & 1.37 & 1.20 & 2.64 & 1.32 \\
Planning: opening & 3.59 & 0.30 & 4.13 & 1.06 \\
Recap & 1.48 & 1.11 & 0.00 & 2.31 \\
Planning: stated approach & 1.62 & 0.96 & 0.72 & 1.70 \\
Derivation & 0.78 & 0.93 & 0.84 & 0.83 \\
Recall & 2.04 & 1.89 & 0.50 & 2.18 \\
Mixed & 0.98 & 1.57 & 1.25 & 1.25 \\
\midrule
Wait-family token & 1.74 & 0.43 & 4.50 & 1.23 \\
Digit-bearing token & 0.65 & 0.85 & 0.40 & 0.79 \\
\bottomrule
\end{tabular}
\caption{\textbf{Enrichment values for the main semantic figure.} Each value compares the frequency of a category among the strongest requests with its frequency among all annotated positions. \Cref{fig:composition} shows the bootstrap intervals for each teacher.}
\label{tab:semantic-enrichment}
\end{table}

\FloatBarrier
\Needspace{10\baselineskip}
\subsection{Conversion estimates by semantic class}
\label{app:collective-conventions}
We now ask which factors in \cref{eq:pac} distinguish the healthy and stalled runs. We report pressure $P$, accessibility $A$, and coherence $C$ separately for the \textsc{control} and \textsc{technical} samples selected for each teacher. The initial comparison asks how the same starting student responds to the two teachers. The endpoint comparison asks how much disagreement remains and how the student's responsiveness has changed.

\paragraph{Measurement and estimation.}
We save the training rollouts and the request vector $B$ computed at every position during training. The initial requests $B_0$ used in the frozen-request probes come directly from these records. For a fixed request at prefix $s$, the parameter gradient is $g_t(s)=\Phi_t(s)^\top B(s)$. We obtain it by backpropagating through $B(s)^\top z_t(s)$ while holding $B(s)$ constant.

The squared norm of a sample mean includes sampling variance. We correct for this before estimating coherence. For $n>1$ sampled gradients, write $\bar g=n^{-1}\sum_i g_i$ and $\widehat D_g=n^{-1}\sum_i\|g_i\|^2$. The correction is
\[
 \widehat C_{\mathrm{corr}}=
 \frac{\left(\|\bar g\|^2-
       [\widehat D_g-\|\bar g\|^2]/(n-1)\right)_+}{\widehat D_g},
 \qquad \widehat D_g>0,
\]
where $(r)_+=\max(r,0)$. Under independent sampling, the numerator before truncation is unbiased for the squared norm of the population-mean gradient (\cref{prop:gradient-norm-correction}). 

\begin{table}[htb]
\centering\small\setlength{\tabcolsep}{4.2pt}
\begin{tabular}{llrrrrr}
\toprule
Run & Class & $P_0$ & $A_0/10^3$ & $C_0$ & $\Xi_0/10^3$ & $\|u_0\|$ \\
\midrule
7B healthy & Control & 0.192 & 119.8 & 0.081 & 9.70 & 43.1 \\
7B healthy & Technical & 0.151 & 106.3 & 0.055 & 5.85 & 29.7 \\
7B stalled & Control & 0.521 & 148.1 & 0.025 & 3.70 & 43.9 \\
7B stalled & Technical & 0.599 & 128.8 & 0.020 & 2.58 & 39.3 \\
1.5B healthy & Control & 6.024 & 39.6 & 0.041 & 1.62 & 98.9 \\
1.5B healthy & Technical & 2.569 & 54.5 & 0.064 & 3.49 & 94.7 \\
1.5B stalled & Control & 0.824 & 41.3 & 0.040 & 1.65 & 36.9 \\
1.5B stalled & Technical & 1.110 & 53.8 & 0.039 & 2.10 & 48.2 \\
\bottomrule
\end{tabular}

\caption{\textbf{Initial conversion estimates by semantic class.} Each row uses $n=80$ positions selected for one teacher and class. $P$ and $A$ use measurements at individual positions. $C$ uses the finite-sample correction above. $\Xi$ and $\|u\|$ are approximate values computed from these estimates. Comparisons are within scale and parameterization.}
\label{tab:collective-initial}
\end{table}

\paragraph{Initial comparison.}
\Cref{tab:collective-initial} shows why gradient norm alone does not explain conversion. At 7B, the healthy run has somewhat lower accessibility but $2.75$--$3.24\times$ higher coherence across the two classes. Its collective gain is $2.27$--$2.62\times$ higher. The stalled run still has a comparable or larger gradient norm because its request pressure is higher.

At 1.5B, accessibility is similar between the two teachers, and the coherence difference depends on the class. The initial collective gains are nearly equal in \textsc{control}: $1.62$ versus $1.65$, in units of $10^3$. In \textsc{technical}, the healthy run has the larger gain: $3.49$ versus $2.10$. Thus its larger gradient norms do not imply a larger conversion gain in every class.

\paragraph{Remaining requests and responsiveness.}
\Cref{tab:collective-endpoints} separates two endpoint comparisons. The live conversion factors use the current requests $B_t$. They measure the response to the remaining disagreements with the teacher. The last column instead reapplies the initial requests $B_0$. With both the requests and prefixes fixed, this column measures how local responsiveness to those requests has changed.

\begin{table}[htb]
\centering\footnotesize\setlength{\tabcolsep}{3.2pt}
\begin{tabular}{llrrrrrr}
\toprule
Run & Class & Step & $P_t/P_0$ & $A_t/A_0$ & $C_t/C_0$ & $\Xi_t/\Xi_0$ & $A_t[B_0]/A_0[B_0]$ \\
\midrule
7B healthy & Control & 200 & 0.005 & 0.983 & 0.012 & 0.012 & 1.414 \\
7B healthy & Technical & 200 & 0.007 & 2.159 & 0.036 & 0.079 & 1.232 \\
7B stalled & Control & 150 & 0.740 & 0.827 & 0.520 & 0.430 & 0.810 \\
7B stalled & Technical & 150 & 0.740 & 0.884 & 0.500 & 0.442 & 0.889 \\
1.5B healthy & Control & 200 & 0.009 & 1.216 & 0.341 & 0.415 & 2.829 \\
1.5B healthy & Technical & 200 & 0.015 & 3.628 & 0.031 & 0.113 & 5.399 \\
1.5B stalled & Control & 200 & 0.602 & 0.836 & 0.725 & 0.606 & 0.831 \\
1.5B stalled & Technical & 200 & 0.790 & 0.793 & 0.564 & 0.447 & 0.789 \\
\bottomrule
\end{tabular}

\caption{\textbf{Remaining requests and responsiveness at the endpoint.} The positions are the same as at initialization. The pressure and live conversion columns use the current requests $B_t$; the last column uses the initial requests $B_0$. All diagnostic columns are ratios to their initial values. Endpoints are step 200, except for stalled 7B at step 150. Coherence estimates are rounded, especially at healthy endpoints.}
\label{tab:collective-endpoints}
\end{table}

The healthy runs retain only $0.5$--$1.5\%$ of their initial request energy. The stalled runs retain $60.2$--$79.0\%$. Meanwhile, accessibility to the initial requests increases in both classes of each healthy run and decreases in both classes of each stalled run.

\FloatBarrier
\Needspace{10\baselineskip}
\subsection{How the local kernel changes}
\label{app:collective-geometry}
To understand the change in responsiveness, we measure both the strength and movement of the kernel's leading directions. At each prefix $s$, let $V_{t,3}(s)$ contain the orthonormal eigenvectors of the symmetrized local kernel $K_t(s,s)$ with the three largest eigenvalues. Denote those eigenvalues by $\{\lambda_{j,t}(s)\}_{j=1}^3$. Main Figure~\ref{fig:conversion-geometry} plots
\[
 G_3(s,t)=\frac{\sum_{j=1}^3\lambda_{j,t}(s)}{\sum_{j=1}^3\lambda_{j,0}(s)},
 \qquad
 O_3(s,t)=\frac{\|V_{0,3}(s)^\top V_{t,3}(s)\|_F^2}{3}.
\]
$G_3$ measures the change in the sum of the three largest eigenvalues. $O_3$ measures how much their subspace overlaps with its initial position. It is the mean squared cosine of the three principal angles: $1$ means identical subspaces and $0$ means orthogonal subspaces. The main figure reports median $G_3$ and $O_3$ over 160 positions; the table below reports $1-O_3$ instead. Moreover, subspace movement alone does not tell us whether the kernel becomes better or worse aligned with the teacher's requests. For this, we measure the share of the initial request in the leading subspace, $w_{1:3}(s,t)=\sum_{j\leq3}|v_{j,t}^\top b_0|^2$, where $b_0$ is the unit direction of the initial request. Its change is $\Delta w_{1:3}$. \Cref{fig:conversion-turn} compares this change for the two teachers' requests in the same checkpoint kernel.

\begin{table}[htb]
\centering\small\setlength{\tabcolsep}{3.4pt}
\begin{tabular}{lrrrrrr}
\toprule
Run & Step & $G_3$ & $1-O_3$ & $\Delta w_{1:3}$ (pp) & Median $a_t/a_0$ & $A_t[B_0]/A_0[B_0]$ \\
\midrule
7B healthy & 50 & 1.19 & 0.034 & -0.23 & 1.04 & 1.23 \\
7B healthy & 200 & 1.25 & 0.041 & -0.36 & 1.07 & 1.34 \\
7B stalled & 50 & 0.79 & 0.018 & +1.12 & 0.93 & 0.90 \\
7B stalled & 150 & 0.74 & 0.032 & +1.51 & 0.91 & 0.85 \\
1.5B healthy & 50 & 1.00 & 0.232 & +5.49 & 1.19 & 3.95 \\
1.5B healthy & 200 & 2.31 & 0.369 & +5.07 & 2.08 & 3.78 \\
1.5B stalled & 50 & 0.80 & 0.022 & +1.06 & 0.91 & 0.86 \\
1.5B stalled & 200 & 0.67 & 0.044 & +0.89 & 0.83 & 0.80 \\
\bottomrule
\end{tabular}

\caption{\textbf{Changes in local geometry.} Except in the last column, ratios and differences are computed at each position before taking the median. The last column compares accessibility averaged across positions with weights proportional to initial request energy. The median and weighted average describe different aspects of the same sample. Here pp denotes percentage points.}
\label{tab:collective-geometry}
\end{table}

Early on in training (by step 50), the healthy 1.5B run has already changed its leading subspace, while the leading strength is near its initial value. The strength increases later. The healthy 7B run mainly increases leading strength, with little subspace movement. Both stalled runs lose leading strength.

The table also distinguishes a typical position from an average weighted by request energy. For example, at step 50 of healthy 1.5B, the median sensitivity ratio is $1.19$, while the ratio of weighted accessibilities is $3.95$. The latter gives more weight to positions with larger initial requests. Their difference reflects the weighting and the way results are combined across positions.

\begin{figure}[htb]
\centering
\begin{subfigure}[t]{.48\linewidth}
\includegraphics[width=\linewidth]{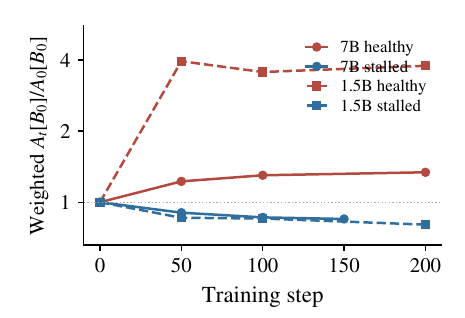}
\caption{Accessibility to the initial requests.}
\end{subfigure}\hfill
\begin{subfigure}[t]{.48\linewidth}
\includegraphics[width=\linewidth]{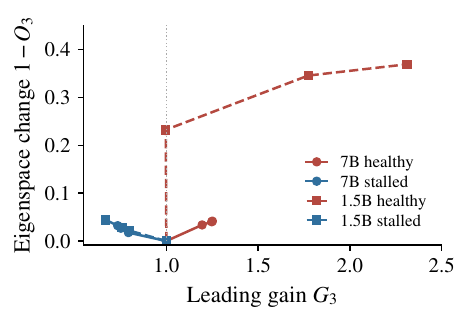}
\caption{Strength and movement of the leading subspace.}
\end{subfigure}
\caption{\textbf{Changes in the kernel and its response to the initial requests.} Left: $A_t[B_0]/A_0[B_0]$ on the combined sample, with equal numbers of positions from the two classes. Right: median leading strength and subspace change across positions. Red/blue denote healthy/stalled runs; circles/squares denote 7B/1.5B.}
\label{fig:collective-supplement}
\end{figure}

\begin{figure}[htb]
\centering
\begin{subfigure}[t]{.48\linewidth}
\includegraphics[width=\linewidth]{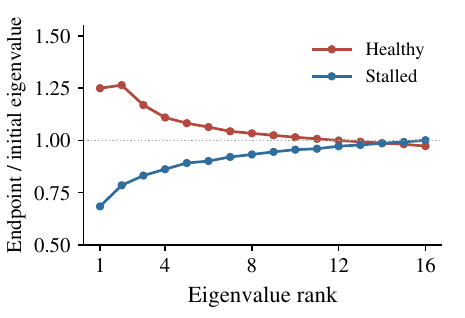}
\caption{7B student.}
\end{subfigure}\hfill
\begin{subfigure}[t]{.48\linewidth}
\includegraphics[width=\linewidth]{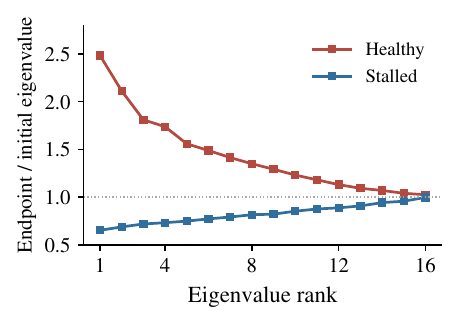}
\caption{1.5B student.}
\end{subfigure}
\caption{\textbf{Endpoint eigenvalues relative to initialization.} At each sorted rank, we compute the endpoint-to-initial ratio at each position and report the median over 160 positions per run. Red/blue denote healthy/stalled runs. Endpoints are step 200, except for stalled 7B at step 150.}
\label{fig:collective-spectra}
\end{figure}

\Cref{fig:collective-spectra} extends the comparison to the full eigenvalue profile. It compares eigenvalues at the same sorted rank across checkpoints. It is clear that the healthy runs experience a lift in the principal eigenvalues as training proceeds, while the stalled runs experience a decline instead.

A separate comparison groups initial directions by their request energy. The reported growth is not concentrated in the groups with stronger requests. We therefore distinguish a turn toward the teacher's requests from preferential growth along the most strongly requested initial directions.

\FloatBarrier
\Needspace{10\baselineskip}
\subsection{Alignment with the teacher}
\label{app:collective-specificity}
The leading subspace rotates in the healthy 1.5B run. Does this movement favor the teacher that trained the student? We compare the training-run teacher's initial requests with the other teacher's initial requests at the same prefixes and token coordinates. Both are evaluated in the same checkpoint kernel. More precisely, at each position, we subtract their changes in projection onto the leading subspace:
\[
 d_s(t)=\Delta w_{1:3}^{\mathrm{trained}}(s,t)
       -\Delta w_{1:3}^{\mathrm{other}}(s,t).
\]
A positive value means that the training teacher's requests gain more projection. \Cref{fig:conversion-turn} uses step 50 for all four runs. The comparison holds the kernel change fixed and asks which field gains more projection.

\begin{table}[htb]
\centering\small\setlength{\tabcolsep}{3.5pt}
\begin{tabular}{lrrrr}
\toprule
Run & Trained gain (pp) & Other gain (pp) & Paired difference (pp) & 95\% interval \\
\midrule
7B healthy & -0.23 & +0.07 & -0.16 & $[-0.33,+0.27]$ \\
7B stalled & +1.12 & +1.25 & +0.00 & $[-0.53,+0.60]$ \\
1.5B healthy & +5.49 & +2.42 & +1.79 & $[+0.94,+3.13]$ \\
1.5B stalled & +1.06 & +1.00 & -0.16 & $[-0.29,+0.05]$ \\
\bottomrule
\end{tabular}

\caption{\textbf{Comparison of the two teachers' requests at step 50.} Values are medians over the same 160 positions. The paired difference is the median of the per-position differences. For each of 3,000 bootstrap samples, we resample the eleven initial rollouts and retain all selected positions in each sampled rollout. The two fields remain paired and the trained checkpoints remain fixed.}
\label{tab:collective-specificity}
\end{table}

The healthy 1.5B run has a median paired difference of $1.79$ percentage points, with a $95\%$ interval of $[0.94,3.13]$. The other three intervals include zero. These measurements show that there is a ``preferential'' turn toward the training teacher's requests in healthy 1.5B. The paired median need not equal the difference between the two separate medians: $5.49-2.42$ and $1.79$ summarize the same positions in different orders. The bootstrap measures variation across the sampled rollouts with the trained checkpoints fixed.

\FloatBarrier
\Needspace{12\baselineskip}
\section{Broadcast: protocols and additional measurements}
\label{app:audit}\label{app:rd}
This appendix gives the experimental details for \cref{sec:broadcast}. We examine how OPD changes prefix visitation and the student's predictions on fixed prefixes. We then ask whether the student can complete its reasoning from those prefixes and reach a correct answer. The theoretical backbone is derived and presented in \cref{app:rd-flow,app:broadcast-prefix-reweighting,app:rd-theory,app:outcome-proof,app:rd-entropy-response}.

\subsection{Models and measurement setup}
\label{app:rd-protocol}

\paragraph{Models and training.}
All three learners use JustRL-1.5B as the teacher. The strong learner is DeepSeek-R1-Distill-Qwen-1.5B which we also studied in \cref{sec:conversion}. The unprepared learner is Qwen2.5-1.5B-Instruct. ``Prepared Qwen'' starts from the same Qwen 1.5B instruct model after four passes of offline distillation on OpenMathInstruct2 (see the preparation section, \cref{app:preparation} for more details). The Qwen runs use 200 training steps, 256 rollouts per update, temperature 1.0, and an 8,192-token generation cap. DeepSeek's generation cap is 4,096 tokens.

\paragraph{Rollout banks.}
We save 32 rollouts from each batch $\ell\in\{1,21,41,\ldots,161,181,196\}$. A bank is this fixed set of rollout texts. Batch $\ell$ is generated after $\ell-1$ updates. Thus batch 41 appears at update 40 in the routing and drift plots. Checkpoint 0 is each learner's starting model, including the offline-prepared model for prepared Qwen.

\paragraph{When does the model initiate checking?}
The WAIT family $\mathcal W$ contains the 13 tokens that decode to ``wait'' (after whitespace is removed and text is lowercased). We examine candidate sentence starts: the first response position and positions after a token ending in a period, question mark, exclamation mark, colon, or newline. The student and teacher probabilities of this family are denoted as $w_t(x)=p_t(\mathcal W\mid x)$ and $q_W(x)=q(\mathcal W\mid x)$. We measure the mean $\log w_t$ across candidate starts. For the above- and below-teacher comparisons, we use log-odds $b_t=\log[w_t/(1-w_t)]$ and $b_q=\log[q_W/(1-q_W)]$.

We form the two groups using a reference checkpoint $r$: checkpoint 0 for DeepSeek and checkpoint 20 for the Qwen learners. We include a prefix only if $\max(w_r,q_W)\ge10^{-3}$ and $|b_r-b_q|\ge0.5$. The sign of $b_r-b_q$ determines whether the student initially assigns more or less probability than the teacher. Each prefix keeps the same group at every evaluated checkpoint.

We also count emitted WAIT tokens per thousand response tokens, and measure the arithmetic mean of $w_t$ at candidate starts. These quantities answer different questions. The emission rate counts what the model actually generates. Mean probability and mean log probability describe its actual predictions at the selected prefixes. 

To assess agreement with the teacher, we use the binary discrepancy $c_W=\KL{\operatorname{Ber}(q_W)}{\operatorname{Ber}(w_t)}$. At a fixed prefix, $\partial_t c_W=(w_t-q_W)\partial_t b_t$. Lowering WAIT probability improves agreement only while the student's probability is above the teacher's.

\begin{table}[ht]
\centering\small
\setlength{\tabcolsep}{4pt}
\begin{tabular}{lrrrrrrrr}
\toprule
&\multicolumn{4}{c}{Local loss}&\multicolumn{4}{c}{Conditional entropy}\\
\cmidrule(lr){2-5}\cmidrule(lr){6-9}
Learner & Initial & Final & $\sum\widehat R$ & $\sum\widehat D$ & Initial & Final & $\sum\widehat R$ & $\sum\widehat D$\\
\midrule
DeepSeek & 0.263 & 0.015 & -0.033 & -0.214 & 0.664 & 0.707 & -0.049 & 0.093\\
Unprepared & 0.579 & 0.073 & 0.474 & -0.980 & 0.650 & 3.988 & 3.666 & -0.327\\
Prepared & 0.707 & 0.362 & 0.108 & -0.453 & 2.171 & 0.842 & -1.183 & -0.146\\
\bottomrule
\end{tabular}
\caption{\textbf{Cumulative routing and drift contributions to loss and entropy.} The comparisons run from initialization through batch 161, including evaluations at substitute checkpoints. Values are rounded independently. At full precision, the routing and drift contributions sum to the change from the initial to the final mean.}\label{tab:rd-summary}
\end{table}

\subsection{DeepSeek: adjusting the probability of checking}
\label{app:rd-checking}

DeepSeek's local loss falls mainly because its predictions change at fixed prefixes (\cref{tab:rd-summary}). We next ask whether these changes bring its checking probabilities closer to the teacher's, as``wait''-initiated checking / backtracking is a dominant part of the model's reasoning traces and observed to initiate sentences in which the model manage its epistemic states \citep{marjanovic2026deepseekr}. We use the classes in \cref{app:rd-protocol}, defined by whether the reference student's WAIT probability lies above or below the teacher's.

In the first 512 tokens, the above-teacher class has $\sum\widehat R=0.036$ and $\sum\widehat D=-1.824$. The below-teacher class has $0.505$ and $1.918$, respectively. The drift contribution therefore lowers mean checking log-odds in the first class and raises them in the second. The main plot uses filled circles and open triangles for these two classes. Routing can still be large within individual intervals, even when its contributions partly cancel over training.

We check whether this two-sided adjustment reduces the discrepancy from the teacher on a fixed set of prefixes (\cref{tab:rd-check-calibration}). For DeepSeek, the discrepancy falls from $0.0789$ to $0.0032$ in the above-teacher class and from $0.1915$ to $0.0149$ in the below-teacher class. Interestingly, the same comparison on held-out prepared-Qwen text gives decreases from about $0.039$ to $0.014$ and from $0.293$ to $0.039$, so the improvement even extends beyond DeepSeek's own generated prefixes.

To examine how closely the changes follow the teacher, we regress the change in student log-odds on the teacher and reference-student log-odds. On DeepSeek's own prefixes at checkpoint 200, the teacher coefficient is about $0.90$, the reference-student coefficient about $-0.88$, and $R^2$ about $0.94$. This detailed relation is much weaker on the held-out Qwen prefixes. The fixed-prefix results support improved calibration on both sets, but the closer match to the teacher's pattern occurs on DeepSeek's own text.

\begin{table}[ht]
\centering\small
\begin{tabular}{llrrrr}
\toprule
Learner & Reference class & Positions & $c_W$ at reference & $c_W$ at 200 & $\Delta$ log-odds\\
\midrule
DeepSeek & Above teacher & 33,010 & 0.0789 & 0.0032 & -2.073\\
DeepSeek & Below teacher & 11,336 & 0.1915 & 0.0149 & +1.381\\
Prepared Qwen & Above teacher & 25,794 & 0.0276 & 0.0580 & -0.606\\
Prepared Qwen & Below teacher & 18,211 & 0.5627 & 0.3813 & +0.721\\
\bottomrule
\end{tabular}
\caption{\textbf{Checking probability on fixed prefixes.} We report discrepancy from the teacher, $c_W$, and the change in log-odds from the reference checkpoint to checkpoint 200. The reference is checkpoint 0 for DeepSeek and checkpoint 20 for prepared Qwen. Each class contains the same prefixes at both checkpoints. The main-text routing and drift plots instead compare successive banks.}\label{tab:rd-check-calibration}
\end{table}

\begin{figure}[ht]
\centering
\includegraphics[width=\linewidth]{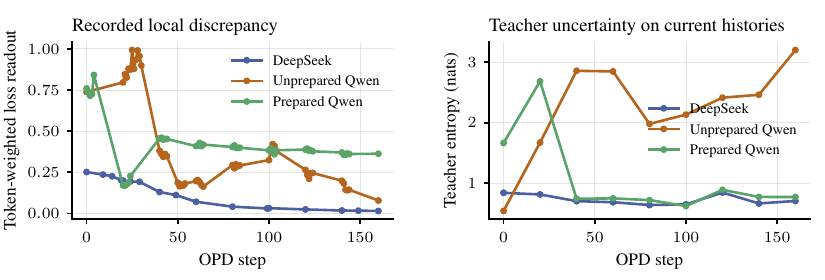}
\caption{\textbf{Local agreement and teacher uncertainty.} Left: local loss averaged over tokens in the full recorded batches, through the period with fixed prompt-language composition. Right: teacher entropy on the 32-rollout samples from each batch. Colors match \cref{fig:rd-contrast,fig:rd-prepared}. The teacher is generally less uncertain on DeepSeek and prepared-Qwen prefixes than on unprepared-Qwen prefixes.}
\label{fig:rd-support-overview}
\end{figure}

\subsection{Unprepared Qwen 1.5B: fitting on changing prefixes}
\label{app:rd-failure}

For unprepared Qwen, the local loss decreases ($0.579\to0.073$). The accumulated contributions are $\sum\widehat R_L=+0.474$ and $\sum\widehat D_L=-0.980$. Thus changes in predictions at fixed prefixes account for the decrease, while changes in prefix visitation raise the loss. This balance varies during training. In the first comparison at step 20, loss rises by $+0.268$: routing contributes $+0.455$, which exceeds the decrease from drift, $-0.187$. Later on, loss mainly falls through changes dominated by drift, as the main text already pointed out. The large rise in entropy has a different origin: it comes mainly from prefix visitation, as shown in the main text.

\subsubsection{Loss and entropy on the same prefixes}
\label{app:rd-entropy-evidence}
The preceding comparisons use batches generated at different training steps. We now hold the prefixes fixed to see what the student learns on the same text. We use the prefixes from batch 41. (The entropy-response identity, how it can change through routing without significant drift, and examples are discussed in the theory presented in \cref{app:rd-entropy-response}.)

On these prefixes, student entropy falls from $4.987$ nats at initialization to $3.989$ at checkpoint 160. The local loss falls from $0.789$ to $0.057$. Teacher entropy is $2.859$ nats, below both student values. The teacher is more uncertain here than on the coherent initial text, but the student still learns sharper predictions on these prefixes.

These measurements answer different questions. The falling loss shows fitting at fixed prefixes. The entropy decomposition shows that the large rise in uncertainty mainly reflects which prefixes the model visits. Note that these are primarily examinations of how the local per-prefix readouts change. To understand how the learnt model's continuations reach a correct answer we need to probe their continuation advantages which involve non-local readouts, as \cref{eq:rd-value-rate} shows. We examine this next.

\FloatBarrier
\Needspace{10\baselineskip}
\subsection{Continuation value at the unprepared student's answer boundaries}
\label{app:steer-probe}
We now examine prefixes at which the initial student was about to answer. Holding these prefixes fixed lets us ask two questions: how does training change the probability of answering now, and how successful are the continuations from the two alternatives? This comparison isolates changes at the selected prefixes from changes in prefix visitation.

\paragraph{Prefix selection and continuation conditions.}
We sample two rollouts per problem from the initial student on 50 held-out MATH-500 problems. ``Cutting'' the rollout immediately before a final-answer marker (\texttt{\textbackslash boxed} or \texttt{Answer:}) gives 86 prefixes. We evaluate the initial student, checkpoints 20--200 of the unprepared run, and the JustRL teacher on these same prefixes. For each model, we record the next-token probability mass $\pi$ on the probe's answer class.

We draw six continuations per prefix under each of \emph{three conditions}: the \emph{answer} condition forces the answer token, the ``continue'' condition forces the model's most probable non-answer token, and the ``natural'' condition samples without forcing either token. All conditions use temperature $0.6$, \mbox{top-$p=0.95$}, and a cap of $16{,}384$ new tokens. \Cref{tab:steer-probe} reports the fraction of correct continuations.

\paragraph{Branch probabilities and continuation values.}
Let $Q_{\rm ans}$ and $Q_{\rm cont}$ denote the correctness values of the two forced branches, ``answer'' vs ``continue'' (recall that $Q$ denotes the continuation success rate from the given prefix distribution after forcing an action choice). Their probability-weighted probe value is $V=\pi Q_{\rm ans}+(1-\pi)Q_{\rm cont}$. This value's change relative to the initial student decomposes as
\[
\Delta V=
\underbrace{\Delta\pi\,(\bar Q_{\rm ans}-\bar Q_{\rm cont})}_{\text{changed branch probability}}
+\underbrace{\bar\pi\,\Delta Q_{\rm ans}+(1-\bar\pi)\,\Delta Q_{\rm cont}}_{\text{changed branch values}},
\]
where bars denote the average of the two endpoint values at the same prefix. The first term describes moving probability between the two branches. It is positive when probability moves \emph{toward} the more successful branch. The second term describes changes in the branch continuation values. We calculate both terms at each prefix before averaging, and obtain intervals from $2{,}000$ bootstrap draws over prefixes. Note that this split applies to the two forced branches. The most probable non-answer token can differ between models. The continue-branch value therefore includes both that first choice and the later continuation. Natural continuations are sampled separately, and their success rate need not equal $V$.

\begin{table}[htb]
\centering\small\setlength{\tabcolsep}{4pt}
\begin{tabular}{lrrrrrrr}
\toprule
 & log-odds of & \multicolumn{3}{c}{correctness} & \multicolumn{3}{c}{change in two-branch value}\\
\cmidrule(lr){3-5}\cmidrule(lr){6-8}
Evaluator & answering & answer & continue & natural & $\Delta V$ & probability term & value term\\
\midrule
initial student & $+2.4$ & 0.30 & 0.22 & 0.27 & --- & --- & ---\\
update 20  & $-3.9$ & 0.30 & 0.29 & 0.29 & $+0.01$ & $-0.03$ & $+0.05$\\
update 30  & $-4.9$ & 0.33 & 0.32 & 0.31 & $+0.04$ & $-0.03$ & $+0.07$\\
update 40  & $-6.1$ & 0.29 & 0.27 & 0.26 & $-0.01$ & $-0.04$ & $+0.03$\\
update 50  & $-6.7$ & 0.31 & 0.17 & 0.21 & $-0.09$ & $-0.08$ & $-0.02$\\
update 100 & $-7.5$ & 0.34 & 0.03 & 0.06 & $-0.24$ & $-0.15$ & $-0.09$\\
update 200 & $-7.4$ & 0.39 & 0.13 & 0.17 & $-0.15$ & $-0.13$ & $-0.02$\\
teacher    & $-3.5$ & 0.42 & 0.58 & 0.60 & $+0.31$ & $+0.04$ & $+0.27$\\
\bottomrule
\end{tabular}
\caption{\textbf{Answer probability and continuation success on 86 fixed prefixes.} Log-odds are averaged over prefixes. The answer and continue columns use forced tokens; natural uses unconstrained continuations. The final columns give the change in the two-branch value and the contributions from changes in branch probability and branch values. We compute these at each prefix before averaging. At update 100, the bootstrap intervals are $[-0.32,-0.16]$ for $\Delta V$, $[-0.20,-0.10]$ for the probability term, and $[-0.14,-0.04]$ for the value term. Values are rounded.}
\label{tab:steer-probe}
\end{table}

\paragraph{How the results change during training.}
Early on, at update 20, the mean log-odds of answering have fallen from $+2.4$ to $-3.9$. Yet the answer and continue branches have similar correctness, $0.30$ and $0.29$. The early shift toward continuing is therefore not yet accompanied by lower continuation success on these prefixes.

By update 100, the answer branch still succeeds with probability $0.34$, but the continue branch succeeds with probability $0.03$. Natural continuations succeed with probability $0.06$. The $8\%$ answer-producing rate in the main text is for the continue branch. It counts any answer, whereas $3\%$ counts a correct answer. Both terms in the two-branch decomposition are negative at this checkpoint. On balance, probability moves toward the less successful branch. Changes in branch values also reduce the probe value. Success partly recovers by update 200, but natural-continuation correctness remains below its initial value.

The teacher also favors continuing over answering on these prefixes. Its continue branch, however, is more successful than its answer branch ($0.58$ versus $0.42$), and its natural continuations succeed with probability $0.60$. This illustrates the role of continuation values in \cref{eq:broadcast-continuations}. In the probability term, a shift toward continuing helps when continuing is more successful than answering now. These answer-boundary probes examine that relation on fixed prefixes. \Cref{app:fixed-reference-continuations} then uses fixed reference continuers to examine changes in prefix visitation.

\FloatBarrier
\Needspace{10\baselineskip}
\subsection{Prepared Qwen: how checking changes during training}
\label{app:rd-prepared}

The prepared run improves its held-out score, even though its rollouts change markedly early in training. We first examine the order of these changes. We then ask whether the later decrease in checking also occurs on unchanged prefixes, how the teacher's requests change, and which representative ``calibration errors'' remain.

\paragraph{The order of the early changes.}
Entropy rises early on. At step 20, entropy increases from $2.17$ to $4.33$ nats. Routing contributes $+1.98$ and drift $+0.18$, while local loss falls from $0.707$ to $0.173$. In the next comparison, from batch 21 to batch 41, loss overshoots to $0.437$. Its routing contribution is $+0.363$ and its drift contribution $-0.099$. Entropy on batch 41 is already low (about $0.70$ nats), and remains low (about $0.6$--$0.9$) on later batches. Thus the loss overshoot follows the initial entropy rise. It mainly reflects a change in prefix visitation, similar to the early phase of the unprepared run.

\begin{figure}[ht]
\centering
\includegraphics[width=\linewidth]{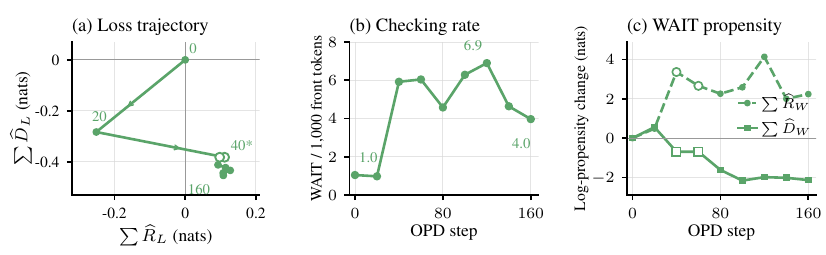}
\caption{\textbf{Loss and checking during the prepared run.} Left: cumulative routing and drift contributions to local loss. Middle: WAIT-family tokens per thousand response tokens, measured in the first 512 tokens. Right: cumulative routing (dashed) and drift (solid) of mean log-WAIT probability at candidate sentence starts in the same window. The last panel decomposes log probability, which is distinct from emitted frequency. Open markers denote substituted evaluator checkpoints.}
\label{fig:rd-prepared-full}\label{fig:rd-prepared}
\end{figure}

\begin{figure}[ht]
\centering
\includegraphics[width=\linewidth]{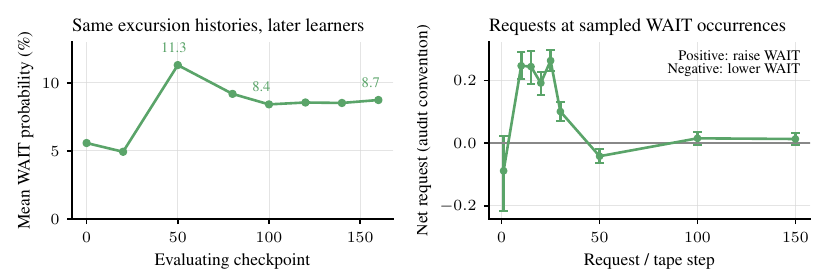}
\caption{\textbf{Changes on fixed prefixes and changes in teacher requests.} Left: mean WAIT probability at the same sentence starts in the first 512 tokens of batch-41 rollouts, evaluated by each checkpoint. Right: mean request at sampled WAIT occurrences before an answer in full training batches, with the reported 95\% intervals. Positive requests favor the sampled WAIT token; negative requests suppress it. The two panels use different prefix sets.}
\label{fig:rd-prepared-audit}
\end{figure}

The WAIT emission rate also rises and then falls (\cref{fig:rd-prepared-full}, middle). To see whether the model changes its predictions on the same prefixes, we evaluate each checkpoint on the candidate sentence starts in the first 512 tokens of batch 41 (\cref{fig:rd-prepared-audit}, left). Mean WAIT probability rises from $0.0494$ at checkpoint 20 to $0.1130$ at checkpoint 50. It then falls to $0.0842$ at checkpoint 100 and is $0.0874$ at checkpoint 160. The later decrease therefore includes a change in predictions on unchanged prefixes. The probability remains above its level at checkpoint 20.

Mean log-WAIT probability gives another view of the change across rollout batches (\cref{fig:rd-prepared-full}, right). Drift is positive in the first comparison ($+0.56$) and negative by checkpoint 50 ($-1.25$). Routing is positive during the expansion ($+2.89$). Further decreases from drift accumulate through checkpoint 100. Later changes, including much of the decline between batches 121 and 141, mainly come from prefix visitation. These contributions sum to a change in mean log probability, not in emitted frequency. On some other fixed prefix sets, mean log probability and arithmetic mean probability change in opposite directions. We therefore keep these measurements separate.

\paragraph{How the teacher's requests change.}
Overall, the requests change from favoring WAIT to suppressing it on average. More specifically, we examine requests at sampled WAIT occurrences in full training batches (\cref{fig:rd-prepared-audit}, right). The mean request is about $+0.19$ to $+0.26$ at batches 10--25, with intervals above zero. At batch 50, it is $-0.0418$ (95\% interval $[-0.0635,-0.0182]$), and $0.528$ of the sampled requests are negative. Later means are nearer zero. 

\Needspace{9\baselineskip}
\paragraph{Remaining calibration errors.}
We finally ask whether these changes bring checking probabilities closer to the teacher's. The prefix classes are fixed at checkpoint 20, and we compare checkpoints 20 and 200 on the same pooled prefix set (\cref{tab:rd-check-calibration}). The discrepancy increases from $0.0276$ to $0.0580$ in the above-teacher class. It decreases from $0.5627$ to $0.3813$ in the below-teacher class. On average, calibration improves only in the initially below-teacher class.

Over-use also grows within the above-teacher subset with $q_W\ge0.05$. Mean student probability rises approximately $0.33\to0.50$ on its own text and $0.29\to0.46$ on held-out text, against teacher probability about $0.15$. The changes still depend on teacher probability after accounting for the reference student's probability, including on held-out prefixes. Thus the student's adjustments remain related to the teacher, but the calibration improvement appears imperfect and weaker than the DeepSeek run.

\FloatBarrier
\Needspace{10\baselineskip}
\subsection{Uncertainty across rollout depths}
\label{app:rd-sensitivity}\label{app:broadcast-evidence}

\begin{figure}[ht]
\centering
\begin{subfigure}{.49\linewidth}
\includegraphics[width=\linewidth]{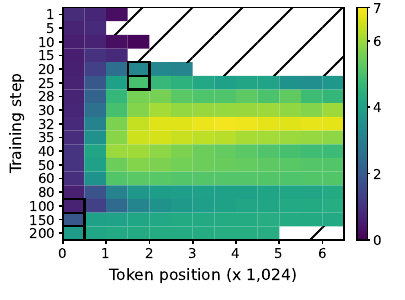}
\caption{Student surprisal for sampled tokens.}\label{fig:broadcast-entropy}
\end{subfigure}\hfill
\begin{subfigure}{.49\linewidth}
\includegraphics[width=\linewidth]{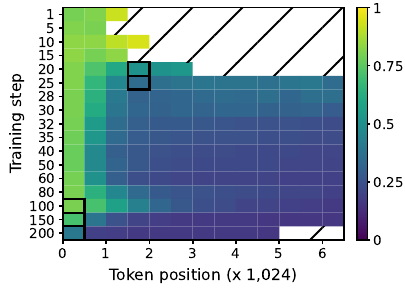}
\caption{Teacher top-1 probability.}\label{fig:broadcast-teacher}
\end{subfigure}
\caption{\textbf{Uncertainty at different response depths during training.} Student surprisal and teacher top-1 probability are measured on the same student-generated prefixes, in 512-token windows. Hatched cells have no reported value. The plots describe the observed training trajectory.}\label{fig:broadcast-onset}
\end{figure}

We next examine when uncertainty rises and where it appears in the response. \Cref{fig:broadcast-onset} uses full training batches from the unprepared Qwen run. Each training update samples 256 rollouts on DAPO prompts at temperature one, with an 8,192-token generation cap.

The left heatmap reports student surprisal: the negative log probability of the token the student generated. The right heatmap reports the teacher's highest next-token probability at the same prefix. Surprisal uses the sampled token, whereas the entropy measurements above use the full next-token distribution. The heatmaps summarize 512-token windows. Rows index rollout batches, including those labeled ``tape step'' in the main figure. Batch $\ell$ follows $\ell-1$ updates. Hatched cells and table dashes indicate that no value is reported there, as the model does not reach that generation length yet.

\Cref{tab:broadcast-onset} gives two example windows. Between batches 20 and 35, surprisal at positions 1536--2047 rises from $3.24$ to $6.47$ nats. In the first 512 tokens, it rises only from $0.62$ to $0.90$. Teacher top-1 probability falls from $0.51$ to $0.29$ in the deeper window, compared with $0.81$ to $0.78$ in the front window. Thus the early increase in uncertainty is much larger in the deeper of these two windows. The change is not monotone: deep-window surprisal falls to $3.19$ by batch 100. Our main routing and drift analysis ends at batch 161.

\begin{table}[ht]
\centering\small
\setlength{\tabcolsep}{6pt}
\begin{tabular}{r rr rr}
\toprule
&\multicolumn{2}{c}{Student surprisal (nats)}&\multicolumn{2}{c}{Teacher top-1 probability}\\
\cmidrule(lr){2-3}\cmidrule(lr){4-5}
Batch & First 512 & 1536--2047 & First 512 & 1536--2047\\
\midrule
1 & .85 & --- & .79 & ---\\
5 & .69 & --- & .81 & ---\\
10 & .58 & .12 & .83 & .94\\
15 & .58 & --- & .83 & ---\\
20 & .62 & 3.24 & .81 & .51\\
25 & .66 & 4.97 & .81 & .35\\
28 & .70 & 5.55 & .80 & .31\\
30 & .74 & 5.71 & .80 & .33\\
32 & .83 & 6.38 & .79 & .30\\
35 & .90 & 6.47 & .78 & .29\\
40 & 1.04 & 6.00 & .77 & .27\\
50 & 1.05 & 5.69 & .77 & .27\\
60 & .95 & 5.20 & .79 & .24\\
80 & .65 & 3.70 & .80 & .34\\
100 & .66 & 3.19 & .80 & .43\\
150 & 1.89 & 4.14 & .71 & .20\\
200 & 3.85 & 4.26 & .40 & .17\\
\bottomrule
\end{tabular}
\caption{\textbf{Two response windows from the heatmaps.} The first column gives the rollout batch. Batch $\ell$ is generated after $\ell-1$ updates. Student values are sampled-token surprisal. A dash indicates no reported value.}
\label{tab:broadcast-onset}
\end{table}

% Supplement for Appendix E (sec_app_routing_drift.tex).
% Prose paragraphs are stored as single source lines, without manual breaks.
%
% PROVENANCE / INTEGRATION NOTES:
% The continuation design and rounded contrasts were supplied in the editing conversation. The original Overleaf archive contains neither the raw continuation records nor the aggregate export. These are transcribed point estimates, not a new computation from raw outcomes.
% Confirm decoder settings, stopping rules, inherited-prefix serialization, total-budget handling, and grader implementation against the experiment records. No settings or uncertainty intervals were inferred from the separate answer-boundary study.
% In particular, verify that the exported averages retain earlier terminal responses with their original scores and the full rollout-bank denominator. A mean conditional on surviving to the cut is a different quantity.

\newpage
\subsection{Fixed-reference continuation experiments}
\label{app:continuation-profile}
\label{app:fixed-reference-continuations}

The answer-versus-continue experiment in \cref{app:steer-probe} conducted a case study probing continuation from fixed prefixes. This, however, is not enough in understanding the decline in terminal score of the trained student model. Recall that, at a high level, the broadcast not only affects continuation behaviors of the model, but also changes its prefix visitation probabilities. Therefore, to understand the terminal-score decline during OPD, we must also consider how the prefixes that the student checkpoints generates as training proceeds. At any rollout depth, successful completion depends on two quantities: which prefixes the student reaches, and how successfully it continues from those prefixes. We first express terminal score in terms of these two quantities, then show how the reference-policy experiment separates their changes.

\paragraph{Prefix visitation and continuation.}
Fix a cut depth $h$. Recall our notation from before, that $d_t(x)$ is the probability that checkpoint $t$ reaches prefix $x$, and $V_t^Y(x)$ is its expected terminal score when it completes that prefix. The set of rollout cuts $\mathcal C_h$ includes depth-$h$ prefixes and responses that ended earlier (an earlier \emph{completed} response just retains its terminal score). Conditioning on $\mathcal C_h$ gives
\[
 J_Y(t)=\sum_{x\in\mathcal C_h}d_t(x)V_t^Y(x).
\]
This is just the usual average over the ``continuation problems'' presented to the student by its own generation. Training changes both the weights $d_t$ and the continuation values $V_t^Y$. Comparing each checkpoint's prefixes using that checkpoint as the continuer therefore only measures their \emph{combined} effect.

\paragraph{Hold the continuer fixed.}
Using the initial student as the reference, add and subtract $\sum_{x\in\mathcal C_h}d_t(x)V_0^Y(x)$ to obtain
\begin{equation}
 \begin{aligned}
 J_Y(t)-J_Y(0)
 &=\underbrace{\sum_{x\in\mathcal C_h}[d_t(x)-d_0(x)]V_0^Y(x)}_{\text{prefix visitation, evaluated by the initial continuer}}
 \quad+\underbrace{\sum_{x\in\mathcal C_h}d_t(x)[V_t^Y(x)-V_0^Y(x)]}_{\text{continuation change on the same current prefixes}}.
 \end{aligned}
 \label{eq:continuation-reference-split}
\end{equation}
The first term \emph{compares the initial and trained prefix distributions} while keeping the continuer fixed. A negative value means that the initial student finds the trained prefixes harder to finish than its own prefixes. The second term keeps the trained prefix distribution fixed and compares the two continuers. A negative value means that the trained student is less successful than the initial student on those same prefixes. Thus the comparison changes the prefix-generating policy first, then the continuing policy.

Our experiment implements the intermediate case directly: checkpoint $t$ generates the first $h$ tokens, and the initial student completes the rest of the rollout. Let us write its expected score as
\begin{equation}
 F_t^0(h)=\sum_{x\in\mathcal C_h}d_t(x)V_0^Y(x).
 \label{eq:continuation-handoff}
\end{equation}
The two terms in \cref{eq:continuation-reference-split} are therefore $F_t^0(h)-J_Y(0)$ and $J_Y(t)-F_t^0(h)$. They can be estimated by averaging completed responses. The first supplies a common measure of prefix difficulty across checkpoints; the second compares continuers on each checkpoint's own prefix distribution. 

\paragraph{Experimental setup and results.}
The study uses 250 held-out MATH problems, two source rollouts per problem from each of checkpoints 50, 100, and 200, and cuts at 256, 512, and 1,024 response tokens. Each active prefix is continued four times by the generating checkpoint, the initial student, and the teacher. All continuers receive the remainder of an 8,192-token total response budget. The readout is extracted-answer success on the full prefix and continuation; the answer need not be immediately followed by an end-of-turn token. Stopping is recorded separately. The estimator averages continuation scores within each prefix, then over the two source rollouts and uniformly over problems, retaining earlier completed responses as above. This population and budget differ from those of the selected answer-boundary probe in \cref{app:steer-probe}. The reported values below are rounded sample means; independently sampled self-continuations estimate the same $J_Y(t)$ but need not give identical means at different cuts.
% This experiment does not inherit T=0.6, top-p=0.95, or the 16,384-new-token budget from app:steer-probe. Raw records are needed to verify the aggregation convention and estimate uncertainty while preserving prompt and prefix pairing.

Consider the 256-token cut at step 50. The initial student scores $48.4\%$ on its own rollouts and $50.5\%$ when it completes the trained checkpoint's prefixes. The trained checkpoint scores $39.5\%$ on those same prefixes. The first comparison therefore contributes $+2.1$ percentage points, while the second contributes $-11.0$ points, accounting for the total $-8.9$-point difference. At this cut, the prefixes remain usable by the initial student; the trained student loses performance in completing them.

\begin{table}[ht]
\centering\small
\setlength{\tabcolsep}{8pt}
\begin{tabular}{r rrr}
\toprule
Cut depth & Step 50 & Step 100 & Step 200\\
\midrule
256   & $+2.1$ & $-1.7$ & $-5.9$\\
512   & $-1.3$ & $-2.7$ & $-6.1$\\
1,024 & $-8.1$ & $-7.6$ & $-8.5$\\
\bottomrule
\end{tabular}
\caption{Change in initial-policy completion success, $100[F_t^0(h)-J_Y(0)]$, in percentage points. The initial student completes prefixes from each trained checkpoint; its own-rollout baseline is $48.4\%$. These are the values plotted in \cref{fig:rd-initial-continuation}, reported as rounded aggregate point estimates from one training run.}
\label{tab:continuation-reference}
\end{table}

\Cref{tab:continuation-reference} shows that at step 50, the initial continuer already scores lower on deep prefixes: the 1,024-token contrast is $-8.1$ points. At later checkpoints, the deficit also reaches the shallow cuts, while the 1,024-token deficit remains similar in magnitude. On the 256-token prefixes, the trained-minus-initial continuation contrast remains negative at steps 100 and 200 ($-12.1$ and $-6.2$ points). By step 200, both terms in \cref{eq:continuation-reference-split} are negative: the initial student finds the new prefixes harder, and the trained student completes those prefixes less successfully than the initial student.

\paragraph{Connection to \Cref{eq:broadcast-continuations}.}
The same two factors describe the instantaneous dynamics. Differentiating the cut identity gives
\begin{equation}
 \dot J_Y(t)=\sum_{x\in\mathcal C_h}\dot d_t(x)V_t^Y(x)+\sum_{x\in\mathcal C_h}d_t(x)\dot V_t^Y(x).
 \label{eq:continuation-infinitesimal}
\end{equation}
Under the regularity assumptions in \cref{prop:rd-value-cut}, the first term computes the contributions to \cref{eq:broadcast-continuations} from decisions before the cut, and the second collects those from the cut onward. Both arise from the same shared parameter update. If the reference is checkpoint $t$ instead of checkpoint 0, and the compared policy is an infinitesimally later checkpoint $t+dt$, the differences in \cref{eq:continuation-reference-split} (divided by elapsed training time) recover the derivative, hence recovering these two terms. Such measurements are unrealistic to implement and too noisy to measure in practice, however, so our experiment instead use the initial policy as the reference continuer to magnify the trend of visited-prefix quality / continuation difficulty.

\paragraph{What does moving the cut measure?}
Another interesting question is, at a fixed checkpoint $t$, moving the rollout cut from $h_1$ to $h_2$ lets the trained student generate an additional block of tokens before the initial student takes over. A decrease in $F_t^0$ means that this additional trained-student generation leaves the initial continuer less able to reach a correct answer. To connect it to continuation advantages, let us follow the initial student's value along a trained rollout. For each action decision, $\mathcal A_0^Y(x_j,a_j)=V_0^Y(x_{j+1})-V_0^Y(x_j)$. Summing over a block yields a telescoping series over depth of the cuts, then averaging over the rollouts gives
\begin{equation}
 F_t^0(h_2)-F_t^0(h_1)=\sum_{\substack{x\in\mathcal S\\h_1\le h(x)<h_2}}d_t(x)\Ex_{a\sim\pi_t(\cdot\mid x)}[\mathcal A_0^Y(x,a)].
 \label{eq:continuation-block-identity}
\end{equation}
Here $\mathcal S$ contains decision prefixes; completed responses contribute no further decisions and retain their scores. At depth zero, the initial student generates the whole response, so $F_t^0(0)=J_Y(0)$. At the total response cap, the trained student has generated the whole response, so $F_t^0(H_{\max})=J_Y(t)$. Consequently, the differences across any partition of the full response sum to $J_Y(t)-J_Y(0)$. Fixed cuts therefore measure aggregate contributions of token blocks to this finite policy comparison, using completed outcomes in place of separate continuation-value estimates at every token.

\FloatBarrier
%\Needspace{12\baselineskip}
%\input{sec_app_broadcast}
%\FloatBarrier
\Needspace{12\baselineskip}
\section{Preparation: protocols and additional measurements}
\label{app:preparation}

% Three-family preparation comparison.
\subsection{Three-family AIME comparison}
\label{app:preparation-multisetup}
% AIME statistics use n = 32 responses per problem, batch a only.
% Regenerate tables with scripts/regenerate_preparation.py and figures with scripts/plot_preparation.py.
% Matched continuation budgets: 200 steps for Qwen 1.5B and Llama, 100 for Qwen 7B.

\paragraph{Students and teachers.}
The pairs are Qwen2.5-1.5B-Instruct / JustRL-1.5B, Qwen2.5-7B-Instruct / AceReason-Nemotron-7B, and Llama-3.1-8B-Instruct / Llama-3.1-Nemotron-Nano-8B-v1. Each pair shares a tokenizer vocabulary. The fixed teacher supplies the preparation text, the offline-control targets, and the on-policy log-probabilities. The Llama teacher uses its reasoning mode with the system prompt ``detailed thinking on.'' 

\paragraph{Stage 1: offline preparation on OMI2.}
Each teacher generates solutions to problems drawn from OpenMathInstruct-2 \citep{toshniwal2024openmathinstruct2}, at temperature one with an 8,192-token response cap. The student uses token-level forward KL on the teacher's recorded top-16 next-token support. The plotted cold start uses 512 optimizer steps of 256 sequences, or 131,072 sequence presentations. The Qwen 1.5B, 7B and Llama corpora each contain 32,768 teacher solutions, so this is four corpus sweeps.

The offline settings are AdamW, learning rate $10^{-6}$, momentum coefficients $(0.9,0.95)$, no weight decay, gradient clipping at 1.0. The cold-start sequence limit is 8,192 tokens.

\paragraph{Stage 2: DAPO comparisons.}
The prompt pool is DAPO-Math-17k \citep{yu2025dapo}, recorded as 17,398 unique problems. Training uses consecutive blocks of 6,400 rows in the released ordering. Each block contains 6,235--6,318 unique problems.

OPD samples four student responses for each of 64 prompts per step, giving 256 responses and one optimizer step per update. Sampling is at temperature one, capped at 8,192 tokens for Qwen and 14,336 for Llama. 

The offline continuation instead uses four teacher-generated solutions for each prompt row in the same blocks: 25,600 sequences per 6,400-row block. It uses the teacher forward-KL objective and optimizer settings as preparation, at 256 sequences per step. The sequence limits are 8,192 for Qwen and 15,360 for Llama, the latter retaining the reported teacher solutions generated with a 14,336-token response cap. Thus the comparison matches the implemented offline and on-policy recipes in prompt blocks, sequence counts, and total optimizer steps.

\begin{table}[t]
\centering\small
\setlength{\tabcolsep}{5pt}
\begin{tabular}{lrrr}
\toprule
Student & Offline-control steps & Steps in the OPD arm & Sequence presentations per arm\\
\midrule
Qwen 1.5B & 200 offline & 200 OPD & 51,200\\
Qwen 7B & 100 offline & 100 OPD & 25,600\\
Llama 8B & 200 offline & 200 OPD & 51,200\\
\bottomrule
\end{tabular}
\caption{Comparison budgets from the same prepared checkpoint within each model pair, after 512 OMI2 preparation steps. Sequence presentations include repetitions.}
\label{tab:preparation-budgets}
\end{table}

\paragraph{Evaluation and sampled coverage.}
The evaluation pools AIME 2024 and 2025, 30 problems each. Each checkpoint is evaluated using 32 responses per problem from the first sampling batch (batch~a), with vLLM, the student's template, temperature $0.6$, top-$p=0.95$, and a 16,384-new-token cap. Scoring is lenient: answer extraction uses a boxed answer when present, otherwise an explicit answer marker including ``the answer is''.

Following \citet{chen2021evaluatingcode}, if $c_i$ of the 32 responses solve problem $i$, we estimate
\begin{equation}
\widehat{\mathrm{pass@}k}=\frac1{60}\sum_{i=1}^{60}\left[1-\frac{\binom{32-c_i}{k}}{\binom{32}{k}}\right],\qquad 1\le k\le32,
\label{eq:preparation-passk}
\end{equation}
with a zero numerator when $32-c_i<k$. In particular, pass@1 is the average observed success rate, and pass@32 is the fraction solved at least once in the sample. A problem never solved in 32 attempts can still lie within the model's support, and one solved on all 32 attempts can still have success probability below one.

The lower panels plot the empirical survival curve $60^{-1}\sum_i\mathbf1\{c_i/32\ge r\}$ for $r>0$. Its limit as $r\downarrow0$ is observed coverage and its area on $[0,1]$ is the average solve rate. At exactly $r=0$, the literal weak-inequality expression equals one; the plotted left endpoint represents the positive-threshold limit. These are marginal distributions across problems.

\paragraph{Endpoints and paired problem changes.}
\Cref{tab:preparation-aime-summary} gives the pooled pass@$k$ endpoints and the number of problems at each empirical reliability threshold.

\begin{table}[t]
\centering\small
\setlength{\tabcolsep}{5pt}
\begin{tabular}{llrrrrr}
\toprule
Student & Arm & pass@1 & pass@32 & Coverage & $\ge 1/2$ & All 32\\
\midrule
Qwen 1.5B & Base & $0.0115$ & $0.1333$ & 8 & 0 & 0\\
 & Cold start & $0.0099$ & $0.1333$ & 8 & 0 & 0\\
 & + offline & $0.0078$ & $0.0833$ & 5 & 0 & 0\\
 & + OPD & $0.0281$ & $0.1833$ & 11 & 0 & 0\\
\midrule
Qwen 7B & Base & $0.0953$ & $0.3167$ & 19 & 4 & 0\\
 & Cold start & $0.1302$ & $0.4167$ & 25 & 7 & 0\\
 & + offline & $0.1906$ & $0.4333$ & 26 & 10 & 1\\
 & + OPD & $0.2693$ & $0.6000$ & 36 & 15 & 5\\
\midrule
Llama 8B & Base & $0.0219$ & $0.1833$ & 11 & 1 & 0\\
 & Cold start & $0.0385$ & $0.2333$ & 14 & 1 & 0\\
 & + offline & $0.0349$ & $0.2333$ & 14 & 0 & 0\\
 & + OPD & $0.0693$ & $0.3500$ & 21 & 3 & 0\\
\bottomrule
\end{tabular}

\caption{Pooled AIME endpoints and observed reliability at 32 samples per problem (batch~a). Coverage, at-least-half, and all-32 columns count problems out of 60. Both Llama comparison arms start from the same prepared checkpoint and use the budgets in \cref{tab:preparation-budgets}.}
\label{tab:preparation-aime-summary}
\end{table}

Among problems solved at least once by both the OPD and offline endpoints, mean observed success rates (OPD/offline) are $0.281/0.104$ for Qwen 1.5B (3 problems), $0.599/0.455$ for Qwen 7B (25), and $0.264/0.159$ for Llama (13). These intersections are selected using both arms' samples, so their means are descriptive. Against the offline endpoint, OPD gains coverage on 8, 11, and 8 problems and loses it on 2, 1, and 1, respectively.

\paragraph{Uncertainty.}
For each student and year, we resample the 30 problem indices with replacement, keeping the OPD and offline outcomes paired. For pass@1 the per-problem difference is $(c_i^{\mathrm{OPD}}-c_i^{\mathrm{offline}})/32$; for pass@32 it is $\mathbf1\{c_i^{\mathrm{OPD}}>0\}-\mathbf1\{c_i^{\mathrm{offline}}>0\}$. \Cref{tab:preparation-aime-ci} reports the mean difference and the 2.5th and 97.5th percentiles of its exact conditional bootstrap distribution, computed by convolution of the empirical per-problem difference distribution. This avoids Monte Carlo error in the interval endpoints. The intervals quantify problem-resampling uncertainty conditional on the checkpoints and recorded outcomes.

\begin{table}[t]
\centering\small
\setlength{\tabcolsep}{5pt}
\begin{tabular}{llrr}
\toprule
Student & Year & $\Delta$pass@1 [95\% interval] & $\Delta$pass@32 [95\% interval]\\
\midrule
Qwen 1.5B & 2024 & $+0.032\ [+0.005,+0.065]$ & $+0.200\ [+0.033,+0.367]$\\
 & 2025 & $+0.008\ [-0.004,+0.027]$ & $+0.000\ [-0.100,+0.100]$\\
Qwen 7B & 2024 & $+0.086\ [+0.047,+0.130]$ & $+0.133\ [+0.033,+0.267]$\\
 & 2025 & $+0.071\ [+0.033,+0.114]$ & $+0.200\ [+0.033,+0.367]$\\
Llama 8B & 2024 & $+0.040\ [+0.005,+0.082]$ & $+0.100\ [+0.000,+0.233]$\\
 & 2025 & $+0.029\ [-0.006,+0.069]$ & $+0.133\ [+0.000,+0.300]$\\
\bottomrule
\end{tabular}

\caption{OPD minus offline control at 32 samples per problem (batch~a), with 95\% paired problem-bootstrap percentile intervals computed separately within each year. Values are in probability units.}
\label{tab:preparation-aime-ci}
\end{table}

\FloatBarrier

\subsection{Preparation-dose study}
\label{app:preparation-legacy}

We follow the same preparation (cold start) procedure as before, except we vary the number of passes over the cold-start corpus and study subsequent OPD performance. Evaluation uses the first 250 MATH-500 problems, four responses per problem, temperature $0.6$, top-$p=0.95$, a 16,384-new-token cap, and the deployment template. \Cref{tab:dose-figure} reports strict pass@1 and 95\% problem-bootstrap intervals with 2,000 resamples.

The 1.5B OPD series uses the step-100 checkpoint. The 7B series uses the best checkpoint evaluated every ten steps within the first 100 OPD steps: steps 30, 50, and 90 for one, two, and four preparation passes, respectively. Its intervals are computed at the selected checkpoint and do not adjust for checkpoint selection.

\begin{table}[htbp]
\centering
\small
\setlength{\tabcolsep}{5pt}
\begin{tabular}{lrrr}
\toprule
Student & Passes & Cold start & After OPD\\
\midrule
Qwen 1.5B
 & 1 & $0.459$ [0.411, 0.506] & $0.459$ [0.409, 0.509]\\
 & 2 & $0.457$ [0.409, 0.504] & $0.530$ [0.476, 0.585]\\
 & 4 & $0.471$ [0.420, 0.522] & $0.562$ [0.510, 0.614]\\
\midrule
Qwen 7B
 & 1 & $0.683$ [0.638, 0.725] & $0.823$ [0.782, 0.862]\\
 & 2 & $0.693$ [0.649, 0.736] & $0.828$ [0.791, 0.864]\\
 & 4 & $0.711$ [0.669, 0.753] & $0.853$ [0.817, 0.887]\\
\bottomrule
\end{tabular}

\caption{The dose panels of \cref{fig:preparation-aime}: strict MATH-500 pass@1 (first 250 problems, 4 samples per problem) after 1, 2 and 4 preparation passes. The OPD column uses step 100 at 1.5B and the best evaluated checkpoint within 100 steps at 7B. Brackets give 95\% problem-bootstrap intervals at each checkpoint.}
\label{tab:dose-figure}
\end{table}

\FloatBarrier
\end{document}